\RequirePackage{snapshot}
\documentclass[12pt]{article}

\usepackage[font=small,labelfont=bf]{caption}

\usepackage{url}
\usepackage{natbib}
\usepackage{booktabs}
\usepackage{longtable}

\usepackage[
  pdfstartpage=1,
  pdfpagemode=UseOutlines,
  bookmarks,
  bookmarksopen,
  pdfstartview=Fit,
  pdfview=Fit,
  colorlinks,
  linktocpage,
  linkcolor=blue,
  citecolor=blue,
  urlcolor=blue,
  pagebackref=true,
  hypertexnames=false
]{hyperref}

\usepackage{multicol}
\usepackage{multirow}
\usepackage{listings}
\usepackage{amsmath,amssymb}
\usepackage{tikz}
\usetikzlibrary{calc,positioning,arrows}

\usepackage{array}

\newcolumntype{M}[1]{>{\centering\arraybackslash}m{#1}}
\usepackage{caption} 
\usepackage{mathtools}
\usepackage{dsfont}
\usepackage[colorinlistoftodos]{todonotes}
\usepackage{booktabs}
\usepackage{xfrac}
\usepackage{bbm}
\usepackage{cleveref}

\usepackage{algorithm}
\usepackage{algorithmicx}

\usepackage[most]{tcolorbox}
\usepackage{xparse}
\usepackage{lipsum}
\usepackage{changepage}
\usepackage{enumitem}

\usepackage{pifont}
\newcommand{\yes}{\ding{51}}%
\newcommand{\no}{\ding{55}}%

\newcommand{\BnoV}{\textrm{B(no vis.)}}
\newcommand{\BnoL}{\textrm{B(no lang.)}}
\newcommand{\B}{\textrm{B}}
\newcommand{\BA}{\textrm{B$\cdot$A}}
\newcommand{\GA}{\textrm{G$\cdot$A}}
\newcommand{\BG}{\textrm{BG}}
\newcommand{\BGA}{\textrm{BG$\cdot$A}}
\newcommand{\mainagent}{\textrm{BGR$\cdot$A}}

\newcommand{\squishlist}{
   \begin{list}{$\bullet$}
    { \setlength{\itemsep}{0pt}      \setlength{\parsep}{3pt}
      \setlength{\topsep}{3pt}       \setlength{\partopsep}{0pt}
      \setlength{\leftmargin}{1.5em} \setlength{\labelwidth}{1em}
      \setlength{\labelsep}{0.5em} } }

\newcommand{\squishlisttwo}{
   \begin{list}{$\bullet$}
    { \setlength{\itemsep}{0pt}    \setlength{\parsep}{0pt}
      \setlength{\topsep}{0pt}     \setlength{\partopsep}{0pt}
      \setlength{\leftmargin}{2em} \setlength{\labelwidth}{1.5em}
      \setlength{\labelsep}{0.5em} } }

\newcommand{\squishend}{
    \end{list}  }

\newcommand{\myvecsym}[1]{\boldsymbol{#1}}

\newcommand{\vphi}{\myvecsym{\phi}}

\newcommand{\vtheta}{\myvecsym{\theta}}

\newcommand{\const}{\mbox{const}}

\newcommand{\KLpq}[2]{\mathrm{KL}\left[{#1}\|{#2}\right]}

\newcommand{\expect}[2]{\mathds{E}_{{#1}} \left[ {#2} \right]}

\DeclareMathAlphabet{\mathpzc}{OT1}{pzc}{m}{n}

\author{
\begin{tabular}{l r}
\end{tabular}\\
Interactive Agents Group\footnote{See Section~\ref{sec:authors} for Authors \& Contributions.} \\
DeepMind
}
\date{}
\title{
\rule[0.4cm]{\textwidth}{2pt}
{\bf Imitating Interactive Intelligence}
\rule{\textwidth}{2pt} 
}

\begin{document}
\maketitle

\begin{center}
{\bf Abstract} 
\end{center}
A common vision from science fiction is that robots will one day inhabit our physical spaces, sense the world as we do, assist our physical labours, and communicate with us through natural language. Here we study how to design artificial agents that can interact naturally with humans using the simplification of a virtual environment. This setting nevertheless integrates a number of the central challenges of artificial intelligence (AI) research: complex visual perception and goal-directed physical control, grounded language comprehension and production, and multi-agent social interaction. To build agents that can robustly interact with humans, we would ideally train them while they interact with humans. However, this is presently impractical. Therefore, we approximate the role of the human with another learned agent, and use ideas from inverse reinforcement learning to reduce the disparities between human-human and agent-agent interactive behaviour. Rigorously evaluating our agents poses a great challenge, so we develop a variety of behavioural tests, including evaluation by humans who watch videos of agents or interact directly with them. These evaluations convincingly demonstrate that interactive training and auxiliary losses improve agent behaviour beyond what is achieved by supervised learning of actions alone. Further, we demonstrate that agent capabilities generalise beyond literal experiences in the dataset. Finally, we train evaluation models whose ratings of agents agree well with human judgement, thus permitting the evaluation of new agent models without additional effort. Taken together, our results in this virtual environment provide evidence that large-scale human behavioural imitation is a promising tool to create intelligent, interactive agents, and the challenge of reliably evaluating such agents is possible to surmount. See videos for an \href{https://www.youtube.com/watch?v=b-fvsi9YIP4&feature=youtu.be}{overview} of the manuscript, \href{https://www.youtube.com/watch?v=NwzRR7XD898&feature=youtu.be}{training time-lapse}, and \href{https://www.youtube.com/watch?v=510xBEcef_o&feature=youtu.be}{human-agent interactions}.

\section{Introduction}
Humans are an interactive species. We interact with the physical world and with one another. We often attribute our evolved social and linguistic complexity to our intelligence, but this inverts the story: the shaping forces of large-group interactions selected for these capacities  \citep{dunbar1993coevolution}, and these capacities are much of the material of our intelligence. To build artificial intelligence capable of human-like thinking, we therefore must not only grapple with how humans think in the abstract, but also with how humans behave as physical agents in the world and as communicative agents in groups. Our study of how to create artificial agents that interact with humans therefore unifies artificial intelligence with the study of natural human intelligence and behaviour. 

This work initiates a research program whose goal is to build embodied artificial agents that can perceive and manipulate the world, understand and produce language, and react capably when given general requests and instructions by humans. Such a holistic research program is consonant with recent calls for more integrated study of the ``situated'' use of language \citep{mcclelland2019extending, lake2020wordmeaning}. Progress towards this goal could greatly expand the scope and naturalness of human-computer interaction \citep{winograd1972understanding, card1983psychology, gwern2018} to the point that interacting with a computer or a robot would be much like interacting with another human being -- through shared attention, gesture, demonstration, and dialogue \citep{tomasello2010origins,winograd1972understanding}. 

Our research program shares much the same spirit as recent work aimed to teach virtual or physical robots to follow instructions provided in natural language \citep{hermann2017grounded, lynch2020grounding} but attempts to go beyond it by emphasising the interactive and language production capabilities of the agents we develop. Our agents interact with humans and with each other by design. They follow instructions but also generate them; they answer questions but also pose them. 

\section{Our Research Program}

\subsection{The Virtual Environment}
We have chosen to study artificial agent interactions in a 3D virtual environment based on the Unity game engine \citep{ward2020using}. Although we may ultimately hope to study interactive physical robots that inhabit our world, virtual domains enable integrated research on perception, control, and language, while avoiding the technical difficulties of robotic hardware, making them an ideal testing ground for any algorithms, architectures, and evaluations we propose. 

\begin{figure}
    \centering
    \includegraphics[width=1.0\textwidth]{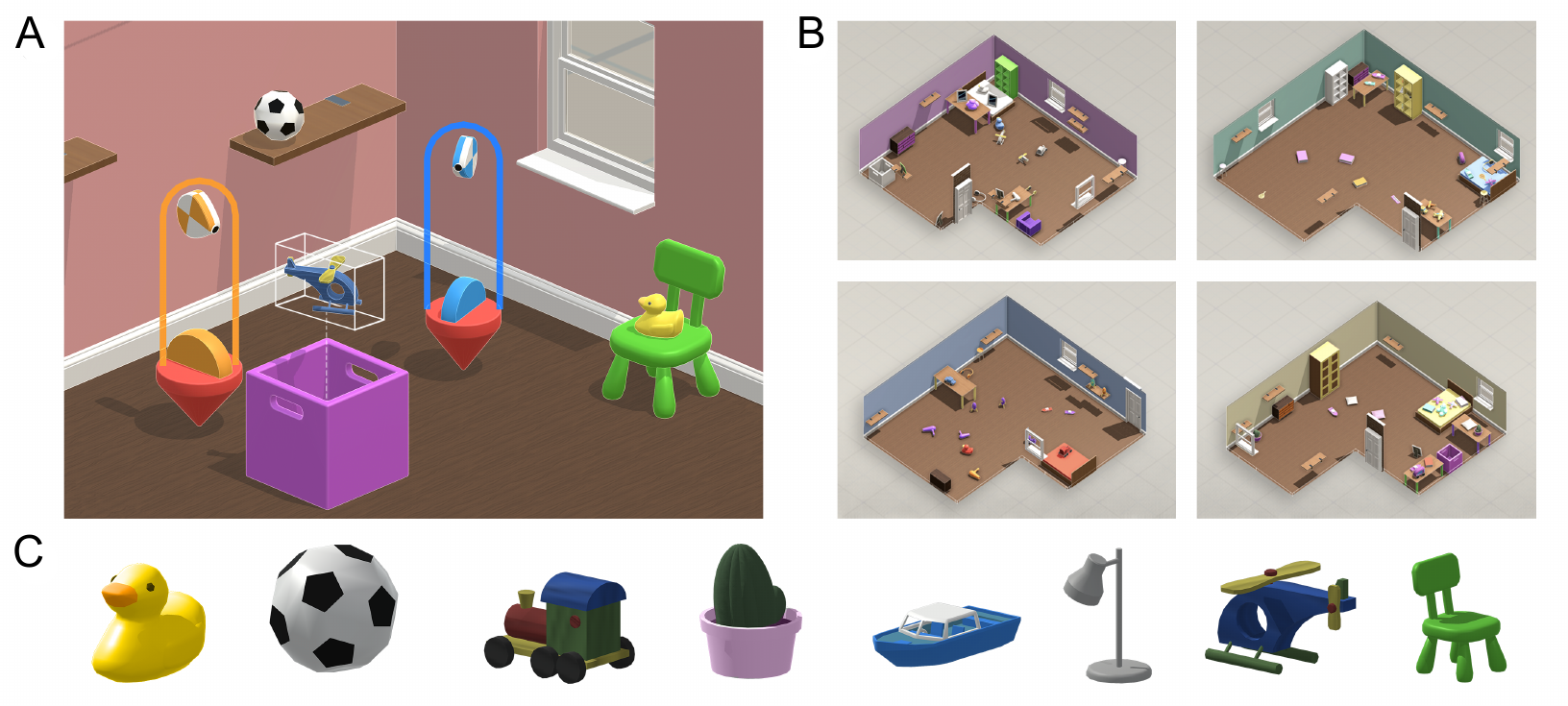}
    \caption{\textbf{The ``Playroom''.} The 3-D ``Playroom'' environment comprises a randomised set of rooms with children's toys and domestic objects, as well as containers, shelves, furniture, windows, and doors. The diversity of the environment enables interactions involving reasoning about space and object relations, ambiguity of references, containment, construction, support, occlusion, and partial observability. Agents interact with the world by moving around, manipulating objects, and speaking to each other. \textbf{A.} Depicts a simple interaction wherein the orange solver agent is placing a helicopter into a container while the blue setter agent watches on. \textbf{B.} Shows four random instantiations of the Playroom, each with a unique combination and arrangement of objects and furniture. \textbf{C.} A sampling of the types of objects available in the room.
    }
    \label{fig:playroom}
\end{figure}

The environment, which we call ``the Playroom,'' comprises a randomised set of rooms with children's toys and domestic objects (Figure \ref{fig:playroom}). The robotic embodiment by which the agent interacts with the world is a ``mobile manipulator'' -- that is, a robot that can move around and reposition objects. This environment supports a broad range of possible tasks, concepts, and interactions that are natural and intuitive to human users. It has containers, shelves, furniture, windows, and doors whose initial positions vary randomly each episode. There are diverse toys and objects that can be moved and positioned. The rooms are \emph{L}-shaped, creating blocked lines of sight, and have randomly variable dimensions. As a whole, the environment supports interactions that involve reasoning about space and object relations, ambiguity of references, containment, construction, support, occlusion, and partial observability. The language referring to this world can involve instructed goals, questions, or descriptions at different levels of specificity. Although the environment is simple compared to the real world, it affords rich and combinatorial interactions.

\subsection{Learning to Interact}
We aim to build agents that can naturally interact with and usefully assist humans. As a first step, one might consider optimising for this outcome directly. A critical prerequisite is a metric measuring ``useful'' interactions. Yet defining such a metric is a thorny issue because what comprises ``useful'' (or, simply, ``good'') is generally ambiguous and subjective. We need a way to measure and make progress without interminable Socratic debate about the meaning of ``good'' \citep{adam1902republic}. 

Suppose we do not have such an explicit rule-based metric to apply to any interaction. In principle, we can overcome the issue of the subjectivity of evaluation by embracing it: we can instead rely on a human evaluator's or collective of evaluators' judgements of the utility of interactions. This resolves the problem of codifying these value judgements \emph{a priori}. However, additional challenges remain. For the sake of argument, let's first suppose that an evaluator is only tasked with judging very unambiguous cases of success or failure. In such a scenario, the efficiency of improving an agent by issuing evaluative feedback depends critically on the intelligence of the agent being evaluated. Consider the two cases below:

If the agent is already intelligent (for example, it is another human), then we can expect the ratio of successes to failures to be moderately high. If the evaluator can unambiguously evaluate the behaviour, then their feedback can be informative. The mutual information between behaviour and evaluation is upper-bounded by the entropy in the evaluation\footnote{For any two random variables $B$ (e.g. a behavioural episode of actions taken by humans) and $Y$ (e.g. a binary evaluation), $\mathcal{I}[B;Y] = H[Y] - H[Y \mid B] \leq H[Y]$.}, and this mutual information can be used to provide feedback to the agent that discriminates between successes and failures. 

If, however, the agent is not already intelligent (for example, it is an untrained agent), then we can expect the ratio of successes to failures to be extremely low. In this case, almost all feedback is the \emph{same} and, consequently, uninformative; there is no measurable correlation between variations in agent behaviour and variations in the evaluation. As tasks increase in complexity and duration, this problem only becomes more severe. Agents must accidentally produce positive behaviour to begin to receive discriminative feedback. The number of required trials is inversely related to the probability that the agent produces a reasonable response on a given trial. For a success probability of $10^{-3}$, the agent needs approximately 1,000 trials before a human evaluator sees a successful trial and can provide feedback registering a change in the optimisation objective. The data required then grow linearly in the time between successful interactions.

Even if the agent fails almost always, it may be possible to compare different trials and to provide feedback about ``better'' and ``worse'' behaviours produced by an agent \citep{christiano2017deep}. While such a strategy can provide a gradient of improvement from untrained behaviour, it is still likely to suffer from the plateau phenomenon of indiscernible improvement in the early exploration stages of reinforcement learning \citep{kakade2003sample}. This will also dramatically increase the number of interactions for which evaluators need to provide feedback before the agent reaches a tolerable level of performance. 

Regardless of the actual preferences (or evaluation metric) of a human evaluator, fundamental properties of the reinforcement learning problem suggest that performance will remain substandard until the agent begins to learn how to behave well in exactly the same distribution of environment states that an intelligent expert (e.g., another human) is likely to visit. This fact is known as the \emph{performance difference lemma} \citep{kakade2003sample}. Formally, if $\pi^*(\mathbf{s})$ is the state distribution visited by the expert, $\pi^*(\mathbf{a} \mid \mathbf{s})$ is the action distribution of the expert, $V^{\pi}$ is the average value achieved by the agent $\pi$, and $Q^{\pi}(\mathbf{s}, \mathbf{a})$ is the value achieved in a state if action $\mathbf{a}$ is chosen, then the performance gap between the expert $\pi^*$ and the agent $\pi$ is
\begin{align*}
V^{\pi^*} - V^{\pi} & = \sum_{\mathbf{s}} \pi^*(\mathbf{s}) \sum_{\mathbf{a}} (\pi^*(\mathbf{a} \mid \mathbf{s}) - \pi(\mathbf{a} \mid \mathbf{s})) Q^{\pi}(\mathbf{s}, \mathbf{a}).  
\end{align*}
That is, as long as the expert is more likely to choose a good action (with larger $Q^\pi(\mathbf{s},\mathbf{a})$) in the states it likes to visit, there will be a large performance difference. 
Unfortunately, the non-expert agent has quite a long way to go before it can select those good actions, too. Because an agent training from scratch will visit a state distribution $\pi(\mathbf{s})$ that is substantially different from the expert's $\pi^*(\mathbf{s})$ (since the state distribution is itself a function of the policy), it is therefore unlikely to have learned how to pick good actions in the expert's favoured states, neither having visited them nor received feedback in them. The problem is vexed: to learn to perform well, the agent must often visit common expert states, but doing so is tantamount to performing well. Intuitively, this is the cause of the plateau phenomenon in RL. It poses a substantial challenge to ``human-in-the-loop'' methods of training agents by reward feedback, where the human time required to evaluate and provide feedback can be tedious, expensive, and can bottleneck the speed with which the AI can learn. The silver lining is that, while this theorem makes a serious problem apparent, it also points toward a resolution: if we can find a way to generally make $\pi(\mathbf{a} \mid \mathbf{s}) = \pi^*(\mathbf{a} \mid \mathbf{s})$, then the performance gap disappears.

In sum, while we could theoretically appeal to human judgement in lieu of an explicit metric to train agents to interact, it would be prohibitively inefficient and result in a substantial expenditure of human effort for little gain. For training by human evaluation to merit further consideration, we should first create agents whose responses to a human evaluator's instructions are satisfactory a larger fraction of the time. Ideally, the agent's responses are already very close to the responses of an intelligent, cooperative person who is trying to interact successfully. At this point, human evaluation has an an important role to play in adapting and improving the agent behaviour by goal-directed optimisation. Thus, before we collect and learn from human evaluations, we argue for building an intelligent behavioural \emph{prior} \citep{galashov2019information}: namely, a model that produces human-like responses in a variety of interactive contexts. 

Building a behavioural prior and demonstrating that humans judge it positively during interaction is the principal achievement of this work. We turn to imitation learning to achieve this, which directly leverages the information content of intelligent human behaviour to train a policy. 

\subsection{Collecting Data for Imitation Learning}
Imitation learning has been successfully deployed to build agents for self-driving cars \citep{pomerleau1989alvinn}, robotics and biomimetic motor control \citep{schaal1999imitation}, game play \citep{silver2016mastering, vinyals2019grandmaster}, and language modeling \citep{shannon1951prediction}. Imitation learning works best when humans are able to provide very good demonstrations of behaviour, and in large supply. For some domains, such as pure text natural language processing, large corpora exist that can be passively harvested from the internet \citep{brown2020language}. For other domains, more targeted data collection is currently required. Training agents by imitation learning in our domain requires us to devise a protocol for collecting human interaction data, and then to gather it at scale. The dataset we have assembled contains approximately two years of human-human interactions in real-time video and text. Measured crudely in hours (rather than in the number of words or the nature of utterances), it matches the duration of childhood required to attain oral fluency in language.

To build an intelligent behavioural prior for an agent acting in the Playroom, we could theoretically deploy imitation learning on free-form human interactions. Indeed, a small fraction of our data was collected this way. However, to produce a data distribution representing certain words, skills, concepts, and interaction types in desirable proportions, we developed a more controlled data collection methodology based on events called \emph{language games}.\footnote{Inspired by Wittgenstein's ideas about the utility of communication \citep{wittgenstein1953philosophische}.} 

\begin{figure}
    \centering
    \includegraphics[width=0.8\textwidth]{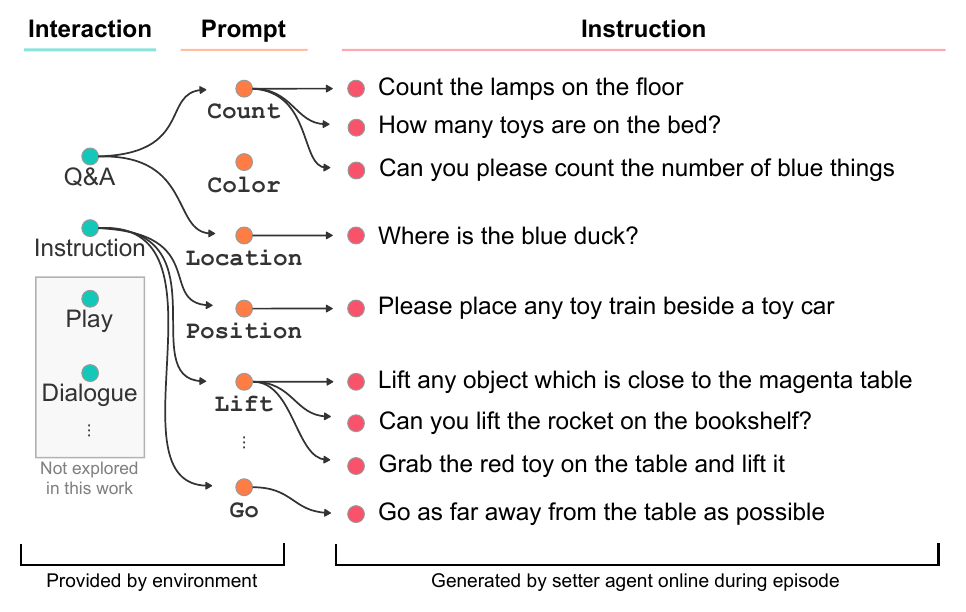}
    \caption{\textbf{Generating Diverse Interactions.} Interactions in the Playroom could take myriad forms. To encourage diverse interactions in the Playroom, we provided prompts (in orange) to humans which they expanded into specific language instructions (in red) for the other human or agent. Prompts shown here are short forms: e.g. {\tt Lift} corresponded to ``Ask the other player to lift something in the room,'' {\tt Color} corresponded to ``Ask the other player about the color of something in the room.'' 
    }
    \label{fig:interactions}
\end{figure}

We categorised the space of interactions into four basic types: question and answer (Q\&A), instruction-following, play, and dialogue (Figure~\ref{fig:interactions}). For this work, we have focused exclusively on the first two. Within each type, we framed several varieties of predefined \emph{prompts}. Prompts included, ``Ask the other player to bring you one or more objects,'' and, ``Ask the other player whether a particular thing exists in the room.'' We used 24 base prompts and up to 10 ``modifiers'' (e.g., ``Try to refer to objects by color'') that were appended to the base prompts to provide variation and encourage more specificity. One example of a prompt with a modifier was: ``Ask the other player to bring you one or more object. Try to refer to objects by color.''

Human participants were divided into two groups: \emph{setters} and \emph{solvers}. Setters received a prompt and were responsible for issuing an instruction based on it. Solvers were responsible for following instructions. Each episode in which a human setter was prompted to provide an instruction to a human solver is what we call a language game (Figure~\ref{fig:lang_games_sequence_diagram}). In each language game, a unique room was sampled from a generative model that produces random rooms, and a prompt was sampled from a list and shown to the setter. The human setter was then free to move around the room to investigate the space. When ready, the setter would then improvise an instruction based on the prompt they received and would communicate this instruction to the solver through a typed chat interface (Figure~\ref{fig:language_games_ui}). The setter and solver were given up to two minutes for each language game.

\begin{figure}
    \centering
    \includegraphics[width=\textwidth]{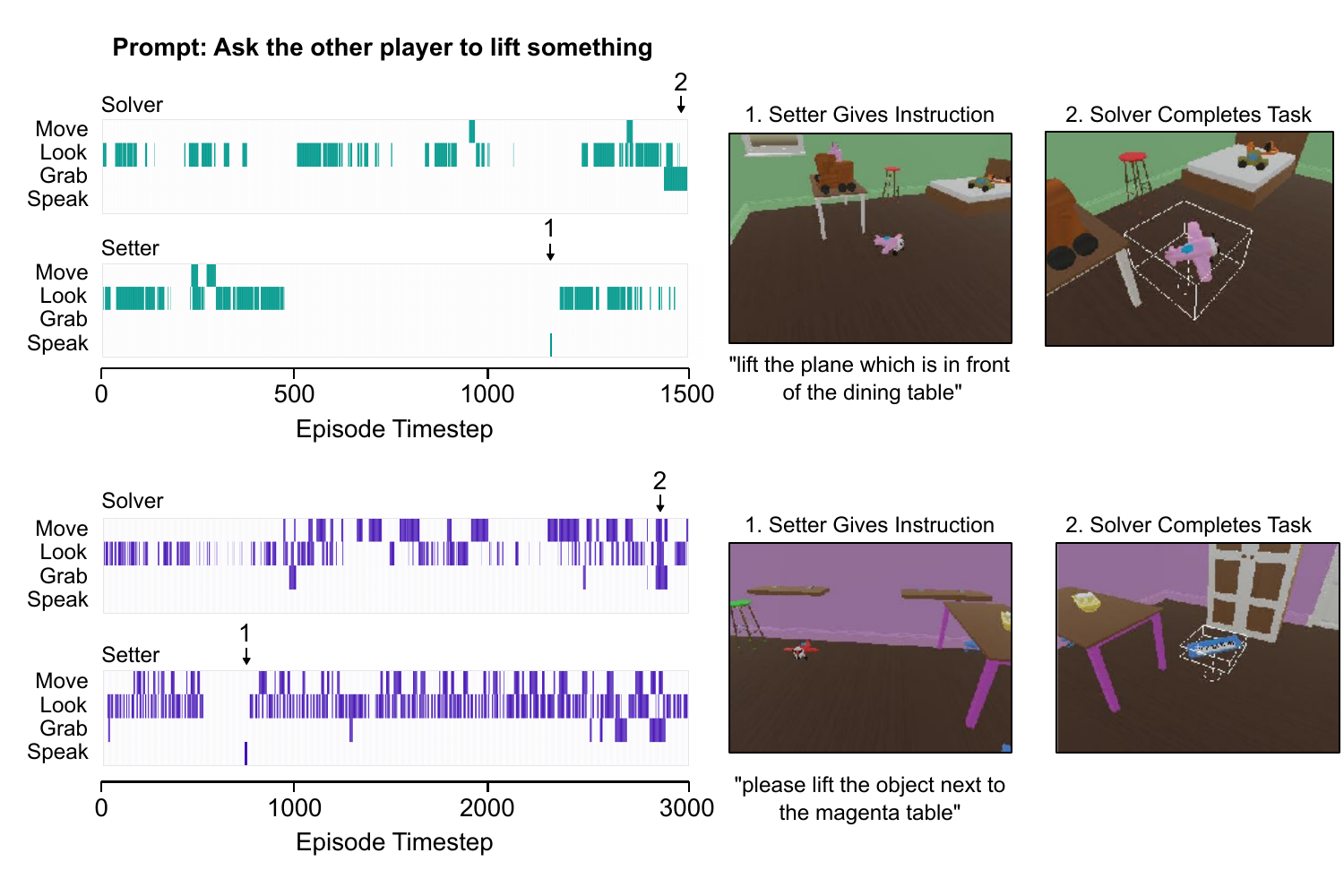}
    \caption{\textbf{Example Trajectories}. In these two human-human episodes, the setter was prompted to ask the solver to lift an object in the room. In the top example, the setter sets the task and the solver completes it in a straightforward manner. In the bottom example, there is some ambiguity: the setter was presumably referring to the red airplane on the ground, but the solver proceeded to lift the blue keyboard, which was also near the magenta table. The task was nevertheless completed successfully.
    }
    \label{fig:trajectory}
\end{figure}

The role of the setter was therefore primarily to explore and understand the situational context of the room (its layout and objects) and to initiate diverse language games constrained by the basic scaffolding given by the prompt (Figure~\ref{fig:interactions}). By defining a simple set of basic prompts, we could utilise humans' creative ability to conjure interesting, \emph{valid} instructions on-the-fly, with all the nuance and ambiguity that would be impossible to define programmatically. While the language game prompts constrained what the setters ought to instruct, setters and solvers were both free to use whatever language and vocabulary they liked. This further amplified the linguistic diversity of the dataset by introducing natural variations in phrasing and word choice. Consider one example, shown in the lower panel of Figure~\ref{fig:trajectory}: the setter looks at a red toy aeroplane, and, prompted to instruct the solver to lift something, asks the solver to ``please lift the object next to the magenta table,'' presumably referring to the aeroplane. The solver then moves to the magenta table and instead finds a blue keyboard, which it then lifts. This constituted a successful interaction even though the referential intention of the instruction was ambiguous.

Altogether, we collected 610,608 episodes of humans interacting as a setter-solver pair. From this total we allocated 549,468 episodes for training, and 61,140 for validation. Episodes lasted up to a maximum of 2 minutes (3,600 steps), with a mean and standard deviation of 55~$\pm$~25s (1,658~$\pm$~746 steps). The relative proportion of language games can be found in Table~\ref{tab:human_episodes_per_prompt} in the Appendix. Setters took 26~$\pm$~16s (784~$\pm$~504 steps) to pose a task for a solver, given the environment prompt (which was communicated at the start of an episode). In the 610,608 episodes there were 320,144 unique setter utterances, and 26,023 unique solver utterances, with an average length of 7.5~$\pm$~2.5 words and a maximum length of 29 words for setters. To put it another way, this signifies that there are 320,144 unique tasks instructed in the dataset. For solvers, the average length was 4.1~$\pm$~2.4 and a maximum length of 26. Upon receiving a setter instruction, the time solvers took to complete the task was 28~$\pm$~18s (859~$\pm$~549 steps). Figure \ref{fig:actions} depicts the average action composition for a solver in an episode. Notably, the density of actions was low, and when actions were taken, the distribution of action choice was highly skewed. This was even more pronounced for language emissions (Figure~\ref{fig:word_freq}A), where approximately one utterance was made per episode for setters, with word choices following a long-tailed distribution for a vocabulary of approximately 550 words. 

\subsection{Agent Architecture}

\subsubsection{Action Representation}
\label{sec:action_space}
Our agents control the virtual robot in much the same way as the human players. The action space is multidimensional and contains a continuous 2D mouse \emph{look} action. The agent space also includes several keyboard buttons, including \emph{forward, left, backward, right} (corresponding to keys `WASD'), along with mixtures of these keys (Figure 3). Finally, a \emph{grab} action allows the agent to grab or drop an object. The full details of the observation and action spaces are given in Appendix~\ref{appendix_agent_outputs}.

\begin{figure}
    \centering
    \includegraphics[width=\textwidth]{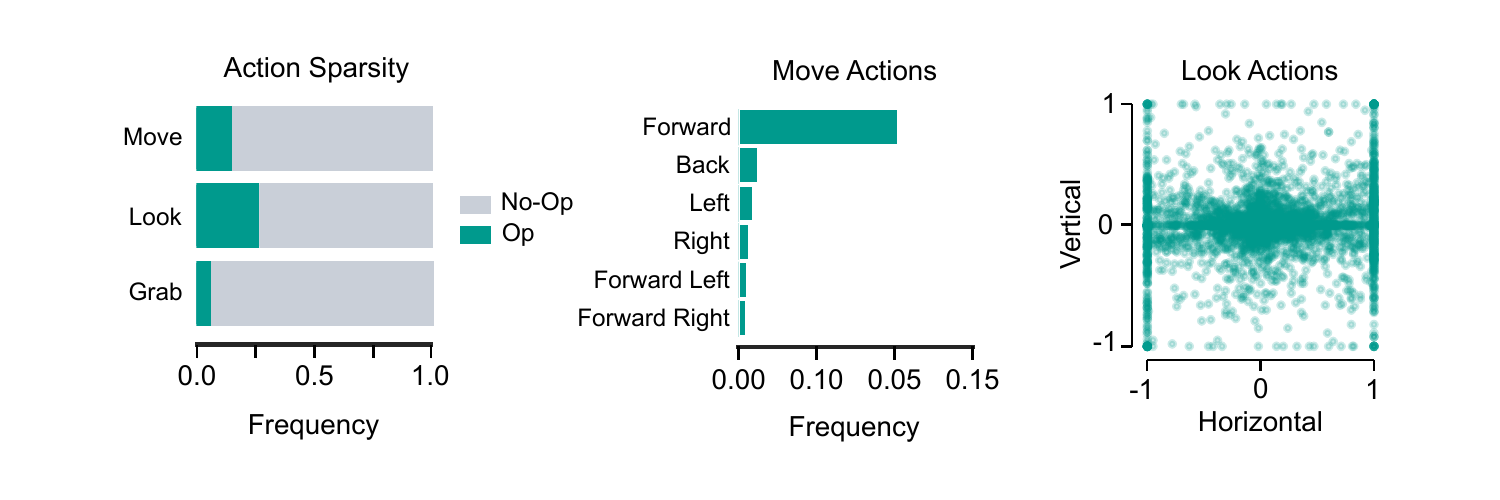}
    \caption{\textbf{Action Composition.} Across each of the move, look, and grab actions we observed a skewed distribution with respect to the chosen actions (middle, right), and whether an action or no-op is chosen (left). For the move action, ``forward'' is heavily represented, whilst look actions are clustered mainly around the origin (corresponding to small shifts in gaze direction), and along the borders (corresponding to large rotations). Each action is relatively rare in the entire trajectory, as seen by the proportion of no-ops to ops.}
    \label{fig:actions}
\end{figure}

The agent operates in discrete time and produces 15 actions per second. These actions are produced by a stochastic \emph{policy}, a probability distribution, $\pi$, defined jointly over all the action variables produced in one time step, $\mathbf{a}$: $\pi(\mathbf{a}) = \pi(\emph{look}, \emph{key}, \emph{grab})$ (At times, we may use the words \emph{agent} and \emph{policy} interchangeably, but when we mean to indicate the conditional distribution of actions given observations, we will refer to this as the policy exclusively.) In detail, we include \emph{no-operation} (``no-op'') actions to simplify the production of a null mouse movement or key press. Although we have in part based our introductory discussion on the formalism of fully-observed Markov Decision Processes, we actually specify our interaction problem more generally. At any time $t$ in an episode, the policy distribution is conditioned on the preceding perceptual observations, which we denote $\mathbf{o}_{\leq t} \equiv (\mathbf{o}_{0}, \mathbf{o}_{1}, \dots, \mathbf{o}_{t})$. The policy is additionally \emph{autoregressive}. That is, the agent samples one action component first, then conditions the distribution over the second action component on the choice of the first, and so on. If we denote the choice of the \emph{look} \emph{no-op} action at time $t$ as $\mathbf{a}_t^{(0)}$, the choice of the \emph{look} action as $\mathbf{a}_t^{(1)}$, the choice of the \emph{key} \emph{no-op} as $\mathbf{a}_t^{(2)}$, the choice of the \emph{key} as $\mathbf{a}_t^{(3)}$, and so on, the action distribution is jointly expressed as:
\begin{align*}
\pi_{\vtheta}(\mathbf{a}_t \mid \mathbf{o}_{\leq t}) & = \prod_{k=0}^{K} \pi_{\vtheta}(\mathbf{a}_t^{(k)} \mid \mathbf{o}_{\leq t}, \mathbf{a}_t^{(< k)}),
\end{align*}
where $\vtheta$ are the parameters of the neural network used to define the policy. The mouse look action distribution is in turn also defined autoregressively: the first sampled action splits the window bounded by $(-1, 1) \times (-1, 1)$ in width and height into 9 squares. The second action splits the selected square into 9 further squares, and so on. Repeating this process several times allows the agent to express any continuous mouse movement up to a threshold resolution. 

\subsubsection{Perception and Language}
Agents perceive the environment visually using ``RGB'' pixel input at resolution of $96 \times 72$. When an object can be grasped by the manipulator, a bounding box outlines the object (Figures \ref{fig:playroom}, \ref{fig:trajectory}, \& \ref{fig:actions}). Agents also process text inputs coming from either another player (including humans), from the environment (agents that imitate the setter role must process the language game prompt), or from their own language output at the previous time step. Language input is buffered so that all past tokens up to a buffer length are observed at once. We will denote the different modalities of vision, language input arriving from the language game prompt, language input coming from the other agent, and language input coming from the agent itself at the last time step as $\mathbf{o}^\textsc{V}$, $\mathbf{o}^\textsc{LP}$, and $\mathbf{o}^\textsc{LO}$, and $\mathbf{o}^\textsc{LS}$, respectively.

Language output is sampled one token at a time, with this step performed after the autoregressive movement actions have been chosen. The language output token is observed by the agent at the next time step. We process and produce language at the level of whole words, using a vocabulary consisting of the approximately 550 most common words in the human data distribution (Section~\ref{sec:vocab}) and used an `UNK' token for the rest.

\subsubsection{Network Components}
\begin{figure}
    \centering
    \includegraphics[width=\textwidth]{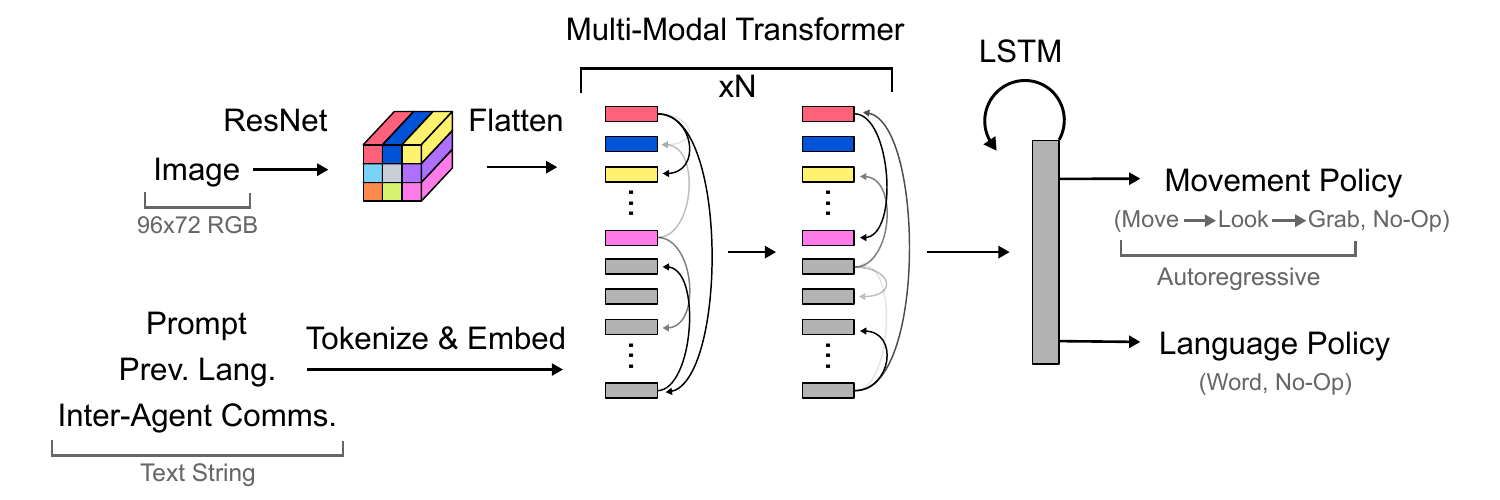}
    \caption{\textbf{Agent Architecture.} The agent receives both RGB images and text strings as inputs. The former gets encoded through a ResNet, and the latter are tokenized by word using a custom vocabulary, and subsequently embedded as distributed vectors. Together the ResNet ``hyper-pixels'' and tokenized words comprise a set of vectors that is the input to a multi-modal transformer. The transformer's output provides the input to an LSTM, which in turn provides input to the motor and language policies.}
    \label{fig:architecture}
\end{figure}

The agent architecture (Figure~\ref{fig:architecture}) uses a ResNet \citep{he2016deep} for vision. At the highest level of the ResNet, a spatial map of dimensions $(\emph{width} \times \emph{height} \times \emph{number-of-channels})$ is produced. The vectors from all the $\emph{width} \times \emph{height}$ positions in this spatial array are concatenated with the embeddings of the language input tokens, which include words comprising the inter-agent communication, the prompt delivered from the environment (to the setter only), and previous language emissions. These concatenated vectors are jointly processed by a transformer network \citep{vaswani2017attention}, which we refer to as the multi-modal transformer (MMT). The output of the MMT consists of a mean-pooling across all output embeddings, concatenated with dedicated output embeddings that function much like the ``CLS'' embedding in the BERT model \citep{devlin2018bert} (see Section~\ref{appendix_mmt} in the Appendix for more information). This output provides the input to an LSTM memory, which in turn provides the input to smaller networks that parameterise the aforementioned policies.

\subsection{Learning}

Our approach to training interactive agents combines diverse techniques from imitation learning with additional supervised and unsupervised learning objectives to regularise representations. We first explain the basic principles behind each method, then explain how they are brought together. 

\subsubsection{Behavioural Cloning}
The most direct approach to imitation learning, known as \emph{behavioural cloning} (BC) \citep{pomerleau1989alvinn, osa2018algorithmic}, frames the problem of copying behaviour as a supervised sequence prediction problem \citep{graves2013generating}. Recalling the discussion of the performance difference lemma, behavioural cloning is an approach that tries to make $\pi(\mathbf{a} \mid \mathbf{s}) = \pi^*(\mathbf{a} \mid \mathbf{s})$, or, in our case, $\pi(\mathbf{a}_t \mid \mathbf{o}_{\leq t}) = \pi^*(\mathbf{a}_t \mid \mathbf{o}_{\leq t})$. It requires a dataset of observation and action sequences produced by expert demonstrators.

A temporal observation sequence $\mathbf{o}_{\leq T} \equiv (\mathbf{o}_0, \mathbf{o}_1, \mathbf{o}_2, \dots, \mathbf{o}_T)$ and a temporal action sequence $\mathbf{a}_{\leq T} \equiv (\mathbf{a}_0, \mathbf{a}_1, \mathbf{a}_2, \dots, \mathbf{a}_T)$ together comprise a \emph{trajectory}. (Length, or \emph{trajectory length}, refers to the number of elements in the observation or action sequence, and while trajectory lengths can vary, for simplicity we develop the fixed length case.) The dataset is distributed according to some unknown distribution $\pi^{*}(\mathbf{o}_{\leq T}, \mathbf{a}_{\leq T})$. For language games, we constructed separate datasets of setter trajectories and solver trajectories.
The loss function for behavioural cloning is the (forward) Kullback-Leibler divergence between $\pi^{*}$ and $\pi_{\vtheta}$:
\begin{align*}
\mathcal{L}^\textsc{bc}(\vtheta) & = \KLpq{\pi^{*}}{\pi_{\vtheta}} \\
& = \expect{\pi^{*}(\mathbf{o}_{\leq T}, \mathbf{a}_{\leq T})}{\ln \frac{\pi^{*}(\mathbf{o}_{\leq T}, \mathbf{a}_{\leq T})}{\pi_{\vtheta}(\mathbf{o}_{\leq T}, \mathbf{a}_{\leq T})}} \\
& = \const(\vtheta) - \expect{\pi^{*}(\mathbf{o}_{\leq T}, \mathbf{a}_{\leq T})}{\ln \pi_{\vtheta}(\mathbf{o}_{\leq T}, \mathbf{a}_{\leq T})}, 
\end{align*}
where $\const(\vtheta)$ collects the demonstrator distribution entropy term, which is a constant independent of the policy parameters. The policy trajectory distribution $\pi_{\vtheta}(\mathbf{o}_{\leq T}, \mathbf{a}_{\leq T})$ is a product of conditional distributions from each time step. The product alternates between terms that are a function of the policy directly, $\pi_{\vtheta}(\mathbf{a}_t \mid \mathbf{o}_{\leq t}, \mathbf{a}_{<t})$, and terms that are a function of the environment and independent of the policy parameters, $p^\textsc{env}(\mathbf{o}_t \mid \mathbf{o}_{<t}, \mathbf{a}_{<t})$. The product is $ \pi_{\vtheta}(\mathbf{o}_{\leq T}, \mathbf{a}_{\leq T}) = \prod_{t=0}^T p^\textsc{env}(\mathbf{o}_t \mid \mathbf{o}_{<t},  \mathbf{a}_{<t}) \pi_{\vtheta}(\mathbf{a}_t \mid \mathbf{o}_{\leq t}, \mathbf{a}_{<t})$. Ignoring constants with respect to the parameters, the argument of the logarithm can therefore be further broken down by time step:
\begin{align*}
\mathcal{L}^\textsc{bc}(\vtheta) & = -\expect{\pi^{*}(\mathbf{o}_{\leq T}, \mathbf{a}_{\leq T})}{\ln \prod_{t=0}^T p^\textsc{env}(\mathbf{o}_t \mid \mathbf{o}_{<t}, \mathbf{a}_{<t}) \pi_{\vtheta}(\mathbf{a}_t \mid \mathbf{o}_{\leq t}, \mathbf{a}_{<t})}  \\
& = -\expect{\pi^{*}(\mathbf{o}_{\leq T}, \mathbf{a}_{\leq T})}{\sum_{t=0}^T \ln p^\textsc{env}(\mathbf{o}_t \mid \mathbf{o}_{<t}, \mathbf{a}_{<t}) + \ln \pi_{\vtheta}(\mathbf{a}_t \mid \mathbf{o}_{\leq t}, \mathbf{a}_{<t})} \\
& = \const(\vtheta) - \expect{\pi^{*}(\mathbf{o}_{\leq T}, \mathbf{a}_{\leq T})}{\sum_{t=0}^T \ln \pi_{\vtheta}(\mathbf{a}_t \mid \mathbf{o}_{\leq t}, \mathbf{a}_{<t})}.
\end{align*}
We have optionally decided to drop explicit conditioning of the policy on past actions, except insofar as they influence the observations, giving 
\begin{align}
\mathcal{L}^\textsc{bc}(\vtheta) & = - \expect{\pi^{*}(\mathbf{o}_{\leq T}, \mathbf{a}_{\leq T})}{\sum_{t=0}^T \ln \pi_{\vtheta}(\mathbf{a}_t \mid \mathbf{o}_{\leq t})}. \label{eq:loss_bc}
\end{align}
We can observe that the expectation is under the demonstration distribution. In practice, we train on the empirical distribution of trajectories in the demonstration dataset. In each evaluation of the loss function, we sample a batch of $B$ trajectories from the dataset:
\begin{align*}
\mathcal{L}^\textsc{bc}(\vtheta) & = -\frac{1}{B} \sum_{n=1}^{B} \sum_{t=0}^{T} \ln \pi_{\vtheta}(\mathbf{a}_{n,t} \mid \mathbf{o}_{n,\leq t}).
\end{align*} 
Although demonstrators interact in the environment to provide data, with BC the agent exclusively learns without acting at all. This feature of BC can be considered an advantage or a disadvantage: an advantage because the agent need not perform trial and error in the world to learn, and a disadvantage because it cannot utilise self-directed environment interaction to learn more. Despite this problem, behavioural cloning is still a principled and reliable algorithm. It performs best when datasets are large, and the policy distribution is able to represent complex correlations among components of the action -- hence our choice of autoregressive action distributions. However, behavioural cloning can be improved, as we will show.

\subsubsection{Auxiliary Learning and Regularisation}
Behavioural cloning, like other supervised learning methods that learn a map from inputs to outputs, can benefit from regularisation. When the agent (policy) acts in the environment, it will encounter observation sequences that are novel. This is an inevitability due to the high dimensionality of the perceptual inputs and the combinatorics of the room and of language itself. But it is more than a statement about combinatorics and dimensionality: when the agent acts it directly alters the state of the world and its own reafferent observations. And, when the policy distribution is conditioned on an observation sequence that is distinct from the training data, $\pi_{\vtheta}(\mathbf{a}_t \mid \mathbf{o}_{\textsc{unseen},\leq t})$, the desired response is nominally undefined and must be inferred by appropriate generalisation. 

In the Playroom (or indeed, in any human-compatible environment), we know that pixels are grouped into higher-order structures that we perceive as toys, furniture, the background, etc. These higher-order structures are multi-scale and include the even higher-order spatial relationships among the objects and features in the room. Together, these perceptual structures influence human behaviour in the room. Our regularisation procedures aim to reduce the number of degrees of freedom in the input data source and the network representations, while preserving information that is correlated with attested human behaviour. These regularisation procedures produce representations that effectively reduce the discriminability of some pairs of observation sequences $(\mathbf{o}_{i,\leq t}, \mathbf{o}_{j, \leq t})$ while increasing the discriminability of others. The geometry of these representations then shapes how the policy network infers its responses, and how it generalises to unseen observations.

We use two kinds of regularisation, both of which help to produce visual representations that improve BC agents with respect to our evaluation metrics. The first regularisation, which we call \emph{Language Matching} (LM), is closely related to the Contrastive Predictive Coding algorithm \citep{oord2018representation, henaff2019data} and Noise Contrastive Estimation \citep{gutmann2010noise} and helps produce visual representations reflecting linguistic concepts. A classifier $D_{\vtheta}$ is attached to the agent network and provided input primarily from the mean-pooling vector of the MMT. It is trained to determine if the visual input and the solver language input (i.e., the instruction provided by the setter) come from the same episode or different episodes (see Appendix section~\ref{appendix_mmt}):
\begin{align}\label{eq:langmatch}
\mathcal{L}^\textsc{lm}(\vtheta) & = 
-\frac{1}{B} \sum_{n=1}^{B} \sum_{t=0}^T  \bigg [ \ln D_{\vtheta}(\mathbf{o}_{n,t}^\textsc{V}, \mathbf{o}_{n,t}^\textsc{LO}) + \ln \big( 1 - D_{\vtheta}(\mathbf{o}_{n,t}^\textsc{V}, \mathbf{o}_{\textsc{Shift}(n),t}^\textsc{LO}) \big ) \bigg ],
\end{align}
where $B$ is the batch size and $\textsc{Shift}(n)$ is the $n$-th index after a modular shift of the integers: $1 \to 2, 2 \to 3 \dots, B \to 1$. The loss is ``contrastive'' because the classifier must distinguish between real episodes and decoys. To improve the classifier loss, the visual encoder must produce representations with high mutual information to the encoded language input. We apply this loss to data from human solver demonstration trajectories where there is often strong alignment between the instructed language and the visual representation: for example, ``Lift a red robot'' predicts that there is likely to be a red object at the centre of fixation, and ``Put three balls in a row'' predicts that three spheres will intersect a ray through the image.

The second regularisation, which we call the ``Object-in-View'' loss (OV), is designed very straightforwardly to produce visual representations encoding the objects and their colours in the frame. We build a second classifier to contrast between strings describing coloured objects in frame versus fictitious objects that are not in frame. To do this, we use information about visible objects derived directly from the environment simulator, although equivalent results could likely be obtainable by conventional human segmentation and labeling of images \citep{girshick2015fast, he2017mask}. Notably, this information is only present during training, and not at inference time.

Together, we refer to these regularising objective functions as ``auxiliary losses.''

\subsubsection{Inverse Reinforcement Learning}
In the Markov Decision Process formalism, we can write the behavioural cloning objective another way to examine the sense in which it tries to make the agent imitate the demonstrator: 
\begin{align*}
\mathcal{L}^\textsc{bc}({\vtheta}) & = \expect{\pi^*(\mathbf{s})}{\KLpq{\pi^*(\mathbf{a} \mid \mathbf{s})}{\pi_{\vtheta}(\mathbf{a} \mid \mathbf{s})}}.
\end{align*}
The imitator learns to match the demonstrator's policy distribution over actions in the observation sequences generated by the demonstrator. Theoretical analysis of behavioural cloning \citep{ross2011reduction} suggests that errors of the imitator agent in predicting the demonstrator's actions lead to a performance gap that compounds.\footnote{Under relatively weak assumptions (bounded task rewards per time step), the suboptimality for BC is linear in the action prediction error rate $\epsilon$ but up to quadratic in the length of the episode $T$, giving $\mathcal{O}(\epsilon T^2)$. The performance difference would be linear in the episode length, $\mathcal{O}(\epsilon T)$, if each mistake of the imitator incurred a loss only at that time step; quadratic suboptimality means roughly that an error exacts a toll for each subsequent step in the episode.} The root problem is that each mistake of the imitator changes the distribution of future states so that $\pi_{\vtheta}(\mathbf{s})$ differs from $\pi^*(\mathbf{s})$. The states the imitator reaches may not be the ones in which it has been trained to respond. Thus, a BC-trained policy can ``run off the rails,'' reaching states it is not able to recover from. Imitation learning algorithms that also learn along the imitator's trajectory distribution can reduce this suboptimality \citep{ross2011reduction}. 

The regularisation schemes presented in the last section can improve the generalisation properties of BC policies to novel inputs, but they cannot train the policy to exert active control in the environment to attain states that are probable in the demonstrator's distribution. By contrast, \emph{inverse reinforcement learning} (IRL) algorithms \citep{ziebart2010modeling, finn2016connection} attempt to infer the reward function underlying the intentions of the demonstrator (e.g., which states it prefers), and optimise the policy itself using reinforcement learning to pursue this reward function. IRL can avoid this failure mode of BC and train a policy to ``get back on the rails'' (i.e., return to states likely in the demonstrator's state distribution; see previous discussion on the performance difference lemma). For an instructive example, consider using inverse reinforcement learning to imitate a very talented Go player. If the reward function that is being inferred is constrained to observe only the win state at the end of the game, then the estimated function will encode that winning is what the demonstrator does. Optimising the imitator policy with this reward function can then recover more information about playing Go well than was contained in the dataset of games played by the demonstrator alone. Whereas a behavioural cloning policy might find itself in a losing situation with no counterpart in its training set, an inverse reinforcement learning algorithm can use trial and error to acquire knowledge about how to achieve win states from unseen conditions.

Generative Adversarial Imitation Learning (GAIL) \citep{ho2016generative} is an algorithm closely related to IRL \citep{ziebart2010modeling, finn2016connection}. Its objective trains the policy to make the distribution $\pi_{\vtheta}(\mathbf{s}, \mathbf{a})$ match $\pi^*(\mathbf{s}, \mathbf{a})$. To do so, GAIL constructs a surrogate model, the \emph{discriminator}, which serves as a reward function. The discriminator, $D_{\vphi}$, is trained using conventional cross entropy to judge if a state and action pair is sampled from a demonstrator or imitator trajectory: 
\begin{align*}
\mathcal{L}^\textsc{disc}({\vphi}) & = -\expect{\pi^*(\mathbf{s}, \mathbf{a})}{\ln D_{\vphi}(\mathbf{s}, \mathbf{a})} - \expect{\pi_{\vtheta}(\mathbf{s}, \mathbf{a})}{\ln (1 - D_{\vphi}(\mathbf{s}, \mathbf{a}))}.
\end{align*}
The optimal discriminator, according to this objective, satisfies $D_{\vphi}(\mathbf{s}, \mathbf{a}) = \frac{\pi^*(\mathbf{s}, \mathbf{a})}{\pi^*(\mathbf{s}, \mathbf{a}) + \pi_{\vtheta}(\mathbf{s}, \mathbf{a})}$.\footnote{As was noted in \cite{goodfellow2014generative} and as is possible to derive by directly computing the stationary point with respect to $D_{\vphi}(\mathbf{s}, \mathbf{a})$: $\pi^*(\mathbf{s}, \mathbf{a}) / D_{\vphi}(\mathbf{s}, \mathbf{a}) - \pi_{\vtheta}(\mathbf{s}, \mathbf{a}) / (1 - D_{\vphi}(\mathbf{s}, \mathbf{a})) = 0$, etc.} We have been deliberately careless about defining $\pi(\mathbf{s}, \mathbf{a})$ precisely but rectify this now. In the discounted case, it can be defined as the discounted summed probability of being in a state and producing an action: $\pi(\mathbf{s}, \mathbf{a}) \equiv (1 - \gamma) \sum_t \gamma^t p(\mathbf{s}_t=\mathbf{s} \mid \pi) \pi(\mathbf{a} \mid \mathbf{s})$. The objective of the policy is to minimise the classification accuracy of the discriminator, which, intuitively, should make the two distributions as indiscriminable as possible: i.e., the same. Therefore, the policy should maximise
\begin{align*}
\mathcal{J}^\textsc{gail}({\vtheta}) & = -\expect{\pi_{\vtheta}(\mathbf{s}, \mathbf{a})}{\ln (1 - D_{\vphi}(\mathbf{s}, \mathbf{a}))}.
\end{align*}
This is exactly a reinforcement learning objective with per time step reward function $r(\mathbf{s}, \mathbf{a}) = - \ln (1 - D_{\vphi}(\mathbf{s}, \mathbf{a}))$. It trains the policy during interaction with the environment: the expectation is under the imitator policy's distribution, not the demonstrator's. Plugging in the optimal discriminator on the right-hand side, we have 
\begin{align*}
\mathcal{J}^\textsc{gail}({\vtheta}) & \approx -\expect{\pi_{\vtheta}(\mathbf{s}, \mathbf{a})}{\ln \frac{\pi_{\vtheta}(\mathbf{s}, \mathbf{a})}{\pi^*(\mathbf{s}, \mathbf{a}) + \pi_{\vtheta}(\mathbf{s}, \mathbf{a})}}.
\end{align*}
At the saddle point, optimised both with respect to the discriminator and with respect to the policy, one can show that $\pi_{\vtheta}(\mathbf{s}, \mathbf{a}) = \pi^*(\mathbf{s}, \mathbf{a})$.\footnote{Solving the constrained optimisation problem $\mathcal{J}^\textsc{gail}({\vtheta}) + \lambda [\sum_{\mathbf{a}} \pi_{\vtheta}(\mathbf{s}, \mathbf{a}) - 1]$ shows that $\frac{\pi_{\vtheta}(\mathbf{s}, \mathbf{a})}{\pi^*(\mathbf{s}, \mathbf{a}) + \pi_{\vtheta}(\mathbf{s}, \mathbf{a})} = \const$ for all $\mathbf{s}, \mathbf{a}$. Therefore, $\pi_{\vtheta}(\mathbf{s}, \mathbf{a}) = \pi^*(\mathbf{s}, \mathbf{a})$.} GAIL differs from traditional IRL algorithms, however, because the reward function it estimates is non-stationary: it changes as the imitator policy changes since it represents information about the probability of a trajectory in the demonstrator data compared to the current policy. 

GAIL provides flexibility. Instead of matching $\pi_{\vtheta}(\mathbf{s}, \mathbf{a}) = \pi^*(\mathbf{s}, \mathbf{a})$, one can instead attempt to enforce only that $\pi_{\vtheta}(\mathbf{s}) = \pi^*(\mathbf{s})$ \citep{merel2017learning, ghasemipour2020divergence}. We have taken this approach both to simplify the model inputs, and because it is sufficient for our needs: behavioural cloning can be used to imitate the policy conditional distribution $\pi^*(\mathbf{a} \mid \mathbf{s})$, while GAIL can be used to imitate the distribution over states themselves $\pi^*(\mathbf{s})$. In this case the correct objective functions are:  
\begin{align*}
\mathcal{L}^\textsc{disc}({\vphi}) & = -\expect{\pi^*(\mathbf{s})}{\ln D_{\vphi}(\mathbf{s})} - \expect{\pi_{\vtheta}(\mathbf{s})}{\ln (1 - D_{\vphi}(\mathbf{s}))}, \\
\mathcal{J}^\textsc{gail}({\vtheta}) & = -\expect{\pi_{\vtheta}(\mathbf{s}, \mathbf{a})}{\ln (1 - D_{\vphi}(\mathbf{s}))}.
\end{align*}
In practice, returning to our Playroom setting with partial observability and two agents interacting, we cannot assume knowledge of a state $\mathbf{s}_t$. Instead, we supply the discriminator with observation sequences $\mathbf{s}_t \approx (\mathbf{o}_{t-s k}, \mathbf{o}_{t-s(k-1)}, \dots, \mathbf{o}_{t})$ of fixed length $k$ and stride $s$; the policy is still conditioned as in Equation \ref{eq:loss_bc}. 

\begin{figure}
    \centering
    \includegraphics[width=\textwidth]{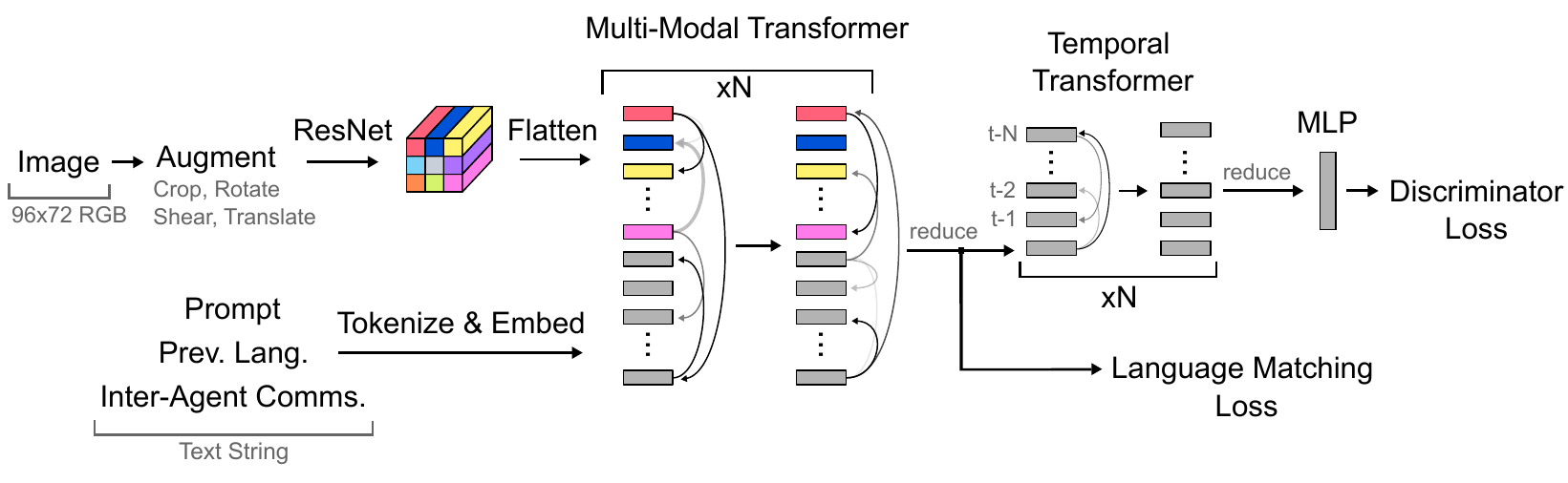}
    \caption{\textbf{GAIL Discriminator Architecture:} The discriminator receives the same inputs as the agent, RGB images and text strings, and encodes them with similar encoders (ResNet, text embedder, and Multi-Modal Transformer) into a single summary vector. The encoded inputs are then processed by a Temporal Transformer that has access to the summary vectors from previous time steps. The mean-pooled output of this transformer is then passed through an MLP to obtain a single output representing the probability that the observation sequence is part of a demonstrator trajectory. The encoders are simultaneously trained by the auxiliary Language Matching objective.}
    \label{fig:rm_architecture}
    
\end{figure}

These observation sequences are short movies with language and vision and are consequently high-dimensional. We are not aware of extant work that has applied GAIL to observations this high-dimensional (see \cite{li2017infogail, zolna2019task} for applications of GAIL to simpler but still visual input), and, perhaps, for good reason. The discriminator classifier must represent the relative probability of a demonstrator trajectory compared to an imitator trajectory, but with high-dimensional input there are many undesirable classification boundaries the discriminator can draw. It can use capacity to over-fit spurious coincidences: e.g., it can memorise that in one demonstrator interaction a pixel patch was hexadecimal colour \#ffb3b3, etc., while ignoring the interaction's semantic content. Consequently, regularisation, as we motivated in the behavioural cloning context, is equally important for making the GAIL discriminator limit its classification to human-interpretable events, thereby giving reward to the policy if it acts in ways that humans also think are descriptive and relevant. For the GAIL discriminator, we use a popular data augmentation technique \emph{RandAugment} \citep{cubuk2020randaugment} designed to make computer vision more invariant. This technique stochastically perturbs each image that is sent to the visual ResNet. We use random cropping, rotation, translation, and shearing of the images. These perturbations substantially alter the pixel-level visual input without altering human understanding of the content of the images or the desired outputs for the network to produce. At the same time, we use the same language matching objective we introduced in the behavioural cloning section, which extracts representations that align between vision and language. This objective is active only when the input to the model is demonstrator observation sequence data, not when the imitator is producing data.

The architecture of the discriminator is shown in Figure \ref{fig:rm_architecture}. RandAugment is applied to the images, and a ResNet processes frames, converting them into a spatial array of vector embeddings. The language is also similarly embedded, and both are passed through a multi-modal transformer. No parameters are shared between the reward model and policy. The top of the MMT applies a mean-pooling operation to arrive at a single embedding per time step, and the language matching loss is computed based on this averaged vector. Subsequently, a second transformer processes the vectors that were produced across time steps before mean-pooling again and applying a multi-layer perceptron classifier representing the discriminator output.

\begin{figure}[h]
    \centering
    \includegraphics[width=0.75\textwidth]{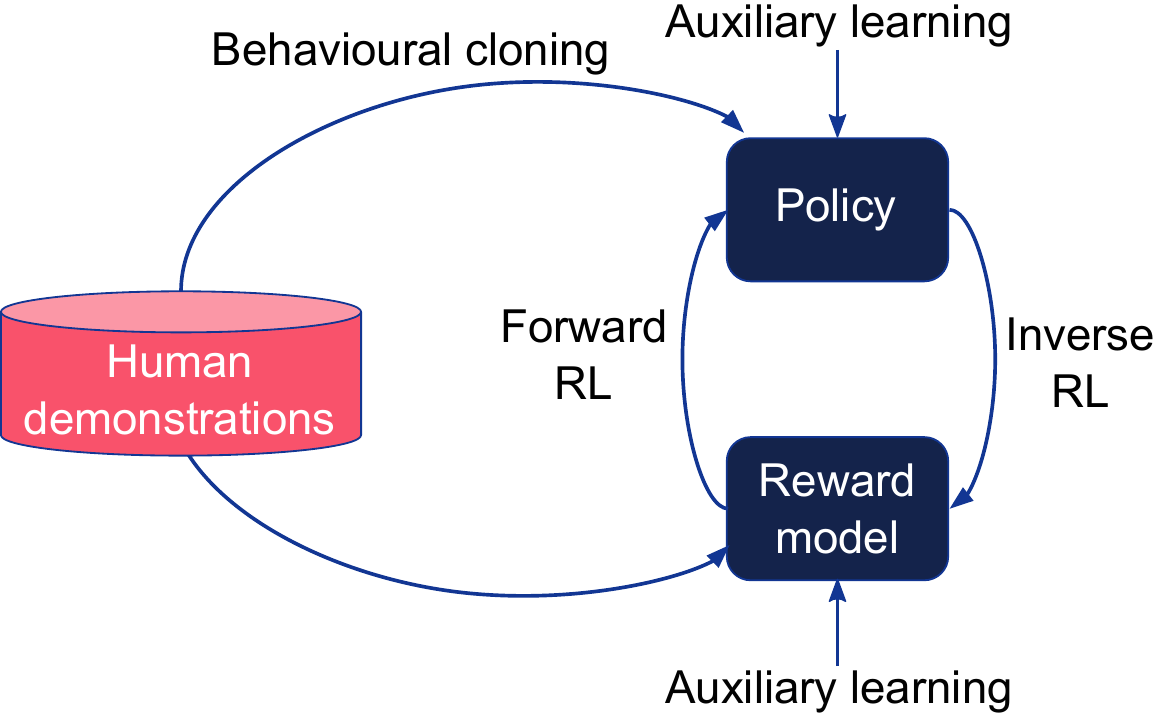}
    \caption{\textbf{Training schematic.} We train policies using human demonstrations via a mixture of behavioural cloning and reinforcement learning on a learned discriminator reward model. The reward model is trained to discriminate between human demonstrations (positive examples) and agent trajectories (negative examples). Both the policy and the reward model are regularised by auxiliary objectives.}
    \label{fig:training}
\end{figure}

Figure \ref{fig:training} summarises how we train agents. We gather human demonstrations of interactive language games. These trajectories are used to fit policies by behavioural cloning. We additionally use a variant of the GAIL algorithm to train a discriminator reward model, classifying trajectories as generated by either the humans or a policy. Simultaneously, the policy derives reward if the discriminator classifies its trajectory as likely to be human. Both the policy and discriminator reward model are regularised by auxiliary learning objectives. 

\begin{figure}
    \centering
    \includegraphics[width=\textwidth]{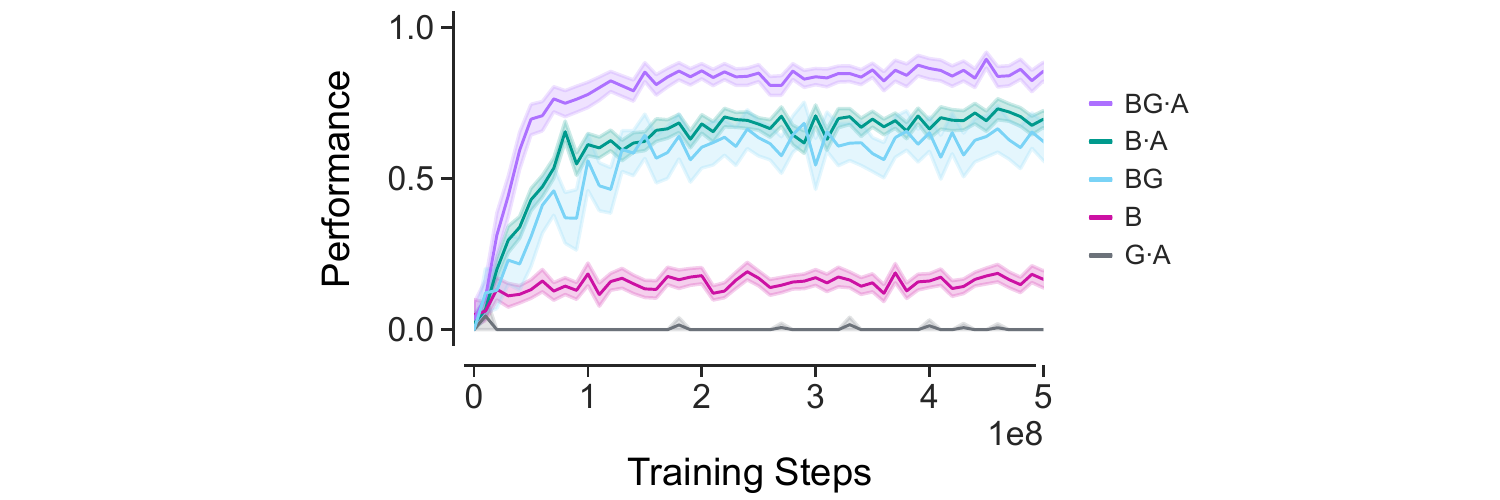}
    \caption{\textbf{Comparison of Imitation Learning Methods on Simple `Put {\em X} on Bed' Task.} In this task, an agent is instructed to put an object in the room on the bed using synthetic language. The data comprised $40,498$ human episodes pre-selected based on success. The GAIL agent (\GA), even with auxiliary loss regularisation of the agent and discriminator, failed to learn, while the simple BC (\B) agent learned to retrieve objects at random but did not identify the correct one. Combining BC with GAIL (\BG) or BC with auxiliary regularisation (\BA) improved performance. Further performance was reached by combining GAIL, BC, and auxiliary losses (\BGA). Note that certain possible comparison models were not run here, including simple GAIL (\textrm{G}), and variations that would use auxiliary losses on the agent but not the discriminator and vice versa.}
    \label{fig:simple_imitation_comparison}
\end{figure}

In Figure~\ref{fig:simple_imitation_comparison}, we compare the performance of our imitation learning algorithms applied to a simplified task in the Playroom. A dataset was collected of a group of subjects instructed using synthetic language to put an object in the room on the bed. A programmatic reward function that detects what object is placed on the bed was used to evaluate performance. Under no condition was the reward function used to train any agent. The agent and discriminator trained by GAIL with the regularisation (\GA; `\textrm{A}' denotes the inclusion of `auxiliary' regularisation, including the LM loss and \emph{RandAugment} on the discriminator) was unable to improve beyond its random initialisation. The behavioural cloning agent (\B{}) was slightly better but did not effectively understand the task: its performance implies it picked up objects at random and put them on the bed. Combining the behavioural cloning with GAIL (\BG) by simply adding the loss terms together achieved reasonable results, implying that GAIL was better at reshaping a behavioural prior than structuring it from scratch. However, behavioural cloning with the additional regularisation (\BA; LM and OV on the policy) achieved essentially the same or better results. Adding the auxiliary LM and OV losses to behavioural cloning and the GAIL discriminator was the best of all (\BGA). While this task is simple, we will show that this rough stratification of agents persisted even when we trained agents with complicated language games data and reported scores based on human evaluations.

\begin{table}[ht]
    \centering
    \small
    \begin{tabular}{rcccccc}
    \toprule
    {} & \multicolumn{2}{c}{Input modalities} & \multicolumn{3}{c}{Training algorithms} & \\
    \cmidrule(r){2-3} \cmidrule(r){4-6}
    Name           & Vision & Language & BC & GAIL & Setter replay & Auxiliary losses \\
    \midrule
    \mainagent{}   & \yes & \yes & \yes & \yes & \yes & \yes \\
    \BGA{}         & \yes & \yes & \yes & \yes & \no  & \yes \\
    \BG{}          & \yes & \yes & \yes & \yes & \no  & \no  \\
    \GA{}          & \yes & \yes & \no  & \yes & \no  & \no  \\
    \BA{}          & \yes & \yes & \yes & \no  & \no  & \yes \\
    \B{}           & \yes & \yes & \yes & \no  & \no  & \no  \\
    \BnoV{}        & \no  & \yes & \yes & \no  & \no  & \no  \\
    \BnoL{}        & \yes & \no  & \yes & \no  & \no  & \no  \\
    \bottomrule
    \end{tabular}
    \caption{Agent Nomenclature. Note that ``no vis.'' and ``no lang.'' indicate no vision and language input, respectively.}
    \label{tab:agent_nomenclature}
\end{table}

\subsubsection{Interactive Training}
While this training recipe is sufficient for simple tasks defined with programmed language and reward, to build agents from language games data requires further innovation to model both the setter and solver behaviour and their interaction. In this work, we train one single agent that acts as both a setter and a solver, with the agent engaged as a setter if and only if the language prompt $\mathbf{o}^\textsc{LP}$ is non-empty. In the original data, two humans interacted, with the setter producing an instruction, and the solver carrying it out. Likewise, during \emph{interactive training}, two agents interact together: one agent in the setter role receives a randomly sampled prompt, investigates the room, and emits an instruction; meanwhile another agent acts as the solver and carries out the instructed task. Together, the setter and solver improvise a small interaction scenario.

Both the setter and solver trajectories from the language games dataset are used to compute the behavioural cloning loss function. During interactive training, the solver is additionally trained by rewards generated by the GAIL discriminator, which is conditioned on the solver observation sequence. In this way, the setter generates tasks for the solver, and the solver is trained by reward feedback to accomplish them. The role of a human in commissioning instructions and communicating their preferences to critique and improve the agent's behaviour is thus approximated by the combined action of the setter agent and the discriminator's reward. 

We will see that interactive training significantly improves on the results of behavioural cloning. However, during the early stages of training, the interactions are wasted because the setter's language policy in particular is untrained. This leads to the production of erroneous, unsatisfiable instructions, which are useless for training the solver policy. As a method to warm start training, in half the episodes in which the solver is training, the Playroom's initial configuration is drawn directly from an episode in the language games database, and the setter activity is replayed step-by-step from the same episode data. We call this condition \emph{setter replay} to denote that the \emph{human} setter actions from the dataset are replayed. Agents trained using this technique are abbreviated `\mainagent{}' (`R' for \emph{Replay}). This mechanism is not completely without compromise: it has limited applicability for continued back-and-forth interaction between the setter and the solver, and it would be impractical to rely on in a real robotic application. Fortunately, setter replay is helpful for improving agent performance and training time, but not crucial. For reference, the abbreviated names of the agents and their properties are summarised in Table~\ref{tab:agent_nomenclature}.

\subsection{Evaluation}
The ecological necessity to interact with the physical world and with other agents is the force that has catalysed and constrained the development of human intelligence \citep{dunbar1993coevolution}. Likewise, the fitness criterion we hope to evaluate and select for in agents is their capability to interact with human beings. As the capability to interact is, largely, commensurate with psychological notions of intelligence  \citep{duncan2010intelligence}, evaluating interactions is perhaps as hard as evaluating intelligence \citep{turing, chollet2019measure}. Indeed, if we could hypothetically create an oracle that could evaluate any interaction with an agent -- e.g., how well the agent understands and relates to a human -- then, as a corollary, we would have already created human-level AI.

Consequently, the development of evaluation techniques and intelligent agents must proceed in tandem, with improvements in one occasioning and stimulating improvements in the other. Our own evaluation methodology is multi-pronged and ranges from \emph{simple automated metrics} computed as a function of agent behaviour, to fixed testing environments, known as \emph{scripted probe tasks}, resembling conventional reinforcement learning problems, to \emph{observational human evaluation} of videos of agents, to Turing test-like \emph{interactive human evaluation} where humans directly engage with agents. We also develop machine learning \emph{evaluation models}, trained from previously collected datasets of human evaluations, whose complexity is comparable to our agents, and whose judgements predict human evaluation of held-out episodes or held-out agents. We will show that these evaluations, from simple, scripted metrics and testing environments, up to freewheeling human interactive evaluation, generally agree with one another in regard to their rankings of agent performance. We thus have our cake and eat it, too: we have cheap and automated evaluation methods for developing agents and more expensive, large-scale, comprehensive human-agent interaction as the gold standard final test of agent quality. 

\section{Results}
As described, we trained agents with behavioural cloning, auxiliary losses, and interactive training, alongside ablated versions thereof. We were able to show statistically significant differences among the models in performance across a variety of evaluation methods. Experiments required large-scale compute resources, so exhaustive hyperparameter search per model configuration was prohibitive. Instead, model hyperparameters that were shared across all model variants (optimiser, batch size, learning rate, network sizes, etc.) were set through multiple rounds of experimentation across the duration of the project, and hyperparameters specific to each model variant were searched for in runs preceding final results. For the results and learning curves presented here, we ran two random seeds for each agent variant. For subsequent analyses, we chose the specific trained model seed and the time to stop training it based on aggregated performance on the scripted probe tasks. See Appendix sections~\ref{app:agent_training}, \ref{app:gail_training}, and \ref{app:distribtrain} for further experimental details. 

In what follows, we describe the automated learning diagnostics and probe tasks used to evaluate training. We examine details of the agent and the GAIL discriminator's behaviour in different settings. We then report the results of large-scale evaluation by human subjects passively observing or actively interacting with the agents, and show these are to some extent predicted by the simpler automated evaluations. We then study how the agents improve with increasing quantities of data, and, conversely, how training on multi-task language games protects the agents from degrading rapidly when specific tranches of data are held out. Using the data collected during observational human evaluation, we demonstrate the feasibility of training evaluation models that begin to capture the essential shape of human judgements about agent interactive performance.

\subsection{Training and Simple Automated Metrics}

The probability that an untrained agent succeeds in any of the tasks performed by humans in the Playroom is close to zero. To provide meaningful baseline performance levels, we trained three agents using behavioural cloning (BC, abbreviated further to B) as the sole means of updating parameters: these were a conventional BC agent (\B{}), an agent without language input (\BnoL{}) and a second agent without vision (\BnoV{}). These were compared to the agents that included auxiliary losses (\BA{}), interactive GAIL training (\BGA{}), and the setter replay (\mainagent{}) mechanism. Since \mainagent{} was the best performing agent across most evaluations, any reference to a default agent will indicate this one. Further agent ablations are examined in Appendix~\ref{app:agent_training}. 

\begin{figure}[ht!]
    \centering
    \includegraphics[width=1\textwidth]{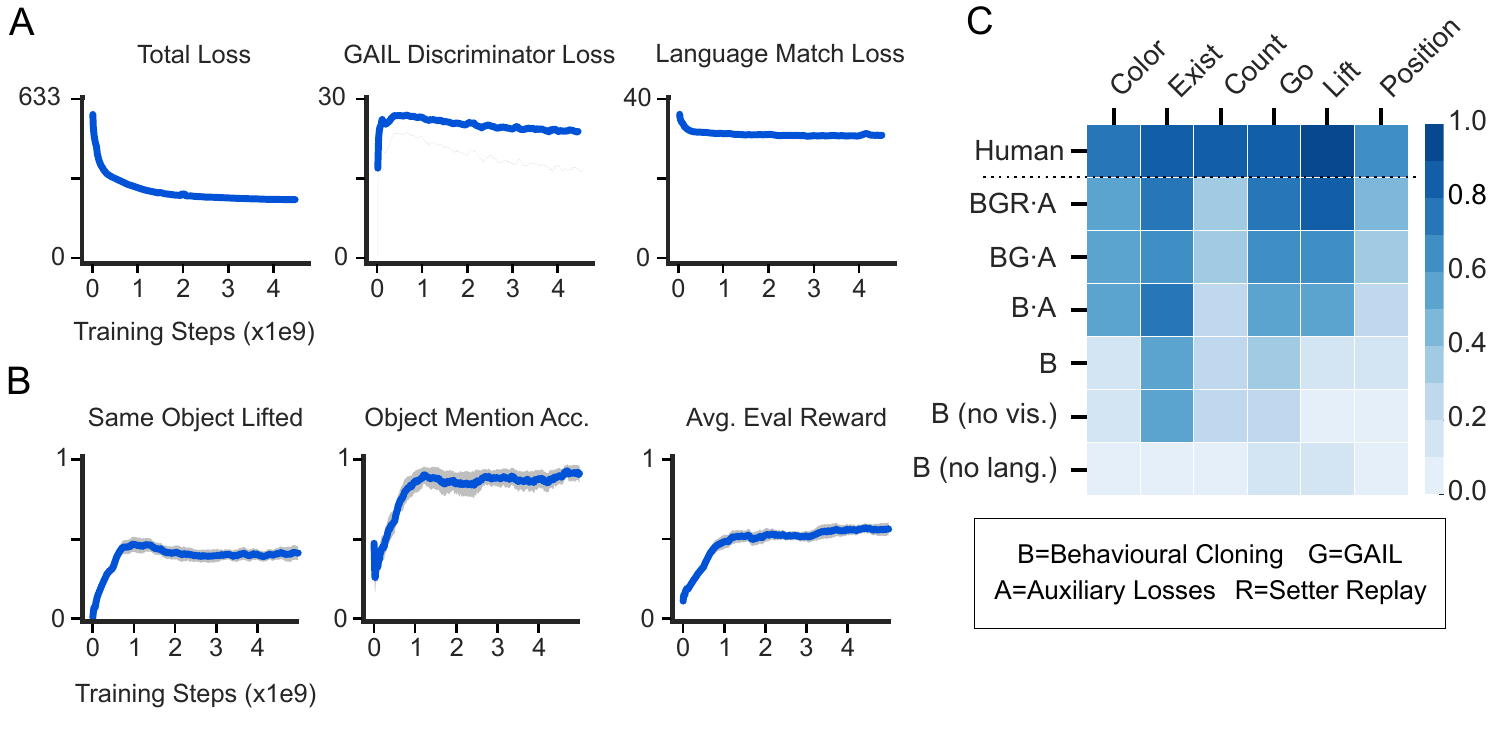}
    \caption{\textbf{Learning Metrics.} \textbf{A.}  The top row shows the trajectory of learning for three training losses: the behavioural cloning loss (top left, total loss which includes losses for the motor actions, language actions, and auxiliary tasks scaled accordingly to their relative contribution), the GAIL discriminator loss (top middle), and the language matching auxiliary loss (top right). \textbf{B.} The bottom row shows tracked heuristic measures along the same trajectory, which proved useful in addition to the losses for assessing and comparing agent performance. Same Object Lifted measures whether the solver agent has lifted the same object as the human in the equivalent validation episode; Object Mention Accuracy measures whether an object is indeed within the room if it happens to be mentioned by the setter in a validation episode; and Average Evaluation Reward measures the reward obtained by a solver agent when trying to solve scripted probe tasks that we developed for agent development. (Rewards in these tasks were not used for training, just for evaluation purposes.) \textbf{C.} Agent and human performance compared on the same scripted probe tasks. Agents were divided based on their included components (e.g., trained purely by behavioural cloning or also by interactive training with GAIL, or whether they were ablated agents that, for example, did not include vision). We observed a gradual improvement in agent performance as we introduced auxiliary losses, interactive training, and setter replay. 
    }
    \label{fig:curves}
\end{figure}

Figure~\ref{fig:curves}A shows the progression of three of the losses associated with training the \mainagent{} agent (top row), as well as three automated metrics which we track during the course of training (bottom row). Neither the BC loss, the GAIL discriminator loss, nor the auxiliary losses directly indicates how well our agents will perform when judged by humans, but they are nonetheless useful to track whether our learning objectives are being optimised as training progresses. Accordingly, we see that the BC and Language Match losses were monotonically optimised over the course of training. The GAIL discriminator loss increased as agent behaviour became difficult to distinguish from demonstrator behaviour and then descended as the discriminator got better at distinguishing human demonstrators from the agent. Anecdotally, discriminator over-fitting, where the discriminator assigned low probability to held-out human demonstrator trajectories, was a leading indicator that an agent would behave poorly. Automated metrics played a similar role as the losses: on a validation set of episodes with a setter replay instruction, we monitored whether the first object lifted by a solver agent was the same as that lifted by a human. We also measured if object and colour combinations mentioned by the agent were indeed in the room. Intuitively, if this metric increased it indicated that the agent could adequately perceive and speak about its surroundings. This was an important metric used while developing setter language. However, it is only a rough heuristic measure: utterances such as, ``Is there a train in the room?'' can be perfectly valid even if there is indeed no train in the room. 

\subsection{Scripted Probe Tasks}
\label{sec:probe_tasks}
In the general case, it is impossible to write a program that checks if an interaction between a human and an agent (or between two agents) has ``succeeded,'' even in the context of a virtual environment. However, for certain very canonical interactions, with a specific flavour of success criterion, it is possible to write down propositions describing physical states of the environment that approximate human judgements about the correctness of following instructions or answering questions. We therefore developed six \emph{scripted probe tasks} in which the linguistic behaviour of the \emph{setter} was scripted to provide clear instructions or questions (e.g., ``Pick up the X''; ``Put the X near the Y''; ``What colour is the X?''). Three of these were instruction following ({\tt Go}, {\tt Lift}, {\tt Position}) and three question answering ({\tt Colour}, {\tt Exist}, {\tt Count}) (see Figure~\ref{fig:curves} and Appendix~\ref{app:probe_tasks} for details) The responses to these instructions or questions could be unambiguously scored (under certain assumptions) by callbacks from the environment engine. Thus, the probe tasks aimed to provide a cheap and unambiguous way of scoring the behaviour of the solver agent in a way that approximates the language games played by humans but without requiring costly human evaluation. During learning we monitored the average performance of our solvers across a set of these probe tasks (Figure~\ref{fig:curves}, Avg.~Eval.~Reward). 
Figure~\ref{fig:curves}B shows the performance of human players and the trained solver agents across these tasks. Overall, the interactively trained agents, with or without setter replay, performed as well as or better than all comparisons. See Appendix Table~\ref{tab:scripted_probes} for precise numeric values.

To establish baselines, we measured human performance on these tasks without providing feedback about success as the humans played. Interestingly, we found that, even though the tasks involve elementary challenges like picking up and placing objects relative to each other, human performance under these conditions (which are the same conditions faced by the agent) was evaluated to be good but not perfect. This underlines the fact that, even for instruction-following and question-answering tasks that require little planning, reasoning, or dexterous motor control, what constitutes success is subjective, and the intuitions human participants brought to bear when deciding they had completed tasks did not always match our own programmed definition of task success. Furthermore, for more nuanced types of interaction, we would have been unable to program rule-based evaluations at all.

\subsection{Action Prediction Metrics}
\label{sec:bc_validation}
\begin{figure}[h]
    \centering
    \includegraphics[width=1\textwidth]{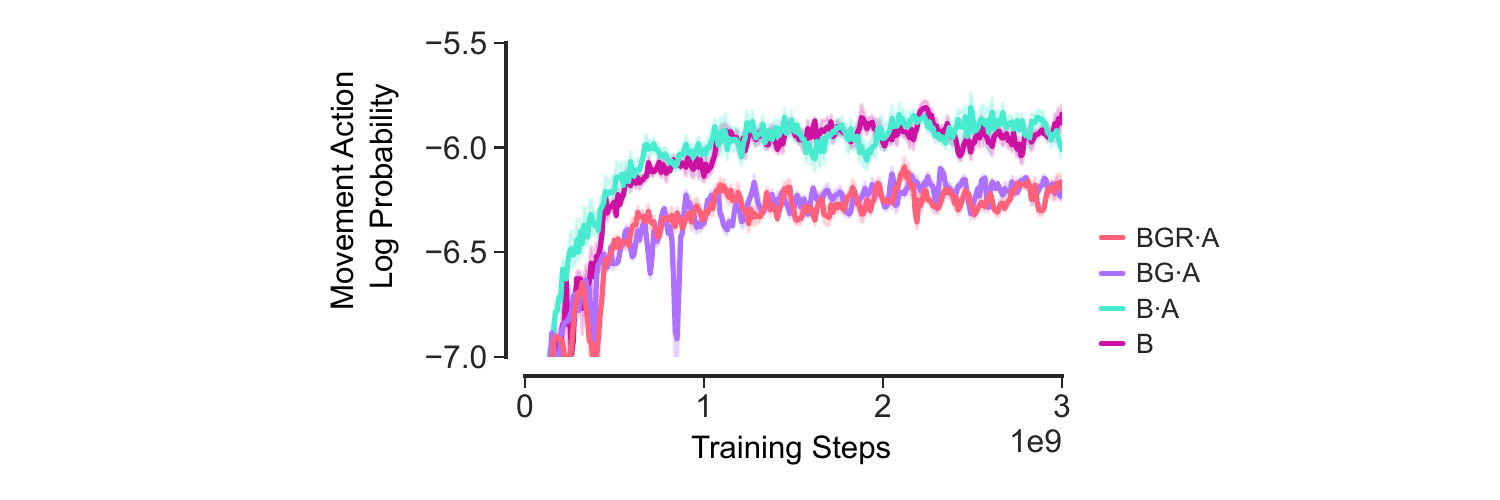}
    \caption{\textbf{Behavioural Cloning Validation Metrics.} Models trained by interaction (\mainagent{} \& \BGA{}) performed better than those that were not (\BA{} \& \B{}) in scripted probe task performance (Figure~\ref{fig:curves}C), but worse in terms of the BC validation set log probability (depicted here).
    }
    \label{fig:bcvalid}
\end{figure}

We also tracked performance at predicting human actions on a validation set of human demonstrations during training -- that is, the behavioural cloning validation set loss. Tracking this metric allowed us to observe over-fitting and other training-related problems. However, as we will see, the BC validation metric was not on its own always a useful guide for understanding agent task performance. To compute the metric, we held out a random subset of the human demonstration data and examined how well our agent predicted the human actions while the agent processed the observations derived from the trajectories. In the Playroom, the agents use motor actions and language actions. Figure~\ref{fig:bcvalid} shows the validation log probabilities for motor actions taken by our agent in the solver role. Training drove performance on this metric up both for our agent and main ablations. Strikingly, both agents trained interactively via GAIL (\mainagent{} and \BGA{}) performed worse on with regard to behavioural cloning loss on the validation set than agents trained to produce actions via BC alone (\B{} and \BA{}). This is notable given what we observed in the scripted probe tasks shown in Figure~\ref{fig:curves}C -- that interactive training produced the best performing agents. As we will see, human judgement of task success agreed more closely with the probe task evaluation. Thus, while convenient and sometimes instructive, BC validation set performance was unreliable for understanding how well agents perform tasks as directed and evaluated by humans. BC validation curves for language actions and the setter role are shown in Appendix~\ref{app:agent_training}. 

\subsection{Automated Setter Metrics}
\label{sec:setter_metrics}


Table~\ref{tab:setter_metrics} shows automated metrics we used to help develop agents' capacities to perform in the role of the setter. These metrics could be measured while training, offering hints about where training was failing, and which agent variations might perform better. We measured: 1.~if setters referred to objects in the room; 2.~the average number of words in an utterance; 3.~the average number of utterances produced in an episode; 4.~the 1-gram entropy of the utterances.
To a first approximation, a model's statistics should roughly match the human distributions, which are also shown in Table~\ref{tab:setter_metrics}. Our agent performed better than the behavioural cloning baseline $\B$, but GAIL was not a key factor (as it was not used directly to optimise the setter behaviour). Rather, the main driver of success was the introduction of auxiliary losses, which we believe helped the model to link visual information with linguistic content. 

\begin{table}[ht]
    \small
    \centering
    \begin{tabular}{rM{2.5cm}M{2.7cm}M{2.6cm}M{2cm}}
\toprule
{} & Obj. mention accuracy & Avg. utterance length (words) & Avg. num. utterances & Entropy \\
\midrule
Human & $0.870\pm{0.011}$ & $6.31\pm{0.04}$ & $1$ & $6.1\pm{0.2}$ \\
\mainagent{} & $0.686\pm{0.007}$ & $5.59\pm{0.02}$ & $0.856\pm{0.003}$ & $5.8\pm{0.2}$ \\
\BGA{} & $0.691\pm{0.007}$ & $5.78\pm{0.02}$ & $0.893\pm{0.003}$ & $5.8\pm{0.3}$ \\
\BA{} & $0.660\pm{0.007}$ & $5.75\pm{0.02}$ & $0.926\pm{0.004}$ & $5.8\pm{0.3}$ \\
\B{} & $0.241\pm{0.007}$ & $5.67\pm{0.02}$ & $0.845\pm{0.003}$ & $5.8\pm{0.2}$ \\
\BnoL{} & $0.255\pm{0.008}$ & $5.29\pm{0.03}$ & $0.846\pm{0.004}$ & $6.0\pm{0.2}$ \\
\BnoV{} & $0.077\pm{0.005}$ & $5.68\pm{0.02}$ & $0.777\pm{0.004}$ & $5.9\pm{0.2}$ \\
\bottomrule
\end{tabular}

    \caption[Automated setter metrics]{
    \textbf{Automated Setter Metrics.} Object Mention Accuracy calculates how often a colour adjective with an object name is found in the room. This measure is not always perfect since humans can use colours that are not detected by our internal dictionary of acceptable answers; hence the imperfect human score. The improvement of auxiliary losses over behavioural cloning is particularly notable. Human episodes were filtered to include one and only one instruction.}
    \label{tab:setter_metrics}
\end{table}

To ground our intuitions, we examined the word frequencies of our agent's utterances when it played as the setter. To compute these metrics consistently across agent variants, we forced the agent observations explicitly along the human demonstration episodes in a held-aside validation set (see Appendix~\ref{app:setter_metrics} for details). Figure~\ref{fig:word_freq}A plots the word frequencies from human setter utterances. For illustrative purposes, Figure~\ref{fig:word_freq}B plots these frequencies versus those computed for human setter utterances for a subset of words. The data are clustered around the unity line, indicating that our agent uttered a particular word about as often as humans did in the same circumstances. For comparison, Figure~\ref{fig:word_freq}C shows the agent produced word frequency versus those for a dataset constructed from Wikipedia \citep{Wiki40B}. 

\begin{figure}[h]
    \centering
    \includegraphics[width=0.8\textwidth]{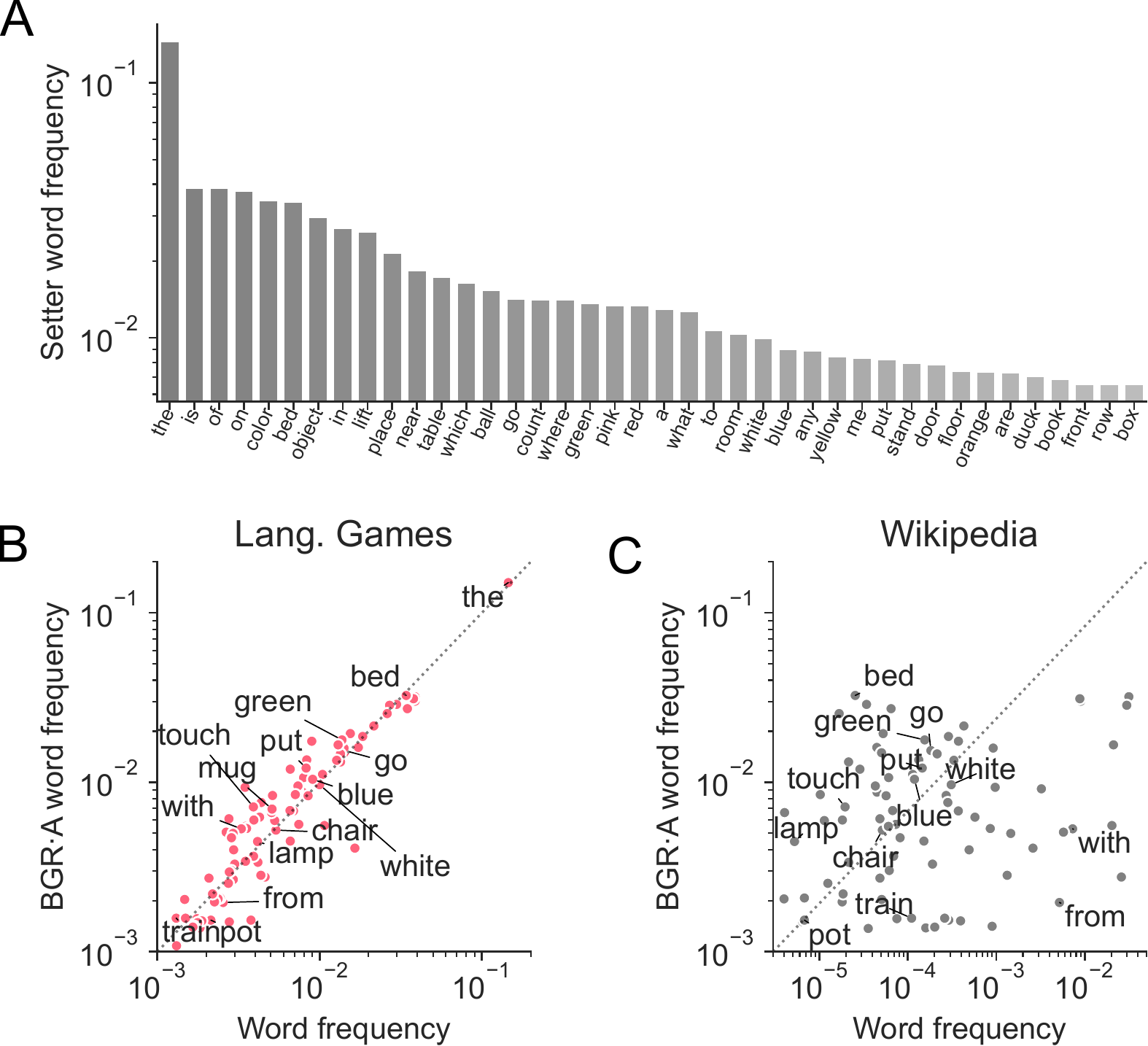}
    \caption{\textbf{Language Diversity in Setter Utterances.} \textbf{A.} Frequency of the most common words in human setter language emissions. \textbf{B.} Frequency of the top-100 most common words in the BGR$\cdot$A agent setter emissions versus human setter language emission and \textbf{C.} versus the English Wiki40B dataset \citep{Wiki40B}. 
    }
    \label{fig:word_freq}
\end{figure}

\subsection{Agent Behaviour and Discriminator Reward Traces}
\label{sec:agent_behaviour}

\begin{figure}[h]
    \centering
    \includegraphics[width=1\textwidth]{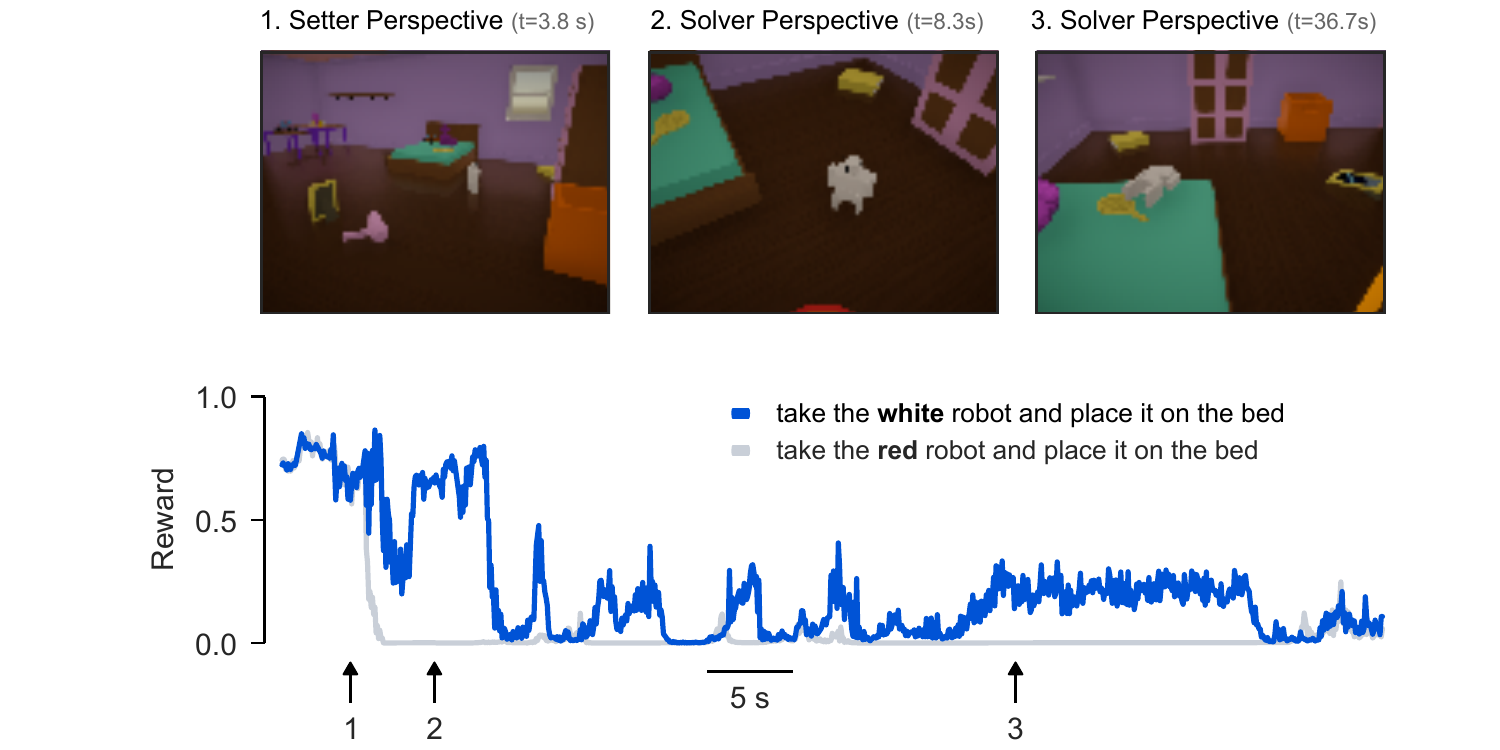}
    \caption{\textbf{Single Episode Agent Behaviour and Discriminator Reward Traces.} The setter viewed the room [1], and asked the solver to ``take the white robot and place it on the bed''. The solver found the correct object [2], and lifted it onto the bed [3]. The GAIL reward model gave positive reward, temporally correlated with finding and depositing the object (blue, at [2] \& [3]). It gave less reward when, instead of the original instruction, the reward model received the counterfactual instruction, ``take the {\em blue} robot and place it on the bed,'' which was inconsistent with the visual observations (grey). In both cases, reward was high at the beginning of the episode because the GAIL discriminator was uncertain about classifying between imitator agent and demonstrator human behaviour while the solver agent awaited the setter instruction.
    }
    \label{fig:reward_model}
\end{figure}

Figure~\ref{fig:reward_model} encapsulates a single episode performed by the \mainagent~agent. The prompt for this episode requested that the setter  ``Ask the player to position something relative to something else''. The setter followed the prompt by asking the solver agent to ``take the white robot and place it on the bed.'' The top row shows the solver finding the object and placing it on the bed. The lower panel of Figure~\ref{fig:reward_model} shows the corresponding output of the GAIL discriminator reward model over the course of the episode. The model gave positive reward at several points during the episode, especially at points where the agent interacted with the correct object. Since our GAIL model takes the setter language as input along with the solver vision, we are also able to examine counterfactual scenarios. We altered the colour in the setter utterance to make ``take the {\em red} robot and place it on the bed,'' and reran the reward model over the episode. This new request was impossible to fulfil given that no red robot existed in the room. Correspondingly, in the counterfactual condition the GAIL discriminator yielded little reward throughout the episode. Thus, the reward model appears to possess some understanding the consistency of a setter instruction and the solver agent behaviour. 

\subsection{Observational human evaluation}
\label{sec:offline_eval}

\begin{figure}[h]
    \centering
    \includegraphics{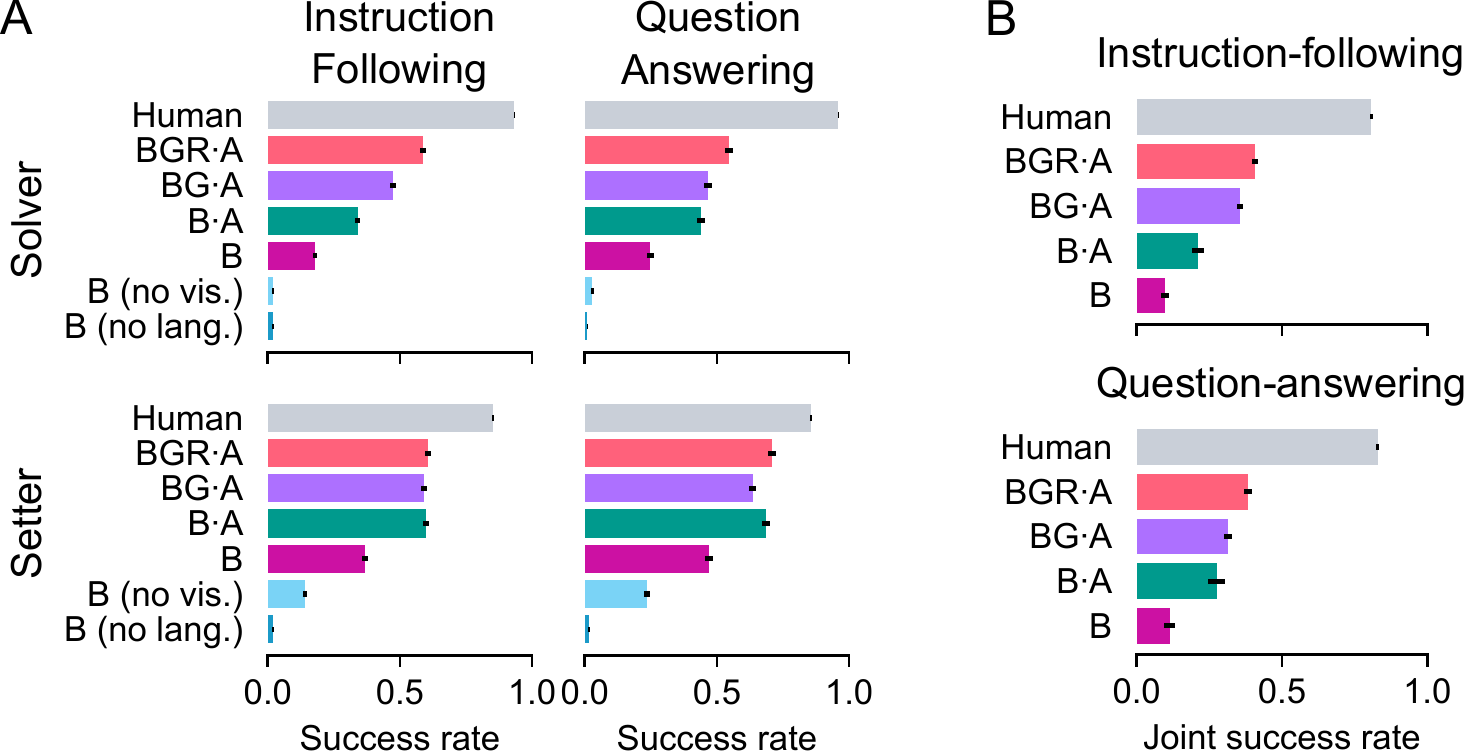}
    \caption[Observational human evaluation of agent performance]{
        \textbf{Observational Human Evaluation of Agent Performance.} \textbf{A:} Success rates for agents performing the role of either solver or setter, as judged by human annotators.
        Agent solver and setter episodes were generated by rolling out a pre-trained policy for $\sim$200 episodes per script.
        The bars represent the proportion of episodes that were marked as ``successful'' by human annotators.
        Each bar represents a weighted average over all prompts within the movement or question-answering categories.
        Each script was weighted according to its frequency within the human demonstration data that was used to train the agents.
        The human baseline was calculated using annotations of episodes from the human demonstration data.
        Error bars represent a 95\% CI of the mean.
        \textbf{B:} Joint success rates for episodes where the same pre-trained policy performed the roles of both setter and solver.
        In this case the setter and solver trajectories for each episode were annotated separately, and only episodes where both the setter and solver were labelled as successful.
    }
    \label{fig:offline_human_eval}
\end{figure}

One step closer to our ultimate interactive evaluation of agent behaviour, we simulated rollouts of agents playing as either the setter or the solver and asked humans to score whether the behaviour was correct (Figure~\ref{fig:offline_human_eval}A).
These rollouts were then evaluated offline using an interface that allowed human raters to skip forwards and backwards through each trajectory of observations and text emissions~\citep{cabi2019scaling}.
The raters were asked to score each episode as either ``successful'' or ``unsuccessful.'' 
For successful episodes, the raters were also asked to mark the moment in time when success first occurred.
This is a relatively high throughput method in comparison to interactive evaluation (Section \ref{sec:online_eval}), since simulated rollouts can be generated much faster than real-time in large batches, and a human rater can typically judge whether or not an episode was successful in much less time than it would take to execute a live interaction with an agent.
Using this paradigm we were able to collect on the order of 10,000 annotated episodes for each of our agents.

To evaluate solvers in this mode, we replayed human setter actions (both language and motor) from episodes in a held out test set of demonstration episodes.
Since setter actions were replayed without regard to the solver's activity, this approach was limited to interactions that do not involve back-and-forth dialogue or active cooperation between the setter and solver (we excluded two prompts -- ``hand me'' and ``do two things in a row'' -- for this reason).
In addition, there are cases where the replayed actions of the setter may impede the solver's ability to complete the task (for example, by disturbing other objects in the room).
These cases make up a very small fraction of episodes and only contribute negatively to agent evaluation. 

To evaluate agents in the setter role, a dummy solver agent with no control policy was placed in the environment. Human observers were asked to determine that the setter produced an utterance which was consistent with the prompt as well as what the setter saw in the room up to the point of the language emission. If no utterance was emitted by the setter, the episode was deemed unsuccessful. 

We used the same interface and instructions to have humans evaluate episodes carried out by pairs of humans in our main dataset.
As expected, humans were judged as completing all of our tasks (setter \& solver, action \& language) with high fidelity ($>$90\% success rate; grey bars in Figure~\ref{fig:offline_human_eval}).
Humans may disagree about what counts as success due to inherent ambiguity (for example whether a particular object is close enough to be considered `near'), or may be be incorrect in their judgement due to a misreading or lack of attention.
We did not attempt to disambiguate between these two cases. 
In order to measure the degree of inter-rater agreement we collected multiple annotations for a subset of human and agent episodes.
We treated the majority label for each episode as the ground truth (in the case of a tie between successful and unsuccessful annotations the episode was considered unsuccessful), and measured the proportion of individual annotations that were in agreement with the majority label.
The proportion of annotations that were in agreement with the majority label was 87.56\%$\pm$0.22 for human solver episodes, and 91.88\%$\pm$0.05 for human setter episodes.
We obtained similar results for annotations of agent episodes (see Table~\ref{tab:rater_agreement} for detailed results).

The top row of Figure~\ref{fig:offline_human_eval}A shows the success rates for human and agent solvers, as judged by human raters.
When evaluated as solvers, the \BnoL{} and \BnoV{} baseline agents were able to successfully complete the setter's instruction in less than 5\% of episodes, and the model trained with BC alone succeeded 20.12\%$\pm$1.13 of the time.
In contrast, the \mainagent{} agent was judged to be successful 57.02\%$\pm$0.89 of the time.
Ablations \BA{} and \BGA{} were judged to perform at an intermediate level (37.28\%$\pm0.84$ and 46.80\%$\pm$0.88 respectively).
The bottom row of panel A shows equivalent results for setter episodes.
The success rates for setter episodes were higher overall in comparison to solver episodes.
In particular the \BnoV{} baseline agent achieved a much higher success rate as a setter than as a solver (17.77\%$\pm$0.69, compared to 2.27\%$\pm$0.30), reflecting the fact that it is often possible for a setter to give a valid instruction without attending to the initial state of the room.
Overall, these results speak clearly to the advantage of using auxiliary objectives and interactive training for improving solver agents beyond straightforward BC in the context of grounded language interactions. 
Although the agents do not yet attain human-level performance, we will soon describe scaling experiments which suggest that this gap could be closed substantially simply by collecting more data.
Perhaps most crucially, even when the \mainagent{} agent failed to perform a given task, it frequently performed sequences of actions that were ``close'' to what was asked.
Thus, we believe it is a good candidate to be optimised further using human evaluative feedback.

We also examined the performance of our best performing agents in {\em joint} episodes, in which the same agent performed the roles of both the setter and the solver in the interaction.
As before, human raters annotated both sides (setter \& solver) of these entirely simulated interactions.
We considered an episode to be a joint success only if {\em both} the setter and the solver were marked as successful by humans.
Figure~\ref{fig:offline_human_eval}B shows that the \mainagent{} was successful in playing both sides of the interaction for 39.58\%$\pm$0.9 of episodes.
Thus, agents were often capable of both setting tasks relevant to their surroundings, as well as responding intelligently to those requested tasks. Combined with automated success labelling, which we will explore later in this document, this capability may open the door to using self-play as a mechanism for optimising behaviour.
As expected, the B, \BA{}, and \BGA{} models were less capable at completing jointly successful episodes, achieving success rates of 10.38\%$\pm$1.15, 23.59\%$\pm$1.67,  and 33.89\%$\pm$0.87 respectively.
Figure~\ref{fig:obs_eval_results_by_prompt} in the Appendix contains a more detailed breakdown of agent performance according to prompt.

\subsection{Interactive Human Evaluation}
\label{sec:online_eval}
Finally, we evaluated the ability of our agents to engage in direct interactions with humans.
In these experiments, humans played the role of the setter\footnote{We did not evaluate setter agents in a fully interactive mode because, for all but one of the tasks we explored, the solver behaviour is largely irrelevant to the success of the setter. That is, setter success is determined by the prompt and what they see up to their first utterance.} just as they do in the human-human episodes we collected: they received a prompt, looked around the room and expanded the prompt to an instruction, observed the agent, and terminated the episode when they considered it solved, or were certain that the solver had failed.
These human-agent interactions were recorded, and then the solver (i.e. agent) side of each interaction was annotated offline by human raters, using the same interface as in Section~\ref{sec:offline_eval}.
Compared to purely observational evaluation, where humans could fast-forward through movies, interactive evaluation is a relatively low throughput method, since each human player can interact with only a single agent at a time, and the interactions must happen in real time.
We collected a total of 27,895 annotated episodes across four different agents.

\begin{figure}[h]
    \centering
    \includegraphics{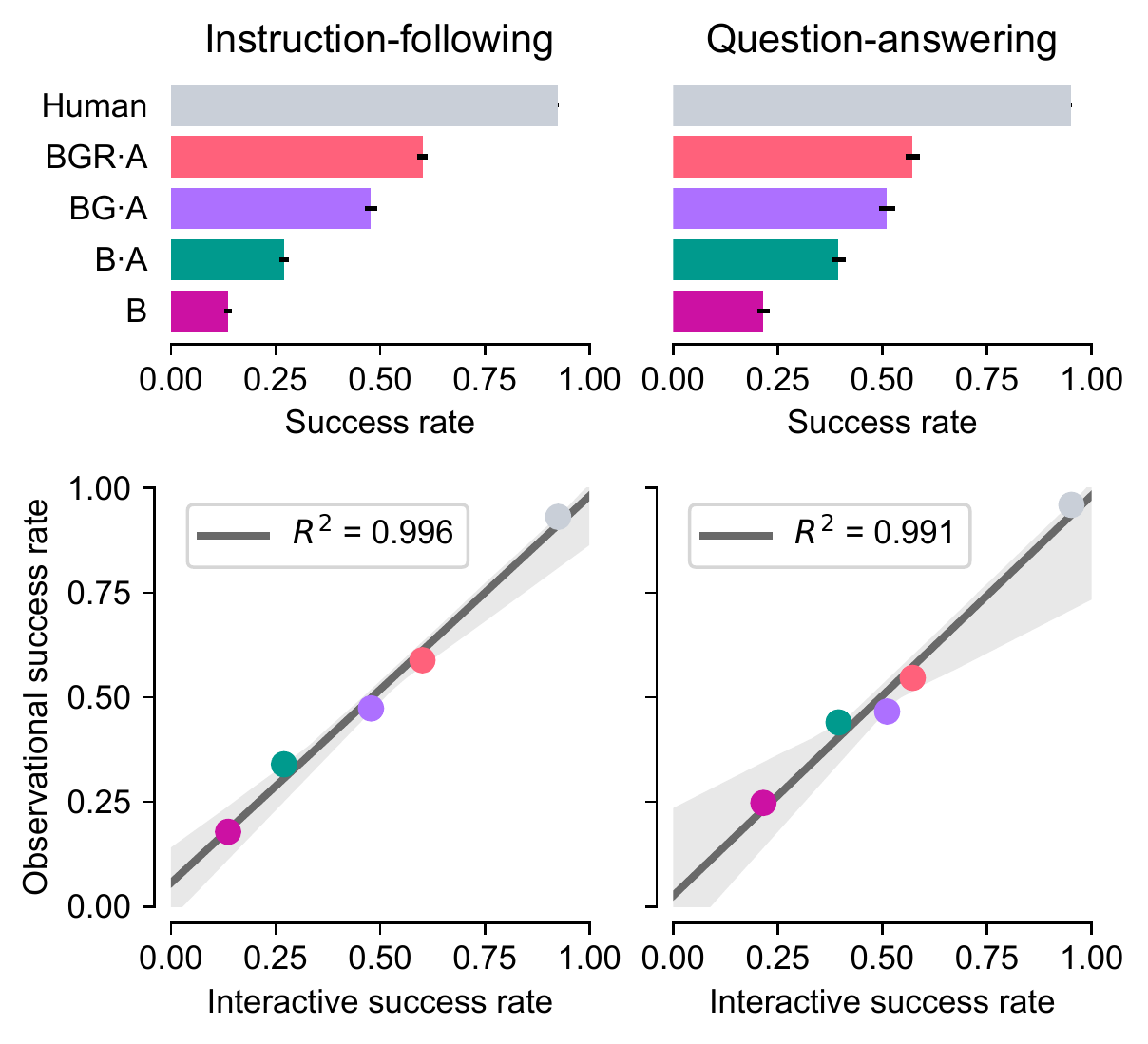}
    \caption[Interactive human evaluation of agent performance]{\textbf{Interactive Human Evaluation.}
    \emph{Top row:} mean solver success rates for live interactions, categorised as instruction-following or question-answering, where a human played the role of the setter, as judged by human raters.
    The human baselines (grey bar) represent live human-human interactions, as shown in the top row of Figure~\ref{fig:offline_human_eval}A.
    Error bars denote a 95\% CI of the mean.
    \emph{Bottom row:} scatter plots comparing the mean success rates achieved for interactive evaluation (x-axis) and observational evaluation (y-axis).
    The observational success rates are the same values plotted in the top row of Figure~\ref{fig:offline_human_eval}A.
    }
    \label{fig:online_human_eval}
\end{figure}

Figure~\ref{fig:online_human_eval} shows the interactive human evaluation results for the agents.
Both the ordering and the absolute magnitudes of the success rates for live human-agent interactions correspond closely to those for observational evaluation.
Our agent was judged to be successful 59.01\%$\pm$1.06 of the time during human-agent interactions (60.10\%$\pm$1.32 and 57.25\%$\pm$1.75 for action and question-answering tasks respectively).
This is slightly higher than the average success rate for this agent in observational evaluations (57.02\%$\pm$0.89).
One possible explanation for this difference is that in the interactive setting the human setter may react to the solver's position and, for example, stay out of its way.

\subsection{Scaling \& Transfer}
\label{sec:scaling}
It is natural to wonder how the highest-performing agent would have improved if we had collected and trained with more data, and how it generalises to unseen situations. We ran experiments to examine the scaling \citep{kaplan2020scaling} and transfer properties of imitation learning for behaviour in the Playroom.

\begin{figure}[hp]
    \centering
    \includegraphics[width=\textwidth]{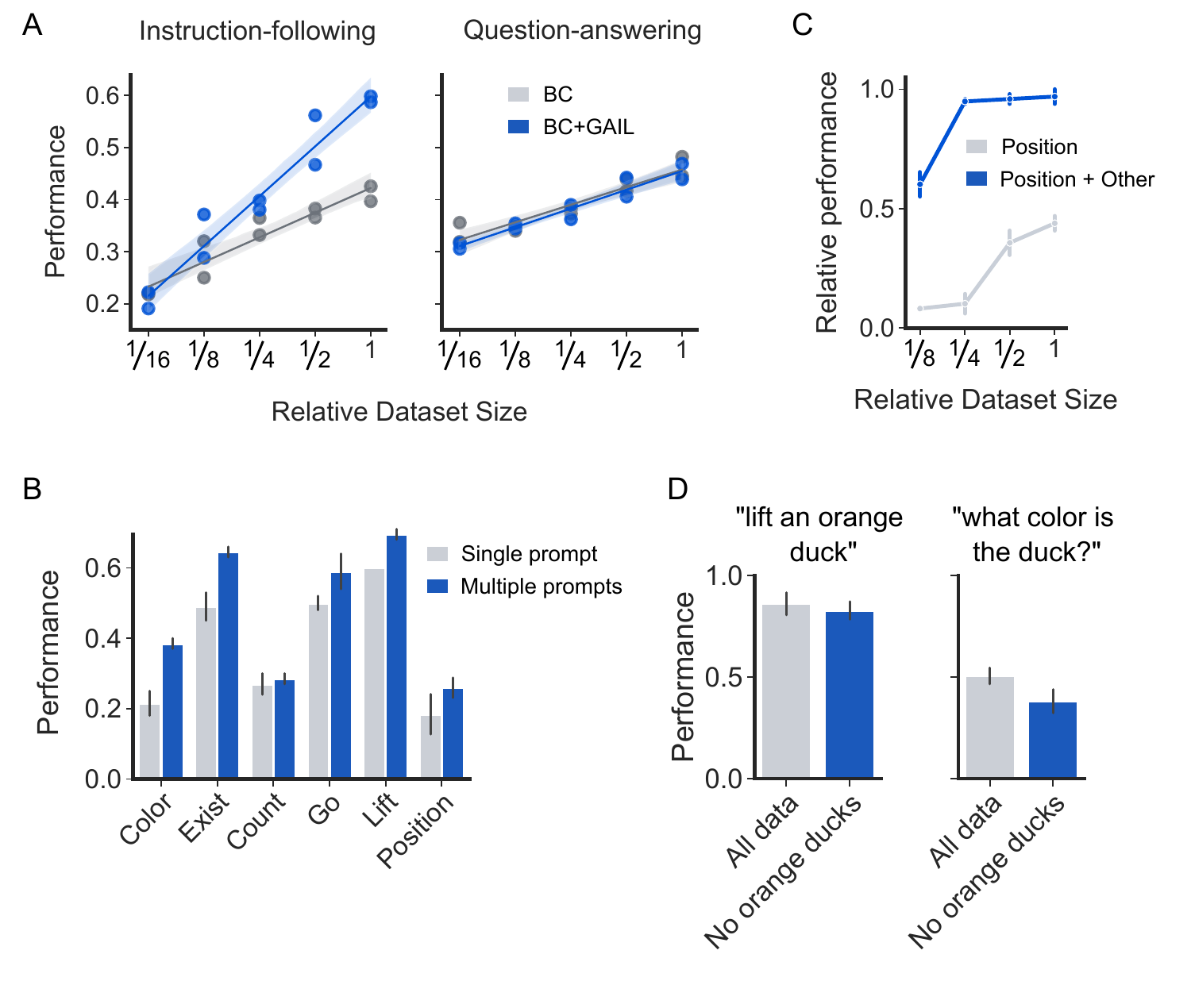}
    \caption{\textbf{Scaling \& Transfer.} \textbf{A}. Scaling properties for two of our agents. The agent's performance on the scripted probe tasks increased as we trained on more data. In instruction-following tasks in particular, the rate of this increase was higher for BC+GAIL compared to BC (scatter points indicate seeds). \textbf{B}. Transfer learning across different language game prompts. Training on multiple language games simultaneously led to higher performance than training on each single prompt independently. \textbf{C}. Multitask training improved data efficiency. We held out episodes with instructions that contain the words ``put,'' ``position'' or ``place'' and studied how much of this data was required to learn to position objects in the room. When simultaneously trained on all language game prompts, using $\frac{1}{8}$ of the {\tt Position} data led to $60\%$ of the performance with all data, compared to $7\%$ if we used the positional data alone. \textbf{D}. Object-colour generalisation. We removed all instances of orange ducks from the data and environment, but we left all other orange objects and all non-orange ducks. The performance at scripted tasks testing for this particular object-colour combination was similar to baseline.}
    \label{fig:scaling}
\end{figure}

First, we examined how the performance of our agents changed as a function of the size of the dataset trained on. We trained the \BA{} and \BGA{} agents using random splits of $\frac{1}{16}$,  $\frac{1}{8}$,  $\frac{1}{4}$, and $\frac{1}{2}$ the size of our full training set. Figure~\ref{fig:scaling}A shows the average performance across the instruction-following and question-answering scripted probe tasks for these dataset sizes. The scripted probe tasks are imperfect measures of model performance, but as we have shown above, they tend to be well correlated with model performance under human evaluation. With each doubling of the dataset size, performance grew by approximately the same increment. The rate of performance, in particular for instruction-following tasks, was larger for the \BGA{} model compared to \BA{}. Generally, these results give us confidence that we could continue to improve the performance of the agents straightforwardly by increasing the dataset size.

We examined the question of whether our agents transferred knowledge from several angles. First, Figure~\ref{fig:scaling}B shows the results of training across multiple prompts at once versus training on the data associated with a single prompt. Assessed via the six scripted probe tasks, a model that trained across all prompts performed as well as or better than a model that only trained on the data corresponding to a single prompt.

A signature of transfer learning is that agents would require less data to learn new tasks given a background of previous knowledge. To test this, we divided our data into two sets: one in which the instruction given by the setter contained the words ``put,'' ``position,'' or ``place'', which we refer to as the positional dataset, and the complement of this set. We then trained on varying fractions ($\frac{1}{8}$, $\frac{1}{4}$, $\frac{1}{2}$, 1) of the positional data in isolation, or in conjunction with the second set of data, that is, all other setter instructions. Figure~\ref{fig:scaling}C shows the performance of \BGA{} models trained using these splits on the {\tt Position} scripted probe task. When trained in conjunction with all other setter instructions, the model performed better with only $\frac{1}{8}$ of the positional data than when trained with all of the positional data alone.

Zooming in further on the question of generalisation, we randomly selected one object-colour combination, orange ducks, and removed all instances of orange ducks from all training data, including both human demonstration data and interactive training episodes. In total we removed 23K episodes containing orange ducks, regardless of whether they where referred to by the setters or not. Importantly, we kept episodes with other orange objects and those with non-orange ducks. This was possible using the game engine to check which object types/colours were present in a given configuration of the Playroom. We then trained the \BGA{} model on either this reduced dataset or on all of the data. After training, we asked the models to ``Lift an orange duck'' or ``What colour is the duck?'' We examined the performance for these requests in randomly configured contexts appropriate for testing the model's understanding. For the {\tt Lift} instruction, there was always at least one orange duck in addition to differently coloured distractor ducks. For the {\tt Color} instruction, there was a single orange duck in the room. Figure~\ref{fig:scaling}D shows that the agent trained without orange ducks performed almost as well on these restricted {\tt Lift} and {\tt Color} probe tasks as an agent trained with all of the data. These results demonstrate explicitly what our results elsewhere suggest: that agents trained to imitate human action and language demonstrate powerful combinatorial generalisation capabilities. While they have never encountered the entity, they know what an ``orange duck'' is and how to interact with one when asked to do so for the first time. This particular example was chosen at random; we have every reason to believe that similar effects would be observed for other compound concepts.

\subsection{Evaluation Models}
\label{sec:eval_models}

Our results thus far show how to leverage imitation learning to create agents with powerful behavioural priors that generalise beyond the instances they have been trained on. We have relied on scripted probe task evaluations during training, but these are labour intensive to build, and we expect they will be increasingly misaligned with human intuitions as the complexity of tasks increases.
Looking forward, we are interested in whether it is possible to automate the evaluation of agents trained to interact with humans. Ultimately, if a model robustly captures task reward, we may wish to directly optimise it. To this end, we trained network models to predict the success/failure labels annotated by humans on our human paired data. Here we report results for instruction-following tasks. Early experiments with similar models for question-answering data are reported in Appendix~\ref{app:eval_models}. 

We trained the evaluation model exclusively on human instruction-following task data. Humans labelled paired human episodes as successful $93.27$\% of the time. Evaluation therefore needs to contend with significant class imbalance, so we tracked balanced accuracy as our main metric for model performance. Though we trained models on only human instruction-following episode data, we selected our best models using balanced accuracy computed on a mixture of human validation data as well as data from two previously trained agents (which we refer to as a ``validation score"; for more details, see Appendix~\ref{app:eval_models_selection}). 
We use balanced accuracy as a metric throughout this section since episodes are unbalanced with respect to success and failure --- a model that merely predicts success 100\% of the time would be correct $93.27$\% of the time for human data. 
Balanced accuracy is computed as the average of the proportion of correct predictions across the two classes: (\% successes predicted correctly + \% failures predicted correctly) / 2. 

\begin{figure}[h]
    \centering
    \includegraphics{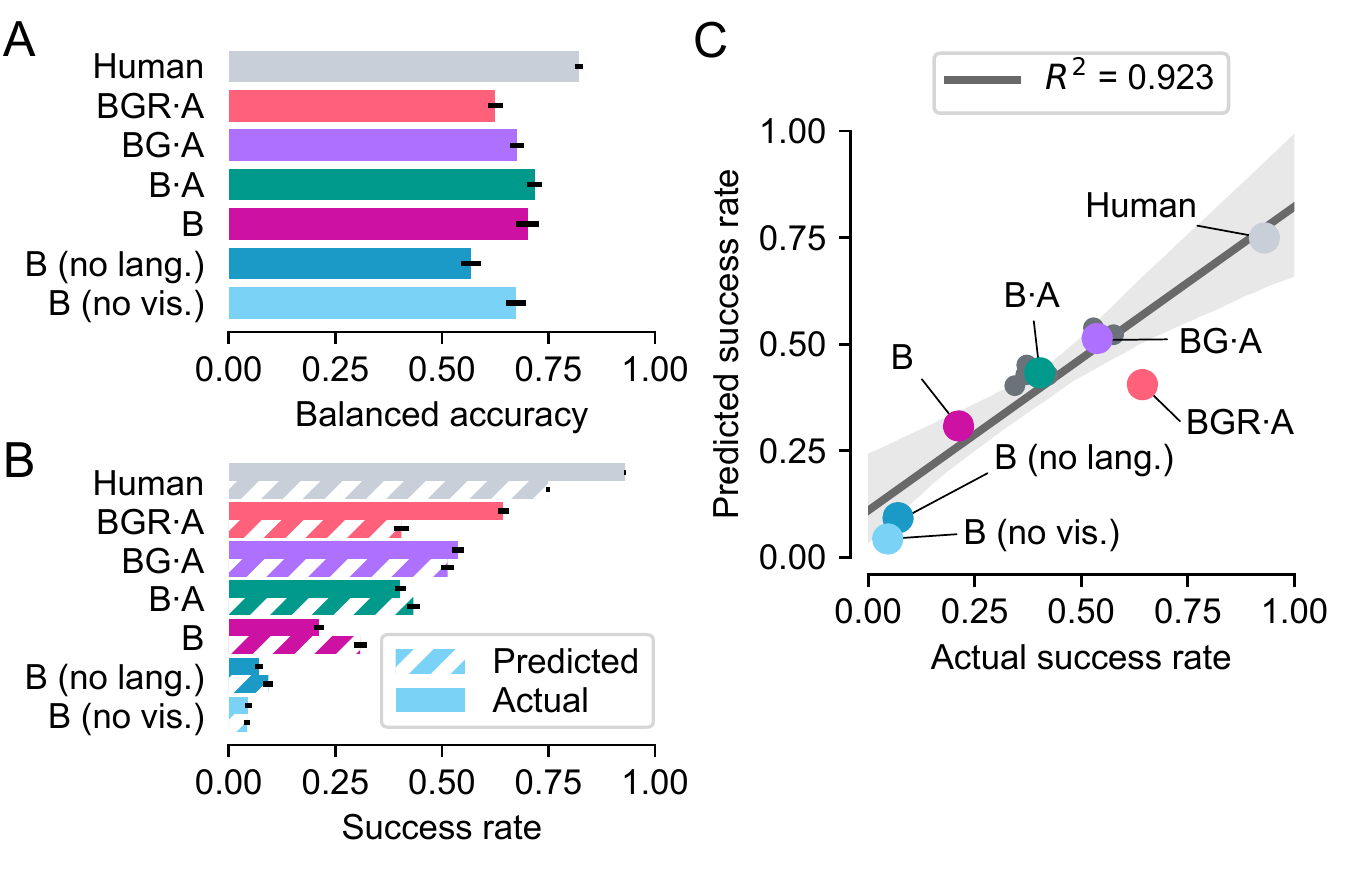}
    \caption[Evaluating agent performance using learned reward models]{\textbf{Evaluation Models.}
    {\bf A.} Balanced accuracy of the evaluation model computed for human validation episodes and for agent rollouts.
    {\bf B.} Actual and predicted success rates for instruction-following episodes across human and agent data. The evaluation model was trained on human data alone, so performance on agent data requires generalisation out of distribution.
    {\bf C.} Correlation between actual versus predicted success rates for ablations. Dark grey dots are ablations presented in Appendix~\ref{app:eval_models}.
    }
    \label{fig:eval_models}
\end{figure}

Our evaluation model consumes a video of the episode from the solver's perspective along with the language instruction emitted by the setter. To reduce the demand of processing whole episodes, the evaluation model processes observations with temporal striding, reducing the number of inputs seen in the episode. It assigns a probability to the episode's success ($y=1$) according to $\hat{y} = r_{\vtheta}(\mathbf{o}^{V}_{\leq T}, \mathbf{o}^{LI})$, where $T$ is the final time of the episode, given the video and language instruction, which we collectively denote as $\tau = [\mathbf{o}^{V}_{\leq T}, \mathbf{o}^{LI}]$ for convenience. 
The video is passed through a standard residual network~\citep{he2016deep}.
Language instructions are embedded and summed along the token dimension to produce a single summary vector.
The video and text representations are then concatenated and fed through a transformer, followed by an MLP and a logistic output unit.
The model was trained by minimising the evaluation loss, $\mathcal{L}^\textsc{ev}$, which was defined as the binary cross-entropy loss over the human data training set: 
\begin{align}
\mathcal{L}^\textsc{ev}(\vtheta) & = \expect{(y,\tau) \sim \mathcal{D}}{-y_i \ln  r_{\vtheta}(\tau_i) + (1-y_i) \ln(1-r_{\vtheta}(\tau_i))}.
\label{eq:eval_loss}
\end{align}
During training, we balanced the positive and negative examples within a batch. We regularised the model's representations via a full-episode variant of the language matching loss presented above in equation~\ref{eq:langmatch}, which we compute on the positive examples in the batch.
\begin{align}
\mathcal{L}^\textsc{elm}(\vtheta) & = 
-\frac{1}{B} \sum_{n=1}^{B}  \bigg [ \ln r_{\vtheta}(\mathbf{o}_{\leq T}^\textsc{v}, \mathbf{o}_{n}^\textsc{lo}) + \ln \big( 1 - r_{\vtheta}(\mathbf{o}_{\leq T}^\textsc{v}, \mathbf{o}_{\textsc{Shift}(n)}^\textsc{lo}) \big ) \bigg ].
\label{eq:elangmatch}
\end{align}
We optimised a convex combination of the $\mathcal{L}^\textsc{ev}$ and $\mathcal{L}^\textsc{elm}$ losses, where the scaling coefficient was chosen by hyperparameter search. The language matching loss was found to be crucial for best performance, contributing to a $3.38$\% improvement in validation score. See Appendix~\ref{app:eval_models} for details of model construction and training.

After training, we applied the model across our entire human validation dataset as well as the simulated rollouts for our $\mainagent$ agent and ablations (from Figure~\ref{fig:offline_human_eval}). Each episode was assigned a label using a threshold determined on a human validation dataset.
Figure~\ref{fig:eval_models}A shows the balanced accuracy of our model applied to the human data (grey, $82.17$\%), our $\mainagent$ agent (magenta,  $62.47$\%), and ablated variants. For comparison, additional human ratings achieved an average balanced accuracy of $90.24$\% across human data and rollouts from ablations. Figure~\ref{fig:eval_models}B compares the success rates for the agents as labelled by humans (solid bars; as in Figure~\ref{fig:offline_human_eval}A) and our evaluation model (dashed bars). The model is imperfect, but is clearly able to distinguish between better and worse performing models. 
Figure~\ref{fig:eval_models}C furthers this point; it shows a scatter of the actual and predicted success rates for the ablations presented in the main text, along with additional ablation agents detailed in Appendix~\ref{app:eval_models}. Our evaluation model agrees with human success evaluations for a wide range of agent configurations, giving a trend line close to unity and with an $R^{2}$ of $0.923$. 

Finally, we trained a variant of the evaluation model which was additionally able to predict the time at which success was achieved, as humans did when annotating videos. This model achieves similar performance to our transformer model with a validation score of $75.84$\% compared to the transformer model's validation score of $76.08$\%. Details for this model, as well as ablations, may be found in Appendix~\ref{app:eval_models}.

Our evaluation model robustly tracked the performance of agents across a vast spectrum of competence in the Playroom, from near-random agents up to human demonstrators.
The reasonable correspondence between machine-learned evaluation models and human judgement strongly suggests the possibility that further improvements to the agents described in this work can be evaluated readily with the same models. 
Future work will explore using these models to evaluate agents during training, select hyperparameters, and directly optimise agent parameters.

\section{Discussion \& Related Work}

\paragraph{Integrated AI Research.} Artificial intelligence research is mostly fragmented into specialized subfields, each with its own repertoire of domain-specific solutions. While the field has made much progress through this reductionist programme, we feel that integrated research is also required to understand how different elements of cognition functionally inter-relate. 
Here, we have taken steps to construct a more general programme of AI research that emphasises the holistic integration of perception, goal-directed, embodied control, and natural language processing, as has been advocated for previously \citep{mcclelland2019extending, lake2020wordmeaning}. 

Central to our integrated research methodology were ``interactions.'' Historically, Turing argued that a machine would be intelligent if it could interact indistinguishably from a human when paired with a human examiner, a protocol he called ``the imitation game,'' \citep{turing}. 
Such work provided clear inspiration to Winograd whose ``SHRDLU'' system comprised an embodied robot (a stationary manipulator) in a simple blocks world that could bidirectionally process limited language while engaging in interactions with a human \citep{winograd1972understanding}.
Winograd envisioned computers that are not ``tyrants,'' but rather machines that understand and assist us interactively, and it is this view that ultimately led him to advocate convergence between artificial intelligence and human-computer interaction \citep{winograd2006shifting}.  

\paragraph{Imitating Human Behaviour at Scale.} 
Our method for building integrated, interactive artificial intelligent rests on a base of imitation of human behaviour.
A central challenge for any attempt to learn models of human behaviour is a process to elicit and measure it. 
 In developmental psychology, several previous projects have attempted large-scale collection of human behavioural data. ~\cite{roy2006human} sought to record video and sound data from all rooms of a family home as a single child grew from birth to three years old. Following ~\cite{yoshida2008s}, ~\cite{sullivan2020saycam} recorded a large dataset of audio-visual experience from head-cameras on children aged 6-32 months.
 These studies have not so far attempted to use data to learn behavioural models. Further, it is at present intrinsically difficult to do so because algorithms and systems have not yet been developed that can perceive and understand the intentions of humans in a way that transfers across radical changes in embodiment, environment, and perspective \citep{stadie2017third, borsa2017observational,aytar2018playing, merel2017learning}.

Massive text corpora are a very different example of large-scale behavioural data that is relatively abundant and easy to collect~\citep{devlin2018bert,radford2019language,brown2020language}. Inter-person dialogue can be recorded in text form, which can capture a form of interactive and goal-driven behaviour. However, modelling text does not satisfy our goal of integrating perception, motor behaviour and language. Moreover, studying how to build agents that understand the ``grounding'' of language \citep{harnad1990symbol} within their sensorimotor embodiment is both fundamentally interesting \citep{hill2019environmental, mcclelland2019extending} and of obvious use for building robotic and other personal assistant artificial intelligences.
Nevertheless, we have observed dramatic progress in artificial intelligence in the language domain, which has been made possible by increasing model and dataset size, the latter made possible by the vast quantity of text available on the internet. While these two ingredients -- model and dataset size -- may not constitute a complete recipe for creating generally intelligent agents, they have proven sufficient to produce sometimes astonishing models \citep{brown2020language}. In this work we have focused on a domain yet to profit from this approach, \emph{embodied, interactive agents}, where natural language, complex motor control, and multi-modal sensory information come together.
A hurdle for this study is that there is no equivalent to a large and publicly available text dataset that can be applied directly to train models.

Computer games provide an alternative possibility for collecting large-scale interactive behaviour. The multi-player Starcraft gameplay data collected by~\cite{vinyals2019grandmaster} is sufficiently rich to produce interesting interactive agents. However, even the most complex and realistic computer games typically make a major simplifying assumption: that there is a  single well-defined objective designed by the game creator, relative to which performance (winning or losing) can be measured unambiguously. 
Our strategies to overcome the absence of such a metric when modelling human behaviour are a key contribution of this work. 

\paragraph{Language, Interactions, and Robotics.} Recent work in robotics demonstrated the possibility of conditioning simulated robotic manipulators with natural language instructions \citep{lynch2020grounding}. Other work on language and interaction based in 3D simulated environments has focused on embodied instruction-following~\citep{hill2019robust}, navigation~\citep{DBLP:journals/corr/abs-1711-07280} or question-answering~\citep{das2018embodied}. These approaches share commonalities with our work here but also present important differences. 
First, in prior work, language has typically described behaviour observed in short, few second windows. (By comparison, interactions in the Playroom can last upwards of a minute.) 
Second, prior work has largely focused on comparatively constrained sets of behaviours, involving uncluttered environments with few objects to manipulate \citep{lynch2020grounding, hill2019environmental}, or has studied navigation absent of environment manipulation altogether \citep{DBLP:journals/corr/abs-1711-07280}. 
Third, our agents not only interpret language but also produce language output. 
While producing context-specific, embodied language is notable in its own right, it has also presented many practical difficulties that were not faced in previous work (including the problem of making language congruent with perception and learning from sparse language output data).
Moving beyond many of these limitations, \citet{hu2019hierarchical} studied a strategy game played by two humans, one in the role of an ``instructor'' who directed strategy via natural language commands, and the second an ``executor'', who carried them out. Recordings of the commands along with game histories were used to train a hierarchical agent that generated intermediate plans in natural language.

In some sense, robotics is the ultimate integrated, interactive research platform (see e.g. ~\cite{tellex2011understanding} for a pioneering study of language understanding in robotics). Ultimately, what we wished to accomplish here \emph{in simulation} was to build a research program to study a way to build intelligent agents \emph{in general}. Compared to a typical robotics platform, our virtual environment allowed for faster iteration and few hardware challenges, making it an ideal place to start this research. An obvious next step is to take the lessons learned from our proposed process model of building AI, and apply them to the real world. 

\paragraph{Imitation Among Humans.} Social learning, imitation, and mimicry are found throughout the animal kingdom~\citep{heyes1996social,laland2004social,byrne2009animal}, and human infants are intrinsically motivated to imitate. They imitate the phonemes, words, and grammatical structures of the language that they encounter in their environment~\citep{chomsky1959chomsky}, as well as observed interactions with objects in their environment~\citep{heyes1996social}. 
Infants appear to leverage sophisticated and abstract capacities for imitation for much the same reason we have proposed here: to bootstrap from other agents' behaviour to acquire basic competence. 
``Program-level imitation,'' where an individual recognises the gist of a complex task, shifts the burden of learning from \emph{tabula rasa} exploration to refinement through practice \citep{byrne1998learning, byrne2009animal}. 

\paragraph{Challenges of the Approach}
The approach to building agents that we have pursued so far has relied substantially on imitation learning techniques to approximate the distribution of human behaviour in the Playroom. We have argued that imitation learning jumpstarts initial competency for engaging in human interactions. However, imitation learning has its own limitations for producing ultimately intelligent, interactive agents.
On its own, imitation learning does not distinguish between human skill and human error, what is desirable or what is counterproductive. The full distribution of behaviour in our dataset includes, for example, mispellings, clumsiness, and lapses of attention. Eliminating these errors and producing agents with mastery and grace in their environment will require additional techniques, including adaptation from human evaluative feedback. To record sufficiently diverse behaviour, we have ``gamified'' human-human interaction via the instrument of language games. 
These language games have helped generate data targeting basic and desirable capabilities for agents, but we believe that it is through interacting with and learning directly from humans, not merely imitating pre-existing human interaction datasets, that we can produce broadly capable agents.
To go beyond competence within somewhat stereotyped scenarios toward interactive agents that can actively acquire and creatively recombine knowledge to cope with new challenges may require as yet unknown methods for knowledge representation and credit assignment, or, failing this, larger scales of data.
Multiple avenues, including understanding more deeply the mechanisms of creative, knowledge-rich thought, or transferring knowledge from large, real world datasets, may offer a way forward.

\section{Conclusion}
In this work, we sought to build embodied artificial agents that interact with their world, with each other, and with us. The agents could perceive and manipulate their environment, produce language, and react capably when given general requests and instructions by humans. They also generalised and transferred knowledge to new tasks. 
Although the agents undertook tasks without easily programmed success criteria, we were able to develop a variety of robust and effective strategies for evaluating their performance.
While the agents' behaviours were not perfect, even when they failed to satisfy instructions, they routinely undertook actions that seemed to reflect some understanding of the original instruction, thus exhibiting behaviour primed to profit from interactive feedback. 

Ultimately, we endeavour to create agents that assist us in our daily lives. 
Therefore, they will need to understand and learn from us while we interact with them.
If the agents introduced into human environments are not reasonably capable from the start, we believe there will be little incentive to engage with them subsequently.
Here, we have made some material progress by creating agents that may be interesting enough to entertain continued interaction, and, in a virtuous circle, it is this interaction that promises to select for increasingly intelligent, useful agents.

\section{Authors \& Contributions}
\label{sec:authors}

\textbf{Josh Abramson} contributed to agent development, imitation learning, data and tasks, running and analysis of experiments, engineering infrastructure, writing, and as the technical lead. \\
\textbf{Arun Ahuja} contributed to agent development, imitation learning, data and tasks, running and analysis of experiments, engineering infrastructure, writing, and as a sub-effort lead for imitation. \\
\textbf{Iain Barr} contributed to running and analysis of experiments, and engineering infrastructure. \\
\textbf{Arthur Brussee} contributed to environment development. \\
\textbf{Federico Carnevale} contributed to imitation learning, running and analysis of experiments, and writing. \\
\textbf{Mary Cassin} contributed to environment development. \\
\textbf{Rachita Chhaparia} contributed to environment development. \\
\textbf{Stephen Clark} contributed to environment development and data and tasks. \\
\textbf{Bogdan Damoc} contributed to environment development \\
\textbf{Andrew Dudzik} contributed to engineering infrastructure and running and analysis of experiments. \\
\textbf{Petko Georgiev} contributed to agent development, imitation learning, data and tasks, running and analysis of experiments, engineering infrastructure, writing, and as a sub-effort lead for agent development. \\
\textbf{Aurelia Guy} contributed to agent development, imitation learning, data and tasks, running and analysis of experiments, engineering infrastructure, and writing. \\
\textbf{Tim Harley} contributed to data and tasks and engineering infrastructure. \\
\textbf{Felix Hill} contributed to data and tasks, environment development, writing, and as a sub-effort lead for environment development. \\
\textbf{Alden Hung} contributed to agent development, imitation learning, data and tasks, running and analysis of experiments, engineering infrastructure, writing, and as a sub-effort lead for imitation learning. \\
\textbf{Zachary Kenton} contributed to evaluation model development and running and analysis of experiments. \\
\textbf{Jessica Landon} contributed to evaluation model development, engineering infrastructure, running and analysis of experiments, and writing. \\
\textbf{Timothy Lillicrap} contributed to agent development, imitation learning, data and tasks, environment development, evaluation model development, writing, and as an effort lead.
\textbf{Kory Mathewson} contributed to agent development. \\
\textbf{So{\v{n}}a Mokr{{\'a}}} contributed to agent development, and running and analysis of experiments. \\
\textbf{Alistair Muldal} contributed to data and tasks, environment development, evaluation model development, writing, and as a sub-effort lead for evaluation model development. \\
\textbf{Adam Santoro} contributed to agent development, data and tasks, imitation learning, running and analysis of experiments, writing, and as a sub-effort lead for agent development. \\
\textbf{Nikolay Savinov} contributed to evaluation model development and running and analysis of experiments. \\
\textbf{Vikrant Varma} contributed to evaluation model development and running and analysis of experiments. \\
\textbf{Greg Wayne} contributed to agent development, imitation learning, data and tasks, evaluation model development, writing, and as an effort lead. \\
\textbf{Duncan Williams} contributed to engineering infrastructur. \\
\textbf{Nathaniel Wong} contributed to environment development and as a sub-effort lead for environment development. \\
\textbf{Chen Yan} contributed to agent development, running and analysis of experiments, and writing. \\
\textbf{Rui Zhu} contributed to agent development, running and analysis of experiments, and engineering infrastructure. \\

\vspace{1mm}
\noindent
{\bf Corresponding Authors:} \\
Greg Wayne (gregwayne@google.com) \& Timothy Lillicrap (countzero@google.com)

\section{Acknowledgments} 
The authors would like to thank Jay McClelland for formative initial discussions; Paola Jouyaux, Vicky Holgate, Esme Sutherland Robson, Guy Scully, and Alex Goldin for organisational support; Duncan Williams and Rachita Chhaparia for infrastructure support; Jason Sanmiya, Sarah York, Dario de Cesare, Charlie Deck, Marcus Mainright for support in building or using the Playroom; Jan Leike, Richard Ngo, Miljan Martic, Remi Lam, Lucas Smaira, Charlie Deck, Daan Wierstra, Matt Botvinick, Nando de Freitas, Adam Marblestone, Koray Kavukcuoglu, Demis Hassabis, Karol Gregor, Danilo J. Rezende, and others for important discussions.

\bibliographystyle{apalike}
\bibliography{references}

\begin{thebibliography}{}

\bibitem[Adam et~al., 1902]{adam1902republic}
Adam, J. et~al. (1902).
\newblock {\em The Republic of Plato}.
\newblock University Press.

\bibitem[Anderson et~al., 2017]{DBLP:journals/corr/abs-1711-07280}
Anderson, P., Wu, Q., Teney, D., Bruce, J., Johnson, M., S{\"{u}}nderhauf, N.,
  Reid, I.~D., Gould, S., and van~den Hengel, A. (2017).
\newblock Vision-and-language navigation: Interpreting visually-grounded
  navigation instructions in real environments.
\newblock {\em CoRR}, abs/1711.07280.

\bibitem[Aytar et~al., 2018]{aytar2018playing}
Aytar, Y., Pfaff, T., Budden, D., Paine, T., Wang, Z., and de~Freitas, N.
  (2018).
\newblock Playing hard exploration games by watching youtube.
\newblock In {\em Advances in Neural Information Processing Systems}, pages
  2930--2941.

\bibitem[Borsa et~al., 2017]{borsa2017observational}
Borsa, D., Piot, B., Munos, R., and Pietquin, O. (2017).
\newblock Observational learning by reinforcement learning.
\newblock {\em arXiv preprint arXiv:1706.06617}.

\bibitem[Branwen, 2018]{gwern2018}
Branwen, G. (2018).
\newblock Tool {AI}.
\newblock \url{https://www.gwern.net/Tool-AI}.

\bibitem[Brown et~al., 2020]{brown2020language}
Brown, T.~B., Mann, B., Ryder, N., Subbiah, M., Kaplan, J., Dhariwal, P.,
  Neelakantan, A., Shyam, P., Sastry, G., Askell, A., et~al. (2020).
\newblock Language models are few-shot learners.
\newblock {\em arXiv preprint arXiv:2005.14165}.

\bibitem[Byrne, 2009]{byrne2009animal}
Byrne, R.~W. (2009).
\newblock Animal imitation.
\newblock {\em Current Biology}, 19(3):R111--R114.

\bibitem[Byrne and Russon, 1998]{byrne1998learning}
Byrne, R.~W. and Russon, A.~E. (1998).
\newblock Learning by imitation: A hierarchical approach.
\newblock {\em Behavioral and brain sciences}, 21(5):667--684.

\bibitem[Cabi et~al., 2019]{cabi2019scaling}
Cabi, S., G{\'o}mez~Colmenarejo, S., Novikov, A., Konyushkova, K., Reed, S.,
  Jeong, R., Zolna, K., Aytar, Y., Budden, D., Vecerik, M., et~al. (2019).
\newblock Scaling data-driven robotics with reward sketching and batch
  reinforcement learning.
\newblock {\em arXiv}, pages arXiv--1909.

\bibitem[Card et~al., 1983]{card1983psychology}
Card, S.~K., Moran, T.~P., and Newell, A. (1983).
\newblock {\em The psychology of human-computer interaction}.
\newblock CRC Press.

\bibitem[Chollet, 2019]{chollet2019measure}
Chollet, F. (2019).
\newblock On the measure of intelligence.
\newblock {\em arXiv preprint arXiv:1911.01547}.

\bibitem[Chomsky, 1959]{chomsky1959chomsky}
Chomsky, N. (1959).
\newblock Chomsky, n. 1959. a review of bf skinner’s verbal behavior.
  language, 35 (1), 26--58.

\bibitem[Chopra et~al., 2005]{chopra2005learning}
Chopra, S., Hadsell, R., and LeCun, Y. (2005).
\newblock Learning a similarity metric discriminatively, with application to
  face verification.
\newblock In {\em Computer Vision and Pattern Recognition, 2005. CVPR 2005.
  IEEE Computer Society Conference on}, volume~1, pages 539--546. IEEE.

\bibitem[Christiano et~al., 2017]{christiano2017deep}
Christiano, P.~F., Leike, J., Brown, T., Martic, M., Legg, S., and Amodei, D.
  (2017).
\newblock Deep reinforcement learning from human preferences.
\newblock In {\em Advances in Neural Information Processing Systems}, pages
  4299--4307.

\bibitem[Cubuk et~al., 2020]{cubuk2020randaugment}
Cubuk, E.~D., Zoph, B., Shlens, J., and Le, Q.~V. (2020).
\newblock Randaugment: Practical automated data augmentation with a reduced
  search space.
\newblock In {\em Proceedings of the IEEE/CVF Conference on Computer Vision and
  Pattern Recognition Workshops}, pages 702--703.

\bibitem[Das et~al., 2018]{das2018embodied}
Das, A., Datta, S., Gkioxari, G., Lee, S., Parikh, D., and Batra, D. (2018).
\newblock Embodied question answering.
\newblock In {\em Proceedings of the IEEE Conference on Computer Vision and
  Pattern Recognition Workshops}, pages 2054--2063.

\bibitem[Devlin et~al., 2018]{devlin2018bert}
Devlin, J., Chang, M.-W., Lee, K., and Toutanova, K. (2018).
\newblock Bert: Pre-training of deep bidirectional transformers for language
  understanding.
\newblock {\em arXiv preprint arXiv:1810.04805}.

\bibitem[Dunbar, 1993]{dunbar1993coevolution}
Dunbar, R.~I. (1993).
\newblock Coevolution of neocortical size, group size and language in humans.
\newblock {\em Behavioral and brain sciences}, 16(4):681--694.

\bibitem[Duncan, 2010]{duncan2010intelligence}
Duncan, J. (2010).
\newblock {\em How intelligence happens}.
\newblock Yale University Press.

\bibitem[Espeholt et~al., 2018]{espeholt2018impala}
Espeholt, L., Soyer, H., Munos, R., Simonyan, K., Mnih, V., Ward, T., Doron,
  Y., Firoiu, V., Harley, T., Dunning, I., et~al. (2018).
\newblock Impala: Scalable distributed deep-rl with importance weighted
  actor-learner architectures.
\newblock {\em arXiv preprint arXiv:1802.01561}.

\bibitem[Finn et~al., 2016]{finn2016connection}
Finn, C., Christiano, P., Abbeel, P., and Levine, S. (2016).
\newblock A connection between generative adversarial networks, inverse
  reinforcement learning, and energy-based models.
\newblock {\em arXiv preprint arXiv:1611.03852}.

\bibitem[Galashov et~al., 2019]{galashov2019information}
Galashov, A., Jayakumar, S.~M., Hasenclever, L., Tirumala, D., Schwarz, J.,
  Desjardins, G., Czarnecki, W.~M., Teh, Y.~W., Pascanu, R., and Heess, N.
  (2019).
\newblock Information asymmetry in kl-regularized rl.
\newblock {\em arXiv preprint arXiv:1905.01240}.

\bibitem[Ghasemipour et~al., 2020]{ghasemipour2020divergence}
Ghasemipour, S. K.~S., Zemel, R., and Gu, S. (2020).
\newblock A divergence minimization perspective on imitation learning methods.
\newblock In {\em Conference on Robot Learning}, pages 1259--1277. PMLR.

\bibitem[Girshick, 2015]{girshick2015fast}
Girshick, R. (2015).
\newblock Fast r-cnn.
\newblock In {\em Proceedings of the IEEE international conference on computer
  vision}, pages 1440--1448.

\bibitem[Goodfellow et~al., 2014]{goodfellow2014generative}
Goodfellow, I., Pouget-Abadie, J., Mirza, M., Xu, B., Warde-Farley, D., Ozair,
  S., Courville, A., and Bengio, Y. (2014).
\newblock Generative adversarial nets.
\newblock In {\em Advances in neural information processing systems}, pages
  2672--2680.

\bibitem[Graves, 2013]{graves2013generating}
Graves, A. (2013).
\newblock Generating sequences with recurrent neural networks.
\newblock {\em arXiv preprint arXiv:1308.0850}.

\bibitem[Guo et~al., 2020]{Wiki40B}
Guo, M., Dai, Z., Vrandecic, D., and Al-Rfou, R. (2020).
\newblock Wiki-40b: Multilingual language model dataset.
\newblock In {\em LREC 2020}.

\bibitem[Gutmann and Hyv{\"a}rinen, 2010]{gutmann2010noise}
Gutmann, M. and Hyv{\"a}rinen, A. (2010).
\newblock Noise-contrastive estimation: A new estimation principle for
  unnormalized statistical models.
\newblock In {\em Proceedings of the Thirteenth International Conference on
  Artificial Intelligence and Statistics}, pages 297--304.

\bibitem[Harnad, 1990]{harnad1990symbol}
Harnad, S. (1990).
\newblock The symbol grounding problem.
\newblock {\em Physica D: Nonlinear Phenomena}, 42(1-3):335--346.

\bibitem[He et~al., 2017]{he2017mask}
He, K., Gkioxari, G., Doll{\'a}r, P., and Girshick, R. (2017).
\newblock Mask r-cnn.
\newblock In {\em Proceedings of the IEEE international conference on computer
  vision}, pages 2961--2969.

\bibitem[He et~al., 2016]{he2016deep}
He, K., Zhang, X., Ren, S., and Sun, J. (2016).
\newblock Deep residual learning for image recognition.
\newblock In {\em Proceedings of the IEEE conference on computer vision and
  pattern recognition}, pages 770--778.

\bibitem[H{\'e}naff et~al., 2019]{henaff2019data}
H{\'e}naff, O.~J., Srinivas, A., De~Fauw, J., Razavi, A., Doersch, C., Eslami,
  S., and Oord, A. v.~d. (2019).
\newblock Data-efficient image recognition with contrastive predictive coding.
\newblock {\em arXiv preprint arXiv:1905.09272}.

\bibitem[Hennigan et~al., 2020]{haiku2020github}
Hennigan, T., Cai, T., Norman, T., and Babuschkin, I. (2020).
\newblock {H}aiku: {S}onnet for {JAX}.

\bibitem[Hermann et~al., 2017]{hermann2017grounded}
Hermann, K.~M., Hill, F., Green, S., Wang, F., Faulkner, R., Soyer, H.,
  Szepesvari, D., Czarnecki, W.~M., Jaderberg, M., Teplyashin, D., et~al.
  (2017).
\newblock Grounded language learning in a simulated 3d world.
\newblock {\em arXiv preprint arXiv:1706.06551}.

\bibitem[Heyes and Galef~Jr, 1996]{heyes1996social}
Heyes, C.~M. and Galef~Jr, B.~G. (1996).
\newblock {\em Social learning in animals: the roots of culture}.
\newblock Elsevier.

\bibitem[Hill et~al., 2019a]{hill2019environmental}
Hill, F., Lampinen, A., Schneider, R., Clark, S., Botvinick, M., McClelland,
  J.~L., and Santoro, A. (2019a).
\newblock Environmental drivers of systematicity and generalization in a
  situated agent.
\newblock In {\em International Conference on Learning Representations}.

\bibitem[Hill et~al., 2019b]{hill2019robust}
Hill, F., Mokra, S., Wong, N., and Harley, T. (2019b).
\newblock Robust instruction-following in a situated agent via
  transfer-learning from text.
\newblock {\em OpenReview}.

\bibitem[Ho and Ermon, 2016]{ho2016generative}
Ho, J. and Ermon, S. (2016).
\newblock Generative adversarial imitation learning.
\newblock In {\em Advances in neural information processing systems}, pages
  4565--4573.

\bibitem[Hu et~al., 2019]{hu2019hierarchical}
Hu, H., Yarats, D., Gong, Q., Tian, Y., and Lewis, M. (2019).
\newblock Hierarchical decision making by generating and following natural
  language instructions.
\newblock In {\em Advances in neural information processing systems}, pages
  10025--10034.

\bibitem[Jouppi et~al., 2017]{jouppi2017datacenter}
Jouppi, N.~P., Young, C., Patil, N., Patterson, D., Agrawal, G., Bajwa, R.,
  Bates, S., Bhatia, S., Boden, N., Borchers, A., et~al. (2017).
\newblock In-datacenter performance analysis of a tensor processing unit.
\newblock In {\em Proceedings of the 44th Annual International Symposium on
  Computer Architecture}, pages 1--12.

\bibitem[Kakade et~al., 2003]{kakade2003sample}
Kakade, S.~M. et~al. (2003).
\newblock {\em On the sample complexity of reinforcement learning}.
\newblock PhD thesis, University of London London, England.

\bibitem[Kaplan et~al., 2020]{kaplan2020scaling}
Kaplan, J., McCandlish, S., Henighan, T., Brown, T.~B., Chess, B., Child, R.,
  Gray, S., Radford, A., Wu, J., and Amodei, D. (2020).
\newblock Scaling laws for neural language models.
\newblock {\em arXiv preprint arXiv:2001.08361}.

\bibitem[Kingma and Ba, 2014]{kingma2014adam}
Kingma, D.~P. and Ba, J. (2014).
\newblock Adam: A method for stochastic optimization.
\newblock {\em arXiv preprint arXiv:1412.6980}.

\bibitem[Lake and Murphy, 2020]{lake2020wordmeaning}
Lake, B.~M. and Murphy, G.~L. (2020).
\newblock Word meaning in minds and machines.
\newblock {\em arXiv preprint arXiv:2008.01766}.

\bibitem[Laland, 2004]{laland2004social}
Laland, K.~N. (2004).
\newblock Social learning strategies.
\newblock {\em Animal Learning \& Behavior}, 32(1):4--14.

\bibitem[Li et~al., 2017]{li2017infogail}
Li, Y., Song, J., and Ermon, S. (2017).
\newblock Infogail: Interpretable imitation learning from visual
  demonstrations.
\newblock In {\em Advances in Neural Information Processing Systems}, pages
  3812--3822.

\bibitem[Lin et~al., 2019]{lin2019tsm}
Lin, J., Gan, C., and Han, S. (2019).
\newblock Tsm: Temporal shift module for efficient video understanding.
\newblock In {\em Proceedings of the IEEE International Conference on Computer
  Vision}, pages 7083--7093.

\bibitem[Lynch and Sermanet, 2020]{lynch2020grounding}
Lynch, C. and Sermanet, P. (2020).
\newblock Grounding language in play.
\newblock {\em arXiv preprint arXiv:2005.07648}.

\bibitem[McClelland et~al., 2019]{mcclelland2019extending}
McClelland, J.~L., Hill, F., Rudolph, M., Baldridge, J., and Sch{\"u}tze, H.
  (2019).
\newblock Extending machine language models toward human-level language
  understanding.
\newblock {\em arXiv preprint arXiv:1912.05877}.

\bibitem[Merel et~al., 2017]{merel2017learning}
Merel, J., Tassa, Y., TB, D., Srinivasan, S., Lemmon, J., Wang, Z., Wayne, G.,
  and Heess, N. (2017).
\newblock Learning human behaviors from motion capture by adversarial
  imitation.
\newblock {\em arXiv preprint arXiv:1707.02201}.

\bibitem[Mnih et~al., 2016]{mnih2016asynchronous}
Mnih, V., Badia, A.~P., Mirza, M., Graves, A., Lillicrap, T., Harley, T.,
  Silver, D., and Kavukcuoglu, K. (2016).
\newblock Asynchronous methods for deep reinforcement learning.
\newblock In {\em International conference on machine learning}, pages
  1928--1937.

\bibitem[Osa et~al., 2018]{osa2018algorithmic}
Osa, T., Pajarinen, J., Neumann, G., Bagnell, J.~A., Abbeel, P., and Peters, J.
  (2018).
\newblock An algorithmic perspective on imitation learning.
\newblock {\em arXiv preprint arXiv:1811.06711}.

\bibitem[Perez et~al., 2018]{perez2018film}
Perez, E., Strub, F., de~Vries, H., Dumoulin, V., and Courville, A.~C. (2018).
\newblock Film: Visual reasoning with a general conditioning layer.
\newblock In {\em AAAI}.

\bibitem[Pomerleau, 1989]{pomerleau1989alvinn}
Pomerleau, D.~A. (1989).
\newblock Alvinn: An autonomous land vehicle in a neural network.
\newblock In {\em Advances in neural information processing systems}, pages
  305--313.

\bibitem[Radford et~al., 2019]{radford2019language}
Radford, A., Wu, J., Child, R., Luan, D., Amodei, D., and Sutskever, I. (2019).
\newblock Language models are unsupervised multitask learners.
\newblock {\em OpenAI Blog}, 1(8):9.

\bibitem[Ross et~al., 2011]{ross2011reduction}
Ross, S., Gordon, G., and Bagnell, D. (2011).
\newblock A reduction of imitation learning and structured prediction to
  no-regret online learning.
\newblock In {\em Proceedings of the fourteenth international conference on
  artificial intelligence and statistics}, pages 627--635.

\bibitem[Roy et~al., 2006]{roy2006human}
Roy, D., Patel, R., DeCamp, P., Kubat, R., Fleischman, M., Roy, B., Mavridis,
  N., Tellex, S., Salata, A., Guinness, J., et~al. (2006).
\newblock The human speechome project.
\newblock In {\em International Workshop on Emergence and Evolution of
  Linguistic Communication}, pages 192--196. Springer.

\bibitem[Schaal, 1999]{schaal1999imitation}
Schaal, S. (1999).
\newblock Is imitation learning the route to humanoid robots?
\newblock {\em Trends in cognitive sciences}, 3(6):233--242.

\bibitem[Shannon, 1951]{shannon1951prediction}
Shannon, C.~E. (1951).
\newblock Prediction and entropy of printed {E}nglish.
\newblock {\em Bell system technical journal}, 30(1):50--64.

\bibitem[Shaw et~al., 2018]{shaw2018self}
Shaw, P., Uszkoreit, J., and Vaswani, A. (2018).
\newblock Self-attention with relative position representations.
\newblock {\em arXiv preprint arXiv:1803.02155}.

\bibitem[Silver et~al., 2016]{silver2016mastering}
Silver, D., Huang, A., Maddison, C.~J., Guez, A., Sifre, L., Van Den~Driessche,
  G., Schrittwieser, J., Antonoglou, I., Panneershelvam, V., Lanctot, M.,
  et~al. (2016).
\newblock Mastering the game of go with deep neural networks and tree search.
\newblock {\em nature}, 529(7587):484--489.

\bibitem[Stadie et~al., 2017]{stadie2017third}
Stadie, B.~C., Abbeel, P., and Sutskever, I. (2017).
\newblock Third-person imitation learning.
\newblock {\em arXiv preprint arXiv:1703.01703}.

\bibitem[Sullivan et~al., 2020]{sullivan2020saycam}
Sullivan, J., Mei, M., Perfors, A., Wojcik, E.~H., and Frank, M.~C. (2020).
\newblock Saycam: A large, longitudinal audiovisual dataset recorded from the
  infant’s perspective.
\newblock {\em PsyArXiv}.

\bibitem[Tellex et~al., 2011]{tellex2011understanding}
Tellex, S.~A., Kollar, T.~F., Dickerson, S.~R., Walter, M.~R., Banerjee, A.,
  Teller, S., and Roy, N. (2011).
\newblock Understanding natural language commands for robotic navigation and
  mobile manipulation.
\newblock {\em AAAI Publications}.

\bibitem[Tomasello, 2010]{tomasello2010origins}
Tomasello, M. (2010).
\newblock {\em Origins of human communication}.
\newblock MIT press.

\bibitem[Turing, 1950]{turing}
Turing, A.~M. (1950).
\newblock Computing machinery and intelligence.
\newblock {\em Mind}, LIX(236):433–460.

\bibitem[{van den Oord} et~al., 2018]{oord2018representation}
{van den Oord}, A., Li, Y., and Vinyals, O. (2018).
\newblock Representation learning with contrastive predictive coding.
\newblock {\em arXiv preprint arXiv:1807.03748}.

\bibitem[Vaswani et~al., 2017]{vaswani2017attention}
Vaswani, A., Shazeer, N., Parmar, N., Uszkoreit, J., Jones, L., Gomez, A.~N.,
  Kaiser, {\L}., and Polosukhin, I. (2017).
\newblock Attention is all you need.
\newblock In {\em Advances in neural information processing systems}, pages
  5998--6008.

\bibitem[Vinyals et~al., 2019]{vinyals2019grandmaster}
Vinyals, O., Babuschkin, I., Czarnecki, W.~M., Mathieu, M., Dudzik, A., Chung,
  J., Choi, D.~H., Powell, R., Ewalds, T., Georgiev, P., et~al. (2019).
\newblock Grandmaster level in starcraft ii using multi-agent reinforcement
  learning.
\newblock {\em Nature}, 575(7782):350--354.

\bibitem[Ward et~al., 2020]{ward2020using}
Ward, T., Bolt, A., Hemmings, N., Carter, S., Sanchez, M., Barreira, R., Noury,
  S., Anderson, K., Lemmon, J., Coe, J., et~al. (2020).
\newblock Using unity to help solve intelligence.
\newblock {\em arXiv preprint arXiv:2011.09294}.

\bibitem[Winograd, 1972]{winograd1972understanding}
Winograd, T. (1972).
\newblock Understanding natural language.
\newblock {\em Cognitive psychology}, 3(1):1--191.

\bibitem[Winograd, 2006]{winograd2006shifting}
Winograd, T. (2006).
\newblock Shifting viewpoints: Artificial intelligence and human--computer
  interaction.
\newblock {\em Artificial intelligence}, 170(18):1256--1258.

\bibitem[Wittgenstein, 1953]{wittgenstein1953philosophische}
Wittgenstein, L. (1953).
\newblock {\em Philosophische Untersuchungen, von Ludwig
  Wittgenstein.-Philosophical investigations, by Ludwig Wittgenstein.
  Translated by GEM Anscombe}.
\newblock B. Blackwell.

\bibitem[Yoshida and Smith, 2008]{yoshida2008s}
Yoshida, H. and Smith, L.~B. (2008).
\newblock What's in view for toddlers? using a head camera to study visual
  experience.
\newblock {\em Infancy}, 13(3):229--248.

\bibitem[Ziebart, 2010]{ziebart2010modeling}
Ziebart, B.~D. (2010).
\newblock Modeling purposeful adaptive behavior with the principle of maximum
  causal entropy.
\newblock {\em Thesis for Carnegie Mellon University}.

\bibitem[Zolna et~al., 2019]{zolna2019task}
Zolna, K., Reed, S., Novikov, A., Colmenarej, S.~G., Budden, D., Cabi, S.,
  Denil, M., de~Freitas, N., and Wang, Z. (2019).
\newblock Task-relevant adversarial imitation learning.
\newblock {\em arXiv preprint arXiv:1910.01077}.

\end{thebibliography}

\clearpage
\newpage

\setcounter{section}{0}

\newpage
\begin{center}
\LARGE{Appendix for Imitating Interactive Intelligence} \\
\large{Interactive Agents Group \\ DeepMind} \\
\vspace{1cm}
\end{center}

\section{Playroom Environment Description}
\label{sec:playroomenv}

The Playroom environment is a configurable room developed in the Unity game engine \citep{ward2020using}. As described below, many aspects of the room are randomised in each episode.  
\begin{table}[ht]
\small
\begin{tabular}{llll}
\toprule
Small objects   & Furniture objects & Object colours & Wall and ceiling colours \\ [-4pt]
\bottomrule
basketball    & arm chair       & aquamarine  & light red  \\ [-6pt]
book          & book case       & blue        & light blue            \\[-6pt]
cushion       & chair           & green       & light yellow          \\[-6pt]
football      & chest           & magenta     & light green           \\[-6pt]
hairdryer     & dining table    & orange      & light purple          \\[-6pt]
headphones    & stool           & purple      & light orange          \\[-6pt]
mug           & wardrobe        & pink        & light aquamarine      \\[-6pt]
picture frame & bed             & red         & light magenta         \\[-6pt]
potted plant  & shelf           & white       &                       \\[-6pt]
rubber duck   & storage box     & yellow      &                       \\[-6pt]
table lamp    &                 &             &                       \\[-6pt]
teddy         &                 &             &                       \\[-6pt]
boat          &                 &             &                       \\[-6pt]
bus           &                 &             &                       \\[-6pt]
car           &                 &             &                       \\[-6pt]
carriage      &                 &             &                       \\[-6pt]
helicopter    &                 &             &                       \\[-6pt]
keyboard      &                 &             &                       \\[-6pt]
plane         &                 &             &                       \\[-6pt]
robot         &                 &             &                       \\[-6pt]
rocket        &                 &             &                       \\[-6pt]
train         &                 &             &                       \\[-6pt]
racket        &                 &             &                       \\ [-6pt]
\end{tabular}
\caption{The total repository of objects and colours. In each episode, small objects and furniture are objects are sampled from these sets and object colours are applied to them at random as well as one of three sizes. The colours of the walls and ceilings are sampled from a list of lighter shades.}
\label{tab:playroom_objects}
\end{table}

\subsection{Objects and furniture in the Playroom}

\begin{figure}[ht]
    \centering
    \includegraphics[width=\textwidth]{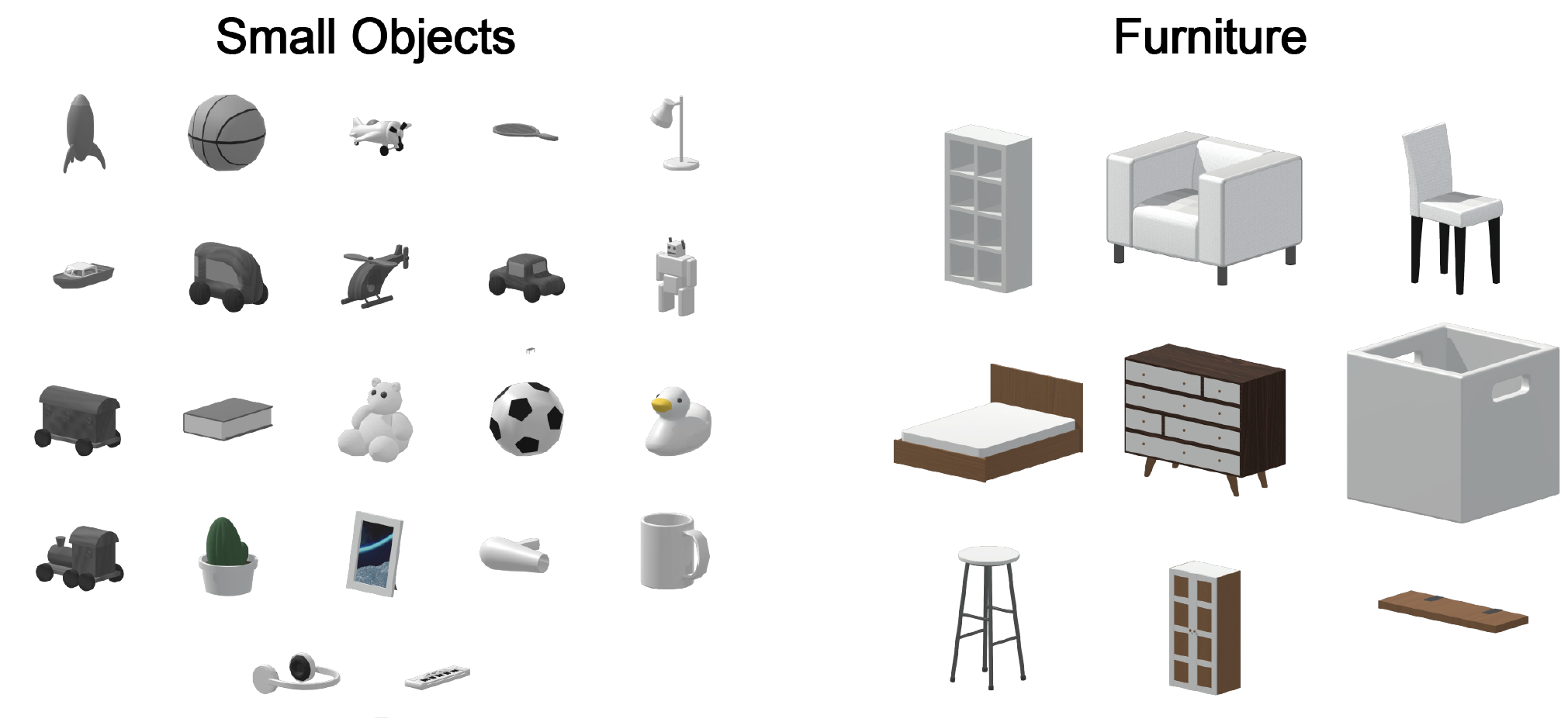}
    \caption{Repository of small objects and furniture in the Playroom environment. The colours of the objects are chosen at random from the list described in Table~\ref{tab:playroom_objects}.}
    \label{fig:playroom_objects}
\end{figure}

Inside the Playroom is a selection of toys and furniture chosen randomly on a per-episode basis from the repository described in Table~\ref{tab:playroom_objects}. Figure~\ref{fig:playroom_objects} illustrates these objects. 

\subsection{Randomisation}

The following properties of the room are randomised per-episode. Where ranges are specified, the sampling interval is closed (inclusive) and the randomisation is uniform over integers (object quantities) or reals (dimensions):

\begin{itemize}
    \item The shape and size of the room: the room is an \emph{L}-shape, with the two longest walls varying in length between 6 and 10 metres, and no part of the room being narrower than 4 metres.
    \item The initial position and orientation of the agent anywhere inside the room. 
    \item The initial position and height of the shelves on the walls (between 0 and 8 shelves).
    \item The initial position of the doors and windows.
    \item The initial location of furniture, against the walls (between 2 and 4 items inclusive)
    \item The initial location and orientation of small objects on the floor (between 2 and 6 inclusive, chosen uniformly). 
    \item The initial location and orientation of small objects on top of furniture items (between 2 and 6).
\end{itemize}

\section{Data}

In this section we provide additional details regarding our data collection process.
The data we collected fall into two main categories: language game demonstrations and human annotations.

\subsection{Human Participants}
Participants were recruited through Google's internal labeling platform, a service that hires contractors to complete tasks. Subjects were given consent forms under DeepMind's HuBREC human subject research review protocol and were paid a fixed hourly rate. 

\subsection{Language Games}
\label{sec:language_games_data_collection}

\begin{figure}[hp]
    \centering
    \includegraphics{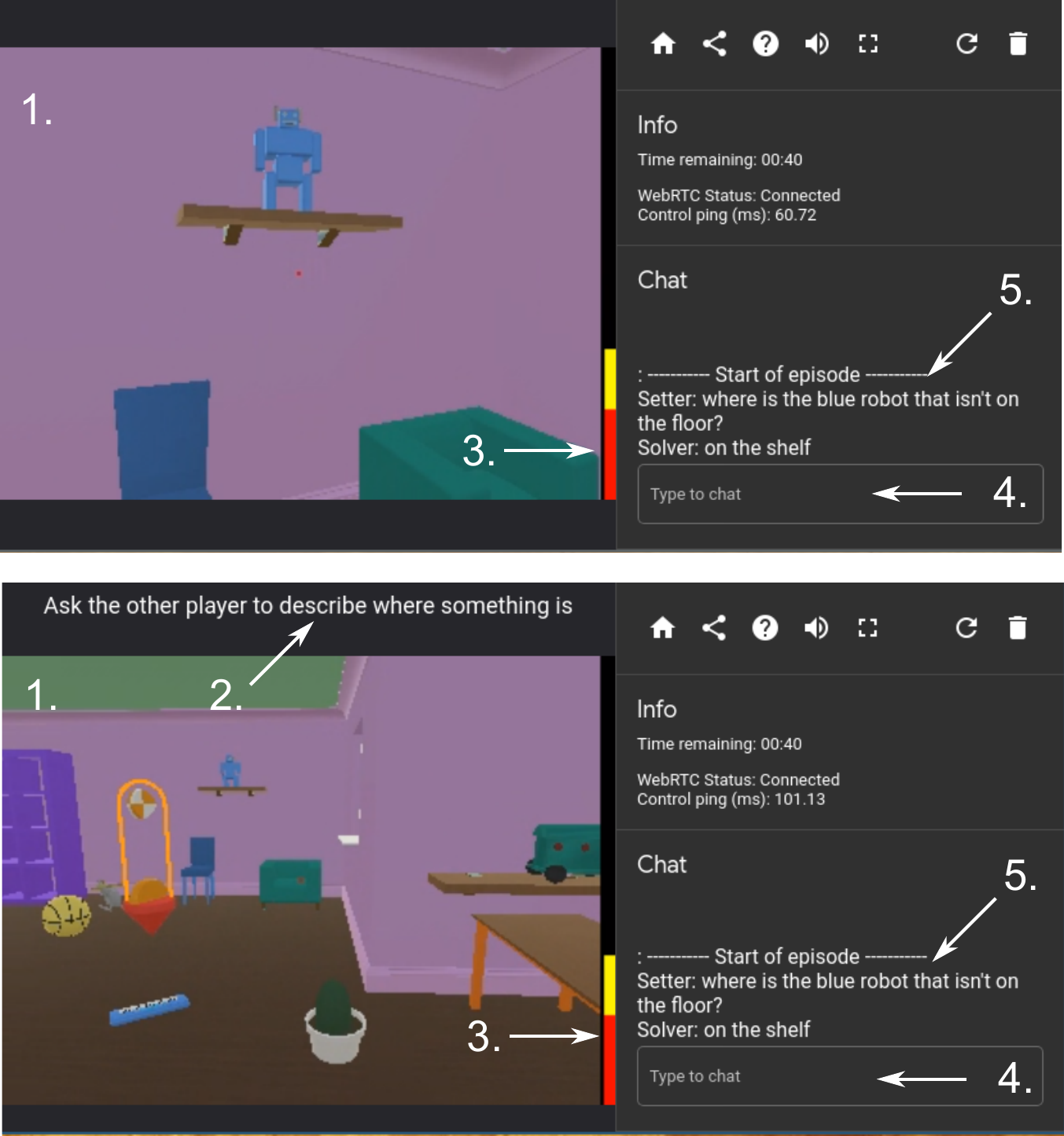}
    \caption[User interface for collecting language games demonstrations.]{
    User interface for collecting language games demonstrations.
    
    \vspace{1ex}
    \emph{Top:}~Solver's view, \emph{bottom:}~setter's view. \emph{Numbered elements:} 1.~First-person camera view; 2.~Game script (only shown to the setter); 3.~Meter showing the amount of time remaining until the episode ends automatically; 4.~Text entry box for typing messages to the other player; 5.~Chat history showing previous messages typed by both players.}
    \label{fig:language_games_ui}
\end{figure}

Each language game episode consists of a two-player interaction where one player (the setter) provides an instruction that the other player (the solver) must complete.
This interaction takes place within the Playroom described in Section~\ref{sec:playroomenv}.
The web interface used for collecting human demonstrations is shown in Figure~\ref{fig:language_games_ui}.
Players controlled their respective avatars with a keyboard and mouse, using the control scheme described in Section~\ref{sec:action_space}.
Players communicated via a chat dialogue in a sidebar.

\subsubsection{Data Collection Procedure}

At the beginning of each recording session the participants were randomly divided into two groups of equal size, A and B, with group~A initially assigned the role of setter and group~B the role of solver.
Pairs of participants were randomly selected, one from group~A and one from group~B, and assigned to play together in a particular game instance.
Participants were not told the identity of the partner they were paired with, and the two groups were seated apart from each other to ensure that the setter and solver could not see each others' screens or communicate with outside the game.
Within a pair, the players switched setter and solver roles every 30~minutes.
The pairs themselves were randomly shuffled every hour, such that each player from group~A was paired with a different partner from group~B.
Each participant therefore spent equal time playing as a setter and as a solver and had the opportunity to interact with multiple different partners over the course of data collection.

\subsubsection{Detailed Instructions}

\begin{figure}[htbp]
    \centering
    \begin{tikzpicture}[node distance=3cm,auto,>=stealth',align=center]
        \node[rectangle, draw] (env) {Environment};
        \node[rectangle, draw, below of=env, node distance=5.5cm] (env_ground) {Environment};
        \node[rectangle, draw, right = of env] (setter) {Setter};
        \node[rectangle, draw, below of=setter, node distance=5.5cm] (setter_ground) {Setter};
        \node[rectangle, draw, right = of setter] (solver) {Solver};
        \node[rectangle, draw, below of=solver, node distance=5.5cm] (solver_ground) {Solver};
        \draw (env) -- (env_ground);
        \draw (setter) -- (setter_ground);
        \draw (solver) -- (solver_ground);
        \draw[->] ([yshift=-1.5cm]env.south) -- node[above, text width=4cm] {\textit{``Ask the other player to count something''}} ([yshift=-1.5cm]setter.south);
        \draw[->] ([yshift=-2.5cm]setter.south) -- node[above, text width=4cm] {\textit{``How many red toys are there?''}} ([yshift=-2.5cm]solver.south);
        \draw[->] ([yshift=-3.5cm]solver.south) -- node[above, text width=4cm] {\textit{``There are 2.''}} ([yshift=-3.5cm]setter.south);
        \draw[->, dashed] ([yshift=-4.5cm]setter.south) -- node[above, text width=4cm] {End episode by key-press} ([yshift=-4.5cm]env.south);
    \end{tikzpicture}
    \caption{Sequence diagram representing the order of events within a single language games episode.}
    \label{fig:lang_games_sequence_diagram}
\end{figure}
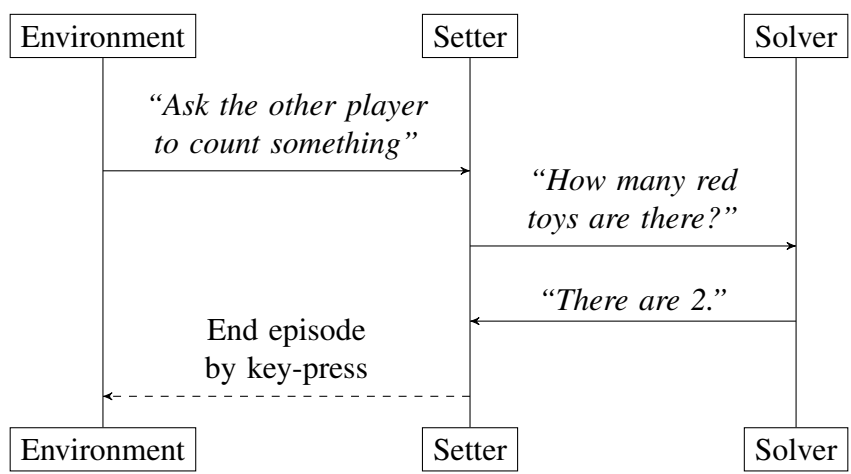

Figure~\ref{fig:lang_games_sequence_diagram} represents the order of events within a single language game episode.
At the beginning of each episode, the setter was given a textual cue indicating what type of instruction or question they should pose to the solver.
This cue consisted of two randomly sampled components: a ``prompt'' specifying the general type of instruction to give and a ``modifier'' that stipulated additional constraints the setter's instruction must satisfy.
For example, the combination of the {\tt Lift} prompt with the ``refer to objects by colour'' modifier resulted in the final cue ``Ask the other player to lift something. Try to refer to objects by colour.''
The modifier was omitted in a random a subset of episodes.
We found that including modifiers helped to increase the overall diversity of the language used by the human setters, and in particular encouraged setters to refer to attributes of objects other than their names (for example, colour or relative position).
Tables~\ref{tab:prompts} and \ref{tab:modifiers} contain the full set of prompts and modifiers respectively.
Table~\ref{tab:human_episodes_per_prompt} contains the total number of human demonstration episodes recorded for each combination of prompt and modifier.

Having given an instruction, the setter then observed the behaviour of the solver, and terminated the episode via key press if they were either satisfied that their instruction was completed successfully by the solver, or if they were certain that the solver would not be able to succeed (for example if the solver made an obvious mistake).
The episode ended automatically after two minutes if the setter did not terminate it manually within that time.

\begin{table}[htbp]
\small
\begin{tabular}{rp{0.5\textwidth}}
\toprule
Prompt                   &                                                                   Full text \\
\midrule
go                       &                                        Ask the other player to go somewhere \\
lift                     &                                      Ask the other player to lift something \\
position object          &       Ask the other player to position something relative to something else \\
position yourself        &              Ask the other player to stand in some position relative to you \\
bring me                 &                       Ask the other player to bring you one or more objects \\
touch                    &                Ask the other player to touch an object using another object \\
push object              &          Ask the other player to push an object around using another object \\
make a row               &         Ask the other player to put three or more specific objects in a row \\
arrange                  &   Ask the other player to move a group of objects into a simple arrangement \\
put on top               &              Ask the other player to put something on top of something else \\
put underneath           &             Ask the other player to put something underneath something else \\
freestyle activity       &                  Ask the other player to perform an activity of your choice \\
say what you see         &  Ask the other player to say what they are looking at or noticing right now \\
question about colour    &                                Ask a question about the colour of something \\
question about existence &          Ask the other player whether a particular thing exists in the room \\
describe location        &                         Ask the other player to describe where something is \\
count                    &                                     Ask the other player to count something \\
\bottomrule
\end{tabular}
\caption{Prompts used in language games.}
\label{tab:prompts}
\end{table}

\begin{table}[htbp]
\small
\begin{tabular}{rp{0.6\textwidth}}
\toprule
                      Modifier &                                                                Full text \\
\midrule
    refer to objects by colour &                                        Try to refer to objects by colour \\
   refer to location by colour &                                   Try to refer to the location by colour \\
               use shape words &  Try to use shape words like: circular, rectangular, round, pointy, long \\
  refer to objects by location &                                      Try to refer to objects by location \\
           use proximity words &                      Try to use words like: near, far, close to, next to \\
 use horizontal position words &      Try to use words like: in front, behind, left of, right of, between \\
   use vertical position words &                     Try to use words like: on top, beneath, above, below \\
            use negation words &                                        Try to use words like: not, isn't \\
          use quantifier words &                       Try to use words like: some, all, most, many, none \\
      not bed, door, or window &                                  Do not use the words: bed, door, window \\
\bottomrule
\end{tabular}
\caption{Modifiers used in language games}
\label{tab:modifiers}
\end{table}

\begin{longtable}{rlr}
\caption[Numbers of human demonstration episodes recorded for each combination of prompt and modifier]{
Numbers of human demonstration episodes recorded for each combination of prompt and modifier. `--' denotes cases where the prompt was given without a modifier.
} \\
\label{tab:human_episodes_per_prompt} \\
\toprule
Prompt & Modifier(s) & Episodes \\
\midrule
\endfirsthead
\caption[]{(continued)} \\
\toprule
Prompt & Modifier(s) & Episodes \\
\midrule
\endhead
arrange & -- &     14215 \\
\cline{1-3}
bring me & -- &     14314 \\
\cline{1-3}
\multirow{5}{*}{count} & -- &     35989 \\
      & refer to objects by colour &      6229 \\
      & refer to objects by location &      6212 \\
      & use negation words &      6106 \\
      & use shape words &      6111 \\
\cline{1-3}
\multirow{4}{*}{describe location} & -- &     35691 \\
      & refer to objects by colour &      6211 \\
      & use negation words &      6213 \\
      & use shape words &      6084 \\
\cline{1-3}
do two things in a row & -- &     13046 \\
\cline{1-3}
freestyle activity & -- &     14582 \\
\cline{1-3}
\multirow{5}{*}{go} & -- &     35777 \\
      & not bed, door, or window &      6274 \\
      & refer to location by colour &      6156 \\
      & use horizontal position words &      6137 \\
      & use proximity words &      6086 \\
\cline{1-3}
\multirow{8}{*}{lift} & -- &     49263 \\
      & refer to objects by colour &      6194 \\
      & refer to objects by location &      6108 \\
      & use horizontal position words &      6209 \\
      & use negation words &      6161 \\
      & use proximity words &      6094 \\
      & use shape words &      6118 \\
      & use vertical position words &      6170 \\
\cline{1-3}
make a row & -- &     14354 \\
\cline{1-3}
\multirow{7}{*}{position object} & -- &     35531 \\
      & refer to objects by colour &      6017 \\
      & refer to objects by location &      6147 \\
      & use horizontal position words &      6230 \\
      & use negation words &      6111 \\
      & use proximity words &      6137 \\
      & use shape words &      6075 \\
\cline{1-3}
position yourself & -- &     14470 \\
\cline{1-3}
push object & -- &     14297 \\
\cline{1-3}
put on top & -- &     14197 \\
\cline{1-3}
put underneath & -- &     14337 \\
\cline{1-3}
\multirow{8}{*}{question about colour} & -- &     35688 \\
      & refer to objects by location &      6074 \\
      & use horizontal position words &      6169 \\
      & use negation words &      6114 \\
      & use proximity words &      6122 \\
      & use quantifier words &      6092 \\
      & use shape words &      6124 \\
      & use vertical position words &      6156 \\
\cline{1-3}
question about existence & -- &     14329 \\
\cline{1-3}
say what you see & -- &     14564 \\
\cline{1-3}
touch & -- &     14544 \\
\bottomrule
\end{longtable}

\subsection{Human Annotations}

The second type of data we collected comprised human annotations of prerecorded episodes, generated either by human players or agents.

\subsubsection{Annotation Interface}

\begin{figure}[htbp]
    \centering
    \includegraphics{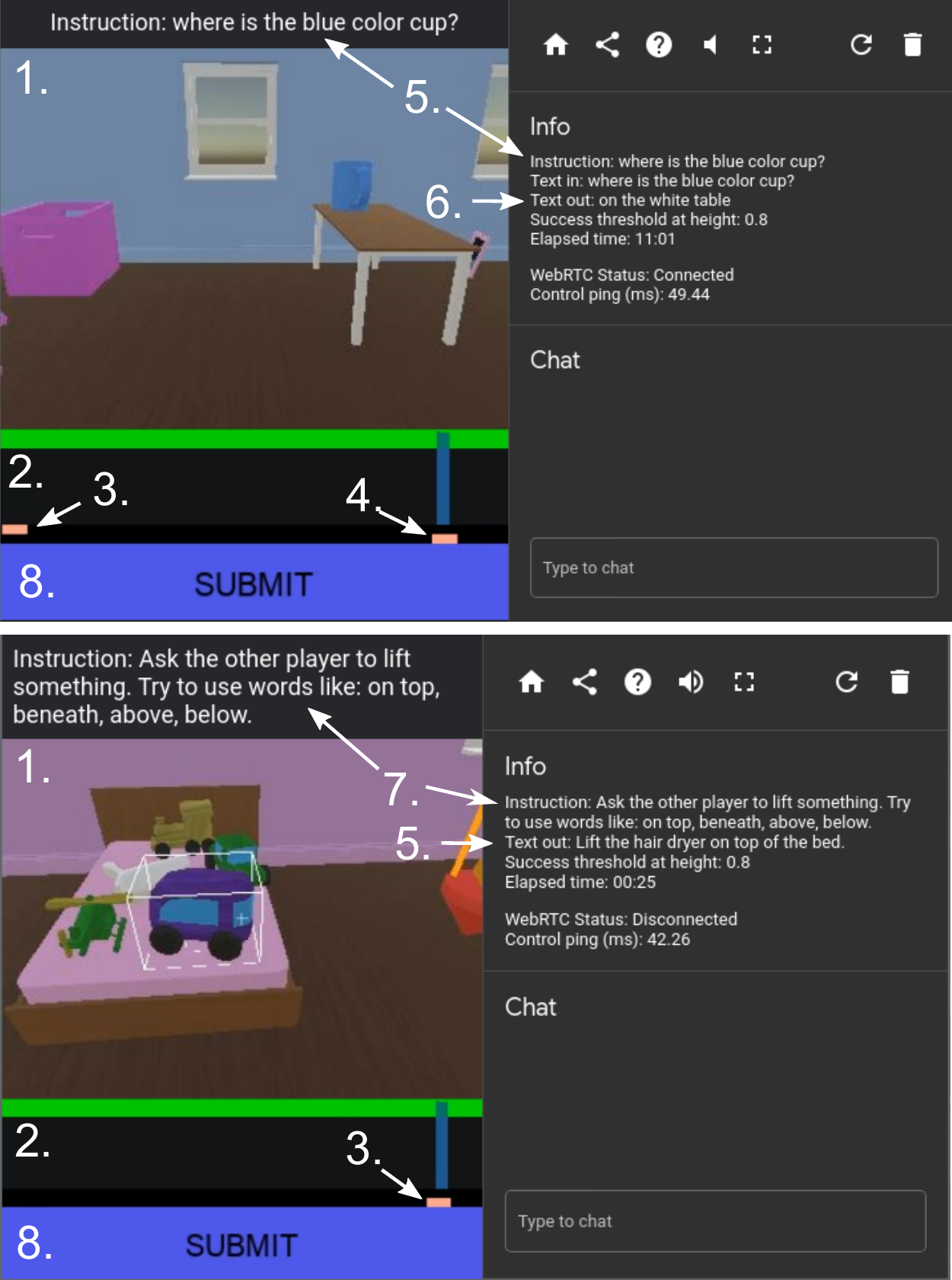}
    \caption[User interface for collecting annotations of language games episodes]{
    \textbf{User interface for collecting annotations of language games episodes}.
    \emph{Top:}~Solver's view; \emph{bottom:}~Setter's view. \emph{Numbered elements:} 1.~First-person camera view; 2.~Sketching interface; 3.~Marker indicating when a setter language emission occurred, 4.~Marker indicating when a solver language emission occurred; 5.~Setter language emission; 6.~Solver language emission; 7.~Prompt and modifier (only shown for setter sketching); 8.~``Submit'' button.
    }
    \label{fig:sketching_ui}
\end{figure}

These data were collected using a ``sketching'' interface similar to that used by \citealt{cabi2019scaling} (Figure~\ref{fig:sketching_ui}).
This interface allows human raters to scan through trajectories of first-person visual and text observations by moving the mouse cursor left and right, and to draw a ``reward sketch'' whose height represents the player's performance over time.

Although the sketching interface can record a graded level of reward across time, we found that this continuous mode of annotation was time-consuming for human raters to perform, and it was difficult to achieve consistency across different prompts and different human raters.
We instead chose to collect binary sketches by setting a height threshold representing the point at which the task is considered ``solved,'' represented by the green horizontal bars in Figure~\ref{fig:sketching_ui}. 
Raters were instructed to decide whether the player succeeded, and if so, to mark the moment of success by drawing a small ``spike'' that enters the green ``success'' region.
Each sketch therefore captures information about whether or not a particular episode was successful, and about when success occurred.
For evaluation purposes, each sketch was binarised and then reduced along the time dimension, yielding a single boolean label indicating whether or not the height of the sketch exceeded the success threshold at any point within the episode.

\subsubsection{Generating Episodes for Annotation}

In addition to collecting annotations for human-human demonstration episodes, we also collected annotations for four different types of episode that were generated by rolling out an agent policy (Table~\ref{tab:annotated_episode_types}). The cases included annotation of solver performance with a replayed setter instruction, annotation of setter success at producing a valid, feasible instruction, annotation of the success of a setter and solver agent interacting together, and annotation of solver success when interacting with a live human setter. In cases where the setter was a live human, episodes were usually terminated manually by the setter before the two minute time limit.
However, in cases where the setter was either a replayed human setter trajectory or an agent, no manual terminations were available, and therefore episodes always had a fixed duration of two minutes.
\begin{table}[htbp]
    \small
    \begin{tabular}{llll}
    \toprule
    {}                   &          Setter &      Solver &                    Termination \\
    \midrule
    Human demonstration  &  Live human     &  Live human &  Key-press or 2 min time limit \\
    Solver offline eval. &  Replayed human &       Agent &               2 min time limit \\
    Setter offline eval. &  Agent          &       No-op &               2 min time limit \\
    Joint offline eval.  &  Agent          &       Agent &               2 min time limit \\
    Solver online eval.  &  Live human     &       Agent &  Key-press or 2 min time limit \\
    \bottomrule
    \end{tabular}
    \caption{Episode types used for annotation.}
    \label{tab:annotated_episode_types}
\end{table}

\subsubsection{Truncation of Frame Sequences for Annotation}

We found that displaying full episodes made the annotation process slower and more difficult, since annotating longer frame sequences requires a greater degree of concentration and manual dexterity than shorter sequences.
We therefore truncated each sequence of frames that was displayed to the annotators in order to exclude frames that were unlikely to have a bearing on whether or not the episode should be judged as successful.

In the case of solver episodes we excluded all of the frames that came before the setter's first language emission, since during this time the solver had no instruction to carry out.
We also excluded all frames that came more than 5 seconds after the solver's first language emission (if there was one), since we required the solver's first emission to be correct in order for an episode to be considered successful.
5 seconds was chosen as the cut-off because over 95\% of human episodes where there was a solver language emission ended less than 5 seconds after the emission occurred.
For example, if the solver made multiple attempts to answer a question then we only counted the first answer they gave.
Finally, we truncated each frame sequence to a maximum duration of 60 seconds.
This time limit was chosen because over 95\% of human episodes terminated within 60 seconds after the setter gave the instruction.

In the case of setter episodes we excluded all frames that came after the setter's first language emission.
The motivation for doing this was that the setter should give an instruction that is valid \emph{given their current knowledge of the state of the room}, so only frames that occur before the instruction was given are relevant for judging its validity.
For example, a setter might say ``lift the blue teddy bear'' without first looking around the room to see if it contains a blue teddy bear.
We considered this to be a failure even if the setter happens to guess correctly, and there is indeed a blue teddy bear in the room.
We also truncated setter episodes to a maximum length of 75 seconds.
This time limit was chosen because it encompassed over 95\% of human setter emissions.

\begin{table}[htbp]
    \centering
    \begin{tabular}{rcccc}
    \toprule
    {} & \multicolumn{2}{c}{Accuracy} & \multicolumn{2}{c}{Balanced accuracy} \\
    \cmidrule(r){2-3} \cmidrule(r){4-5}
    {} &        Setter &        Solver &            Setter &        Solver \\
    \midrule
    Human           &  87.56 $\pm$ 0.22 &  91.88 $\pm$ 0.05 &      86.89 $\pm$ 0.24 &  88.24 $\pm$ 0.10 \\
    \mainagent{}    &  88.30 $\pm$ 0.38 &  88.05 $\pm$ 0.38 &      86.38 $\pm$ 0.47 &  86.32 $\pm$ 0.56 \\
    \BGA{}          &  88.61 $\pm$ 0.37 &  89.51 $\pm$ 0.48 &      86.87 $\pm$ 0.46 &  87.70 $\pm$ 0.82 \\
    \BA{}           &  87.29 $\pm$ 0.38 &  90.30 $\pm$ 0.46 &      85.26 $\pm$ 0.49 &  88.11 $\pm$ 1.41 \\
    \B{}            &  88.13 $\pm$ 0.40 &  94.08 $\pm$ 0.34 &      87.80 $\pm$ 0.46 &  89.90 $\pm$ 1.76 \\
    \BnoV{}         &  87.69 $\pm$ 0.32 &  98.22 $\pm$ 0.13 &      84.05 $\pm$ 0.91 &  84.33 $\pm$ 4.08 \\
    \BnoL{}         &  97.91 $\pm$ 0.14 &  98.01 $\pm$ 0.15 &      89.90 $\pm$ 2.60 &  86.07 $\pm$ 3.39 \\
    \bottomrule
    \end{tabular}
    \caption[Agreement between Human Annotations of Human and Agent Episodes.]{
    \textbf{Agreement between Human Annotations of Human and Agent Episodes.} \emph{Accuracy} corresponds to the proportion of individual annotations that are equal to the majority label for the corresponding episode.
    \emph{Balanced accuracy} was calculated by computing separate accuracies for episodes where the majority label was successful or unsuccessful respectively, and then taking the mean of these two values.
    $\pm$ denotes a 95\% CI of the mean.
    }
    \label{tab:rater_agreement}
\end{table}

\begin{figure}[htbp]
    \centering
    \includegraphics{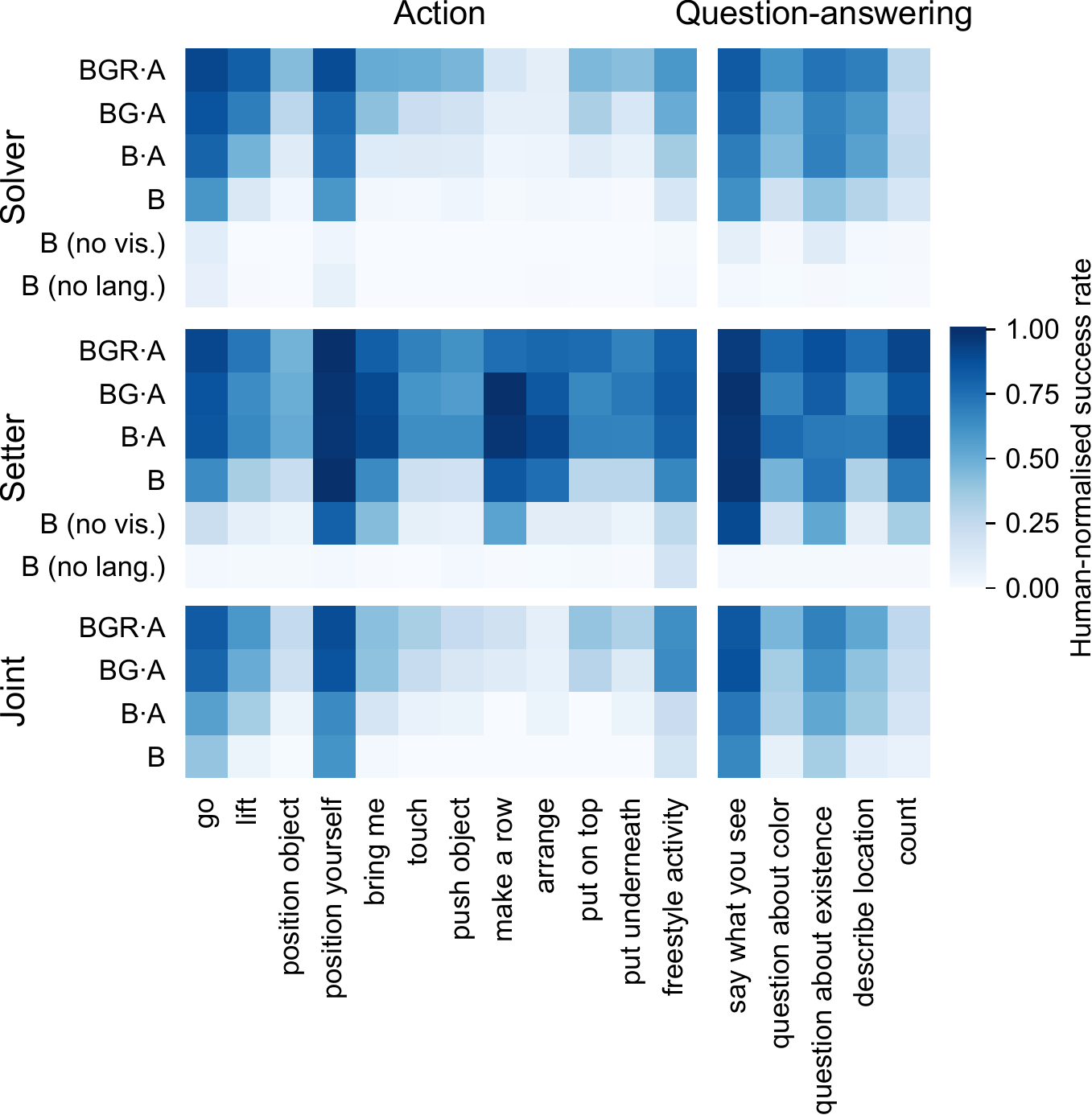}
    \caption[Observational Human Evaluation Results per Prompt.]{
    \textbf{Observational Human Evaluation Results per Prompt}. Each heat map pixel represents the mean success rate of a given agent as judged by human raters, expressed as a fraction of human baseline performance for the corresponding script. 
    }
    \label{fig:obs_eval_results_by_prompt}
\end{figure}  



\section{Agent Architecture}
\subsection{Inputs}
\label{appendix_agent_inputs}
Setter and solver agents inputs comprised multi-modal sensory perceptions and miscellaneous extra information used for auxiliary supervised learning or unsupervised learning, or used as hard-coded features (such as whether an object is currently being grasped, or previously chosen actions).

\subsubsection{Perception}
\label{sec:appendix_agent_perception}

Each agent's multi-modal input comprised $96\times72\times3$ resolution RGB images depicting the agent's first person perspective of the 3-D room, and two types of language, formatted as simple multi-word text strings. The first language text came from the environment and provided information to the setter about the episode's particular interaction type (e.g. ``Tell the other player to lift something''), or an empty string for the solver. The second came from the other agent in the room, providing a dialogue channel used, for example, by setters to communicate an instruction to a solver. 

RGB images were processed by a ResNet architecture \citep{he2016deep}, composed of $5$ residual blocks. Each residual block had two stages of processing. The first consisted of a $3\times3$ convolution followed by an optional max pooling operation with a $3\times3$ window size, downsampling the incoming image by half along each dimension. The second stage consisted of two loops over a sequence of $4$ computations: a ReLU non-linearity, a $3\times3$ convolution, a ReLU non-linearity, and a final $3\times3$ convolution. The input to each pass of the loop is summed with the output, implementing a residual connection. Finally, the output of the entire residual block is passed through a ReLU non-linearity. Therefore, altogether each residual block consisted of $5$ total convolutions, one optional max-pool, and two residual connections. The ResNet architecture as a whole thus had $25$ total convolutional layers. In pseudocode, the ResNet block was:

\begin{footnotesize}
\begin{lstlisting}
def residual_block(input):
    conv_out = conv(input)
    block_input = max_pool(conv_out)
    for _ in range(2):
        conv_out = block_input
        conv_out = relu(conv(conv_out))
        conv_out = relu(conv(conv_out))
        block_input = conv_out + block_input
    return conv_out
\end{lstlisting}
\end{footnotesize}

Each of the $5$ convolutions within a given residual block used the same number of kernels. The number of kernels for each block were $16, 32, 64, 128$, and $256$. We opted to implement max-pooling for every residual block except the first, resulting in $4$ downsampling operations across the ResNet. Therefore, the ResNet computed a $6\times5\times256$ output for a given $96\times72\times3$ input image.
Finally, each of the $6\times5$ ResNet output vectors of length $256$ was linearly projected to $512$ dimensions (i.e., $6\times5\times512$), and then the set were reshaped to be a $30\times512$ matrix by merging the height and width dimensions. Each row, therefore, corresponded to a $512$ dimensional feature vector for a particular ``pixel'' in the ResNet output.

\subsubsection{Text Preprocessing}
\label{app:text_process}
Text inputs underwent minor preprocessing before being provided as inputs to the agent. First, we tokenised the string using a space delimiter, forced lower casing, and stripped punctuation. Next, we applied basic typo correction using the following four-step process to each word token: (1) if the word was already present in the output vocabulary then it was returned unchanged; (2) if the word was a concatenation of two words in the output vocabulary then the missing space was inserted; (3) if there was a predefined correction specified in a custom typo-fix dictionary, which manually mapped common typos to their corrections, then this correction was be applied; (4) if the word was within a closeness threshold, implemented using the standard Python difflib package with a threshold setting of $0.5$, or a word in the output vocabulary then it was replaced by the word from the output vocabulary.

We constructed our agents' vocabulary by processing a sample of human language from our dataset, correcting for typos as just described, and selecting the top $500$ most frequently used words. Next, we appended to this vocabulary words known to be used in the procedural evaluation instructions, resulting in the final vocabulary for our agents. We constructed a spelling correction table to detect common typos. Both the vocabulary and the typo correction table are attached in Section~\ref{sec:vocab}.

Input strings, which at this point are tokenised into words and typo corrected, were then converted to integers using a static word-to-integer mapping and either truncated or padded to a set length of $16$ total integers. Finally, these sequences of $16$ integers were used to look up an learned embedding table, resulting in size $512$ vectors representing each token. Each set of $16$ vectors therefore represented each source of input text to the agent; i.e., text from the environment or inter-agent communication.

\subsubsection{Miscellaneous Features}
\label{appendix_agent_misc_features}
The final source of inputs to the agent were miscellaneous features, comprising an extra text source for auxiliary supervised or unsupervised learning, an extra text source indicating the previous language action, hard-coded features indicating the number of steps since the last non-noop target, and hard-coded features indicating the number of steps since the last time an agent made a decision about whether to emit an action (as opposed to choosing not to act, or no-oping). The latter were represented on a log scale, $\log (\text{steps})$, and were provided as input to the no-op policy, as described below in section \ref{appendix_agent_outputs}.

\subsection{Sensory Integration by the Multi-Modal Transformer (MMT)}
\label{appendix_mmt}
After perceptual processing, the agent had available a set of $30$ $512$-dimensional visual representations, one for each ``pixel'' in the ResNet output, two sets of $16$ $512$-dimensional vector embeddings, one for each word in each of the text inputs, and one $512$-dimensional vector representing the token from the previous step's language emission. These vectors comprised a size $30+16+16+1=63$ set of $512$-dimensional vectors. 

To this set of $63$ vectors we appended two more $512$-dimensional vectors whose initial activations were learned. These additional two vectors were used in a way analogous to the CLS token used in BERT architectures \cite{devlin2018bert}, as will be described. Together, the $65$ vectors comprised the input to an $8$-layer, $8$-head transformer \cite{vaswani2017attention} with size $512$ embeddings and MLP layers, using relative position encoding \cite{shaw2018self}.

The CLS-like channels were free to attend to all of the other input embeddings, acting as a dedicated attention-based ``output aggregator'' for the transformer (since transformer outputs are a set of embeddings, some sort of aggregation or reshaping is needed to pass their output to any downstream module, which in our case was an LSTM). We also performed a feature-size mean-pooling operation across all the others embeddings. These three vectors (the $2$ CLS-like embeddings and the one aggregate embedding) were concatenated together to form a $1536$-dimensional vector that was passed along to an LSTM.

\subsection{Memory}
\label{appendix_agent_memory}
We used a two-layer, $512$-dimensional LSTM as memory in our agent. The output of the LSTM, a $1024$-dimensional vector, was concatenated with the LSTM's input to implement a rudimentary skip connection past the LSTM memory. This vector served as the inputs to the various policy heads in our agent, described next.

\subsection{Outputs}
\label{appendix_agent_outputs}

The output of the agent's memory served as the input to various policy heads: an aggregate motor policy, which produced actions for movement, looking, and grabbing, and a language policy, which produced single word emissions from the agent's vocabulary per timestep. Overriding each of the motor and language policies was a no-op policy, which dictated whether an action should be chosen for the current step or not. When trained with GAIL, motor actions were produced at 15 frames per second and repeated for two steps in a row to reach 30 frames per second. The behavioural cloning loss skipped every other action in the dataset. This was probably not an optimal modelling choice, but it initially helped GAIL training by simplifying the reinforcement learning exploration and credit assignment. For BC agents that did not also train with GAIL, we tried modelling actions at 30 frames per second and at 15, with 30 working better.

\subsubsection{Language Policy}
The input to the agent's language policy was the output from the memory, described in section \ref{appendix_agent_memory}, concatenated with two features: a bit representing the decision about whether or not to act, as determined by the no-op policy (see section \ref{appendix_agent_no_op}), and a bit representing whether the agent had already acted in the episode. 

For the agent's language policy we used a simple one-layer, $512$-dimensional MLP with ReLU non-linearity followed by a $512$-unit linear layer. We then computed weights corresponding to the agent's preferred word emission, $w=\text{softmax}(Ex)$, where E is the row-wise learnable embedding matrix for the vocabulary mentioned previously for tokenizing and embedding input text, and $x$ is the linear layer's output. These weights were used a logits for a categorical distribution across the vocabulary, which allowed us to compute log probabilities of the target word when doing behavioural cloning, or for sampling when running the agent online. 

A notable feature of this language policy was the shared encoding and decoding of language embeddings: the embeddings used to encode text in the agent input were the same as those used to decode the agent's output representation into a word, $E$. Thus, the agent used the same representation for a given word whether it was processing it as input (e.g., when a solver is told to ``lift a duck''), or whether it was choosing a word to utter (e.g., when a setter is asking a solver to ``lift a duck'').

\subsubsection{Motor Policy}
The motor policy had three subcomponents: the movement policy, the grab policy, and the look policy. 
The movement policy consisted of a one-layer, $512$-dimensional MLP with ReLU non-linearity followed by a linear projection to a $9$-dimensional vector representing the logits for a categorical distribution across movement actions: right, left, forward, back, forward right, forward left, backward left, backward right, and no movement (no-op). The grab policy was similar to the movement policy except the categorical distribution was across two actions: grab and no-op. The look policy also started with a one-layer, $512$-dimensional MLP with ReLU non-linearity. This provided the input to a small $100$-unit LSTM that implemented a recursive discrete decision procedure where coarse decisions about where to look were gradually refined over $5$ steps. At each step, each dimension of the continuous ``looking space'' (i.e., the space represented by the current visual RGB input) was divided into $9$ segments, partitioning both the height and width dimension of the space into $3$ discrete partitions. One partition was sampled for each dimension and recursively divided in the same manner. In this way one action in the continuous space was represented as a sequence of discrete actions. This procedure provided a limit to the resolution for ``looking,'' which could increase if the number of steps was increased, but we capped the resolution at $0.01$, assuming an original size of $2$ units for each $x$- and $y$-dimension.

\subsubsection{No-Op Policies}
\label{appendix_agent_no_op}
Both the motor and language policies could be vetoed by a no-op policy, which decided whether an action should be exposed by the agent to the environment at any given timestep (practically, the motor and language policies always sampled actions, but it was the no-op policies' job to determine whether these actions would be passed along to the environment, and hence, whether they would actually be enacted by the agent). The no-op policies were one-layer, $512$-dimensional MLPs with ReLU non-linearities, followed by a linear projection to a $2$-dimensional vector, which acted as the logits to a categorical distribution over two actions: op, and no-op. The input to the MLP was the output described in section \ref{appendix_agent_memory} concatenated with the hard-coded features described in section \ref{appendix_agent_misc_features}: hard-coded features indicating the number of steps since the last non-no-op target, and hard-coded features indicated the number of steps since the last time an agent made a decision about whether to emit an action. 

\section{Agent Training}
\label{app:agent_training}
We used two principal methods to train agents: supervised learning-based behavioural cloning to expert human interactions, and a form of inverse reinforcement learning, specifically Generative Adversarial Imitation Learning \citep{ho2016generative}. 

\subsection{Data Processing}

We preprocessed the language games data, described in Section~\ref{sec:language_games_data_collection}, before it was used in training. When the human player does not move, actions are registered as ``no-ops.'' We removed these actions and their corresponding observations from trajectories. If a trajectory contains a sequence of no-ops, we condensed them to a sequence of just two no-op actions.

The recorded text fields in the data were also preprocessed to correct for typos and match the agent's vocabulary as described in \ref{app:text_process}.

\subsection{Supervised Learning (Behavioural Cloning)}
\label{app:agent_training_bc}
An expert trajectory comprised the observations, or inputs (RGB images, and any text input, see section \ref{appendix_agent_inputs}) and the actions taken (see section \ref{appendix_agent_outputs} for information about the variety of actions). Therefore, for a single trajectory in a batch, expert observations are given sequentially to the agent, which then produces its predicted action distribution for the move, look, grab, no-op, and language policies. Each of these policies was trained to maximise the likelihood of the expert action. The loss terms had unequal coefficients: $\omega_\text{LANG}=50, \omega_\text{MOVE}=1$. We used the ADAM optimizer \citep{kingma2014adam} with a batch size of 192 and sequence (unroll) length of 50. Hyperparameters for all training, including RL, are presented in Table~\ref{tab:hyperparameters}.
    
While expert language productions were multi-word (e.g., ``lift the yellow duck on the table'') and recorded at the time point when the subjects pressed enter, to simplify the model we preprocessed these target language actions in the dataset by smearing the tokens across time, after the emission, ensuring that each step only required the agent to predict a single word token, rather than the full multi-word text. For example, if at time $t$ the language target was ``lift the yellow duck on the table'' according to the expert human data, then after preprocessing the target at time $t$ became ``lift'', the target at $t+1$ became ``the'', and so on. While this method produced a slight distortion between the time the experts actually emitted language and when the agents were asked to emit language, in practice we did not see any detrimental effects. Instead, agents performed better when only tasked with emitting a single token per timestep. While we did not fully explore the exact reasons behind this, we hypothesize a number of effects might be at play: (1) smearing language across time increases the proportion of timesteps that include a language target, decreasing the sparsity of the language gradients, which can have subtle implications for computing, for example, the momentum parameters in the optimizer; (2) smearing language across time allows the agent core to receive an unadulterated gradient signal for any given word prediction, as opposed to the non-smeared case where the gradients across all word predictions are intermingled; (3) the model architecture was simplified. However, we believe these results were context-dependent, and there may be cause to revisit them.

Although agents were trained as both setters and solvers, we did not explicitly indicate the particular role of the agent (i.e., whether it was a setter or a solver for a given episode) because this information was indirectly revealed by the presence for the setter or absence for the solver of the prompt language input. 

\subsection{Unsupervised and Auxiliary Supervised Learning}
A particularly difficult aspect of modeling the expert data using behavioural cloning was the relative density of each policy target. Move and look actions more densely populated the trajectories (though were still relatively sparse compared to no-ops), while the grab and language policies were very sparse. Given that most trajectories involved only a single language emission for the setter (and sometimes zero language emissions for the solver, if it was just performing a motor task), only a single time step out of approximately $2000$ contained a language target (though, after smearing, this resulted in about $6$ timesteps out of every $2000$, with an average emission length of approximately $6$). 

This was a non-ideal circumstance for supervised learning, since batches of data could only be expected to have a handful of language and grabbing targets, significantly reducing the effective batch size for these targets. Unfortunately, the effects of sparsity are even more pernicious and difficult to resolve. With a relatively strong learning signal to train the move and look policies, and a weak signal to learn the language policy, we found that naive supervised training on expert data resulted in very poor language policies regardless of the length of training. We did not complete a full battery of experiments to conclude exactly what the underlying effect was; however, we hypothesise a few: (1) if there is a strong, low-variance gradient for one type of target policy compared to another, then the model parameters may specialise to predict the dense targets and at the expense of the sparse targets; (2) the effective batch size for the sparse targets might simply be too low for effective training, precluding proper learning in any practical amount of time; (3) the sparse, high-variance language action gradients and dense, low-variance gradients may compete to influence the updates for the optimiser parameters (e.g. the normaliser in Adam), and the optimiser may then become even less sensitive to the language gradients.

This sparsity problem was important to overcome since the language target data was a rich source of information to learn about object identities, grounding the words for particular words (``duck'') to the pixel inputs (i.e., the actual shape of a duck in the visual field). This is not only useful for setter language policies, but also motor policies, since being able to recognise objects is a necessary condition for being able to manipulate them. 

We fortunately developed a robust solution with two prongs using both unsupervised learning and auxiliary supervised learning. These methods enabled the agents' perceptual systems to develop the capacity to recognise objects and actions and provided dense and discriminative gradients at each time step.

\subsubsection{Language Matching (LM)}\label{section:lm}
The Language Matching (LM) auxiliary task was partially inspired by developments in contrastive self-supervised learning \citep{chopra2005learning, gutmann2010noise, oord2018representation, henaff2019data}. The idea was that the visual observations in an expert trajectory are correlated with the instruction provided by the expert setter. This was especially true for instructions like manipulating named objects, going to locations, etc. We made use of this observation by effectively doubling the batch size: in the first part of the batch we had the visual observations and associated language input to solver from real trajectories; in the other part of the batch, we had the same visual observations and the language input from other trajectories (shuffled from the same batch by taking the language from the next batch element modulo the batch size $B$).

We added a simple MLP classifier head to the multi-modal transformer taking in the original batch elements and the shuffled ones, training it to classify them correctly using a conventional bernoulli cross entropy loss. This loss was only active during behavioural cloning training of the solver and non-active during interactive training or when training as the setter.

\subsubsection{Object-in-View Auxiliary Supervised Learning (OV)}
Many of the emissions in the expert setter language involved objects in the room. For setter agents, language often referred to objects at a distance as well, where they were harder to recognise. Solvers would often approach and manipulate objects, giving them clearer views, which made the language matching loss work. However, for setter training, language matching was insufficient for training agents to recognise objects at a distance in crowded scenes to enable successful language generation.
We introduced the Object-in-View (OV) auxiliary task, which worked by proposing particular colour-object combinations (e.g., ``yellow duck'') and forcing the agent to decide whether this combination was in view or not. Intuitively, an agent that can successfully learn of this task should have a strong command over basic object and colour identification, invariant to the object's position, angle, partial occlusion, and so on.

To implement this loss we began by choosing a colour-object combination for each timestep, choosing with a $50\%$ probability whether a given step would include a colour-object pair that was within view or not. The colour-object pair was represented by a simple two-word string, which we embedded into two $512$-dimensional vectors using the language embedding method described previously for processing text inputs. We then took the feature-wise mean of these two vectors as the final representation of the colour-object pair.

Next, we took the output of the agent's LSTM memory (concatenated with the LSTM input, as described previously), and passed it through a $2$-layer MLP with $512$ units per layer. We then performed the dot product between the MLP output and the colour-object representation, the result of which was used to compute a bernoulli cross entropy loss with the binary target. Similar to the behavioural cloning losses, we used a scalar coefficient of $20$ for the OV loss.  

\subsection{GAIL and Interactive Training}
\label{app:gail_training}

In addition to training the agent via a supervised method such as behavioural cloning we also used a form of inverse reinforcement learning, specifically Generative Adversarial Imitation Learning (GAIL) \citep{ho2016generative}. GAIL is an algorithm closely related to IRL \citep{ziebart2010modeling, finn2016connection}, which trains a \emph{discriminator} model to distinguish demonstrator trajectories from imitator / agent trajectories. A function of the discriminator's output is converted into a reward for the agent, which trains by RL to make trajectories that appear to the discriminator like the demonstrator trajectories.

\subsubsection{GAIL Data Processing}
When training with GAIL, we additionally preprocessed the data. 
First, the visual observations provided to the discriminator were modified using RandAugment \citep{cubuk2020randaugment}. In particular, two random image geometric image augmentations were performed from the set of rotation, shearing, and translation. In addition, the images were randomly cropped by 10 pixels.

The original data was recorded at 30 frames per second. However, to improve the RL movement policy exploration, we strided the data and used every other observation and original action. When executing the agent, the actions were sampled by the agent at 15 frames per second, with each action repeated for two time steps in a row. Empirically, this substantially improved RL training with GAIL. Future work using stronger RL optimisers may enable this action repeat to be dropped.

\subsubsection{Interactive Training}
Experience for the reinforcement learning updates was generated through two different simulation environments: a multi-player interactive training environment and a setter replay environment. In each of these environments, the agent generated a trajectory and received reward from the reward model.

In the multi-player interactive training mode, one single model was instantiated twice, one acting as a setter and one as a solver. The agent in the setter role received a prompt from the environment and had to produce an instruction or question which is achievable given the current room configuration. The agent in the solver role received this instruction and had to carry out the task or answer the question. The trajectories generated during the interaction were processed by the GAIL discriminator and used to train via reinforcement learning. In this work, we only updated the policy via RL on solver trajectories.

\subsubsection{Setter Replay (SR)}

During early stages of training, when the language policy was still largely untrained, the instructions produced by the setter were often erroneous or not achievable. This produced a significant number of interactions that were not useful for training the solver, and therefore wasted compute time. To mitigate this, in half the episodes we replayed human setter trajectories from the dataset verbatim instead of running the setter agent policy. For this, we also retrieved the Playroom's initial configuration from an episode in our database and followed the human setter activity from that episode step-by-step. 

\subsubsection{GAIL Discriminator Architecture}
\subsubsection{Inputs}
\label{appendix_gail_inputs}

The  discriminator scored short sequences of observations, which were then converted into a reward to train the agent. Both trajectories generated from the multi-player interactive environment and from the setter replay served as negative examples for the discriminator training. Observation sequences from the expert dataset of human interactions served as positive examples.

\subsubsection{Perception}
As in the agent, the discriminator processed multi-modal perceptual inputs with images, depicting the agent's first person perspective of the 3-D room, and language input, formatted as simple multi-word text strings. The text input came from either the agent, from the other agent via setter replay of prerecorded trajectories, or from human interaction when executing the trained agent.

The discriminator used the same ResNet architecture as the agent to process RGB images. As in the agent, each of the $5$ convolutions within a given residual block used the same number of kernels. The number of kernels for each block were $16, 32, 64, 128$ and $256$. The ResNet output was reshaped to be a $30\times256$ matrix by merging height and width dimensions. Each row, therefore, corresponded to a $256$ dimensional feature vector for a particular vector in the ResNet output's spatial array.

The text input was similarly preprocessed by tokenising and typo correcting. The discriminator was also provided with an extra text source indicating the language action from the agent from the last time step.

\subsubsection{Multi-Modal Integration}
After encoding the image and text, the discriminator also used a multi-modal transformer (MMT) to merge visual and text representations (see Section~\ref{appendix_mmt}). 
The output of this module at each timestep was mean-pooled and concatenated to the output from $2$ CLS-like channels, making a $768$d vector $\mathbf{e}_t$, which was passed to a two-layer MLP (hidden size 256) to train a language matching classifier (see Section~\ref{app:clm_disc}). 

\subsubsection{Buffered Memory}
\label{appendix_gail_memory}
We used buffered sequences of the outputs of the MMT within the discriminator. These sequences consisted of the $8$ previous MMT outputs strided by $2$ steps: $\mathbf{e}_{t-16}, \mathbf{e}_{t-14} \dots, \mathbf{e}_{t-2}, \mathbf{e}_t$. With the agent already operating on strided observations of $2$ steps, this extended the observation history for the discriminator to 32 real time frames or about $1$ second of history.

The buffered (over time) input, was passed through a second temporal transformer using relative position encoding \cite{shaw2018self} with $2$-layers and $4$-heads  \cite{vaswani2017attention} with size $256$ embeddings. 
The transformer output was then passed to a final MLP, with hidden size of 256 to produce the discriminator output $D_t$. Reward for the policy was computed as $r_t = - \ln (1-D_t)$.

\subsubsection{Language Matching (LM)}
\label{app:clm_disc}
We applied the same language matching loss $L_{LM}$ that we used in the agent (see Section~\ref{section:lm}) within the discriminator. We primarily relied on language matching to optimise representations in the discriminator, by reducing the relative scale of the discriminator cross entropy loss: $L_\text{LM} + \alpha L_\text{GAIL}$, with $\alpha$ set to $0.01$.
        
$L_\text{LM}$ was applied to the output of the MMT and only trained using data from expert trajectories (shuffled and unshuffled), whereas $L_\text{GAIL}$ was applied to the whole output of the discriminator after processing with the temporal transformer.

\subsubsection{Reinforcement Learning}
We adopted the distributed RL training framework Importance Weighted Actor-Learner Architecture \citep{espeholt2018impala}. 
Agent trajectories were generated on ``actor'' computers on CPUs and then sent to a ``learner'' in a $[T, B]$ format, where $T$ is the unroll length and $B$ the batch size. The trajectories for supervised learning were combined with the trajectories from RL, making a full batch of size $2 \times 192$, with different losses applied to supervised learning and RL batch elements. 
The value function baseline for RL was implemented in the agent by an additional MLP head with a hidden layer size of 512 taking in the same inputs as policy heads do. 
We used a small entropy loss in the policy gradient update \citep{mnih2016asynchronous, espeholt2018impala}.
Both the movement and language policy (Section~\ref{appendix_agent_outputs}) shared the same rewards and value function $V_\theta$. The returns $R_t$ for each policy head were computed independently using the respective off-policy corrections \citep{espeholt2018impala}.
Table~\ref{tab:hyperparameters} contains a list of all the training hyperparameters.
    
\begin{table}[htbp]
    \small
    \begin{tabular}{lll}
    \toprule
    Hyperparameter               &          Value &          Description \\
    \midrule
    $\eta_a$  &  1e-4  & Agent learning rate (BC \& RL) \\
    $\eta_d$  &  1e-4 & Discriminator learning rate\\
    $\beta_1^\pi$  &  0.0 & Agent Adam $\beta_1$\\
    $\beta_2^\pi$  &  0.999 & Agent Adam $\beta_2$\\
    $\beta_1^D$  &  0.9 & Discriminator Adam $\beta_1$\\
    $\beta_2^D$  &  0.999 & Discriminator Adam $\beta_2$\\    
    $\gamma$  &  0.9 & Agent discount factor\\
    $\epsilon$ &  1e-5 & Scale factor for entropy term\\
    $T$ &  50 & Unroll length\\
    $B$ &  192 & Batch size\\
    $\alpha$ &  1e-2 & Balance between GAIL and LM loss in discriminator\\
    $T$ &  50 & Unroll length\\
    $B$ &  192 & Batch size\\
    $\omega_{\text{LANG}}$ &  50 & coefficient for language policy loss  \\
    $\omega_{\text{MOVE}}$ &  1 & coefficient all movement policy losses (move, grab, and look)  \\
    $\omega_{\text{LM}}$ &  1 & coefficient for language matching loss  \\    
    $\omega_{\text{OV}}$ & 20 & coefficient for Object-in-View Loss \\
    
    \bottomrule
    \end{tabular}
    \caption{Hyperparameters for supervised learning and RL.}
    \label{tab:hyperparameters}
    \end{table} 

\section{Distributed Training Infrastructure}
\label{app:distribtrain}

The agent and reward model were trained in a distributed fashion. Overall the setup was similar to IMPALA (Importance Weighted Actor-Learner Architectures) \cite{espeholt2018impala}. \textit{Actors} ran on multiple CPUs. Actors simulated environments and performed inference on agent models to generate actions. \textit{Learners} ran on accelerators, in this case tensor processing units (TPUs) \citep{jouppi2017datacenter}, and performed parameter updates using the data generated on actors. Model parameters were synchronised from learners to actors on a regular basis.

The difference from IMPALA for the experiments presented here was that there were several types of actor. Some ran through setter and solver dataset trajectories for supervised training; some generated both setter and solver trajectories for interactive training; and some generated setter replay episodes where the room layout and the setter actions came from dataset trajectories. We used two separate learners: one for the agent and one for the reward model. In addition, to monitor training, we used two types of evaluation actors: one for the scripted probe tasks and one to calculate metrics like log-probabilities and language output metrics by running through dataset trajectories.
More details follow in the remainder of this section.

\subsection{Actors}

Actors were split into three types, which sync parameters at the start of each unroll:
\begin{enumerate}
    \item \textit{Dataset Actors:} Episodes for the environment on these actors are replays of the episodes in the stored human data, from the view of the setter or the solver (in equal proportion). Teacher forcing is used for agent actions, i.e.\ actions (for both movement and language) are forced to be the same as the actions in the data. For each timestep, inference is run on the agent and reward model and as usual state is maintained between steps (and reset to initial state at the start of each episode). Once enough steps have been taken to complete one unroll (episode boundaries may come in the middle of this) the data is stacked and sent to both the agent and reward model learners, to be used for behavioural cloning and GAIL discriminator learning respectively.
    \item \textit{Interactive Training Actors:} Episodes for the environment on these actors are random instantiations of the Playroom environment described in section~\ref{sec:playroomenv}. The current agent parameters are used to do inference (separately) on observations from the point of view of setter and solver. This inference produces actions for both players that are used to step the environment. Inference is also run on the current reward model, based on visual observations from the solver perspective only, and rewards are thus generated for the solver. Once enough steps have been taken to complete one unroll (episode boundaries may come in the middle of this) the solver data is stacked and sent to both the agent and reward model learners, to be used for reinforcement learning and GAIL discriminator learning, respectively.
    \item \textit{Setter Replay Actors:} Episodes for the environment on these actors are partial replays of the episodes in the stored human data. The initial layout of the room, including the type, colour and position of all objects, is taken from an episode of stored data. The actions of the setter are taken from the human setter trajectory. In all other respects, these actors are then the same as the interactive training actors.
\end{enumerate}

Note that on all these actors, the language output of the setter becomes the language input observation for the solver, and vice versa. The language game prompt is provided as an observation to the setter only.
Note also that each CPU can run multiple environments simultaneously. For the experiments presented here, we used 2,000 dataset actors with 8 environments per actor and 2,000 online environment actors with 4 environments per actor. Online actors were either all interactive training or all setter replay, or 1,000 of each.

\subsection{Learners}

There are two different learners:
\begin{enumerate}
    \item \textit{Agent Learner:} The agent learner updates parameters for the agent. Per step it receives one batch of mixed setter and solver unrolls from the dataset actors, which it uses for behavioural cloning, language matching, and object-in-view losses. It also receives per step a batch of solver unrolls (same batch size) from online environment actors (the two types of online actors, if they are both running, feed to the same queue), which it uses for reinforcement learning losses with the rewards coming from the GAIL reward model (already computed on the actors).
    \item \textit{Reward Model Learner:} The reward model learner updates parameters for the reward model. Per step it receives a batch of solver data from dataset actors and a batch (with the same size) of solver data from online actors (the two types of online actors, if they are both running, feed to the same queue). It uses the dataset batch for the language matching loss and then both batches together for the GAIL discriminator loss.
\end{enumerate}

Note that parameters are synced to separate \textit{cacher} CPU workers regularly and actors sync their parameters from these cachers rather than directly from the learners. The sync frequency from learners to cachers is shorter than the time for either learner to take a single step.
The batch size used in all cases was $2 \times 192$. Each learner ran on 16 TPU chips.

\subsection{Evaluation Actors}

There are two types of evaluation actors, which both sync parameters at the start of an episode:
\begin{enumerate}
    \item \textit{Single Player Online Evaluation Actors:} These actors run all the scripted probe evaluation tasks, with the current agent parameters used for solver inference and action choice. Procedural rewards are logged per episode.
    \item \textit{Dataset Evaluation Actors:} Similar to the dataset actors, these actors take episodes from the human data (training or validation, logged separately) and replay them from the perspective of setter or solver. Agent inference is run on the observations to get log probabilities of actions and various language output metrics.
\end{enumerate}

\section{Evaluation models} 
\label{app:eval_models}

As discussed in Section~\ref{sec:offline_eval}, one way in which we could measure our progress is to have humans directly score how often our agents are successful at completing instructions.
However, collecting human annotations is relatively expensive, and in order to accelerate progress it is desirable to have an automated method for evaluating agent performance. 
Automated evaluations can be employed in several ways: 
\begin{itemize}
    \item They can be used to remove poor quality human demonstration data before we apply imitation learning approaches;
    \item They can be used to perform hyperparameter tuning for imitation learning architectures and algorithms;
    \item They can be used to produce reward to optimise agent performance using reinforcement learning.    
\end{itemize}
We trained supervised models to predict labels given by human annotators who viewed episodes. The models themselves observed strided or decimated sequences of observations to reduce model size.
We chose to predict a binary success/failure label for each episode as a simple, albeit not completely general, approach to evaluation. We found there was a high degree of agreement among human annotators for this type of score on our dataset (about 85-90\%; see Table~\ref{tab:rater_agreement}).
In this work, we focused on building models to evaluate solver behaviour only.
This section presents a detailed view of the evaluation model architecture presented in the main text, the different models with which we experimented, the process we used to select our best models, and additional results.

\subsection{Architecture}

The description of the evaluation model architecture can be divided into three parts: processing the inputs, constructing the model, and defining the losses to optimise. Processing the inputs transforms trajectories of observations into a format that the model can efficiently ingest, while defining the model and losses connect the different modalities in the observations to evaluate if an episode was successful.

\subsubsection{Inputs}
Each episode consists of a sequence of frames, a single setter instruction, and a single solver language emission. We used a majority vote across all human annotations of an episode to determine the label.

The inputs are processed as follows:

\begin{itemize}
\item \textbf{Video:} we selected $x$ frames (where $x$ is a hyperparameter with default $x=32$) evenly spaced, starting at the index of the setter instruction and ending at the end of the episode.
\item \textbf{Setter Instruction:} we take the first setter emission, use the same typo correction system used in the agent, and pad with zeros to fill 16 tokens.
\item \textbf{Solver Emission:} we take the first solver emission, use the same typo correction system used in the agent, and pad with zeros to fill 10 tokens. 
\item \textbf{Binary Reward:} we binarised the reward sketches by labeling a sketch as a success if any frame of the sketch passed the success threshold.
We then took the majority vote across all annotations for a single episode if we had multiple sketches. 
\item \textbf{Binarised Evaluation Sequence:} for moment of success prediction, we reduce the annotation sequences down to a one-hot encoding of moment of success of length $\text{num-frames-selected} + 1$. The $1$ occurred at the time index of the first frame on or after the median moment of success marked in the reward sketches, or at the last index if the episode was unsuccessful. Note: this was only used for the success frame prediction loss. 
\end{itemize}

Because the human training data was heavily imbalanced, with the vast majority of episodes being successful, we constructed batches of episodes (default batch size was $32$) by selecting an equal number of successful episodes and unsuccessful episodes.

\subsubsection{Models}

One of the biggest challenges in developing evaluation models is that we had long episodes with multiple modalities to combine: video frames, setter instruction, and solver emission. The model thus had to learn to determine what constituted success for a particular instruction based on the video and the solver emission (in the question-answering case). We explored different model architectures to aid in solving this problem in a way that generalised from human episodes to agent episodes.

One of our models was based on a ResNet architecture. This model first computed embeddings for each of the modalities: video, setter instruction, and solver emission. For the vision stack, we had a hyperparameter controlling whether to use a standard ResNet-50 \citep{he2016deep} or a TSM ResNet, which adds a temporal shift module inside the residual block \citep{lin2019tsm}. We used the standard dm-haiku embedding module \citep{haiku2020github} to calculate an embedding for the setter instruction and an embedding for solver emission. We then had two methods for combining modalities:
\begin{enumerate}
        \item \textit{Concatenation:} Concatenate the embeddings of each modality, then pass the concatenated embeddings through an MLP head to get the output of the model.
        \item \textit{Product:} Multiply the embeddings of each modality, then take the mean across the embedding size as the output of the model. 
\end{enumerate}
    
Another of our models was a transformer-based architecture. In addition to the three inputs from video, setter instruction, and solver emission, we additionally introduced two dummy embeddings analogous to the CLS input in BERT \citep{devlin2018bert}. For the setter instruction and solver emission embeddings, the token embedding for each modality used a separate learnable parameter lookup embedding, with embedding dimension $512$, with the same vocabulary as used by the agent architecture. The embedding of the video frames was produced by a ResNet-50 \citep{he2016deep}, where the normal output was replaced with a $512$ dimensional vector. We concatenate the embeddings from all modalities and added to them segment and position embeddings to form a total embedding. The segment and position embeddings were also learnable embeddings, with dimension $512$. The segment embedding encoded which of the four modalities the input was from. The position embedding encoded the position in the sequence, with frames and words appearing in time order. Correspondingly, the vocab sizes was $4$ for the segment embeddings and $60$ for the position embeddings (the sum of number of frames, 32, setter instruction length, 16, solver emission length, 10, and dummy inputs, 2) respectively. The total embedding was then passed through a transformer with 16 self-attention heads, and 16 transformer-block layers, without dropout. We use the same transformer block as in \citep{radford2019language}, except we used standard rather than masked attention. We took a mean over the non-dummy outputs, and concatenated this with the dummy outputs, then flattened the result before passing it through an MLP head with $2$ hidden layers each of size $512$. We trained with batch size 32. We grid searched over learning rates $3e^{-3}, 1e^{-3}, 3e^{-4}, 1e^{-4}$. In the next section, we will describe the losses in more detail. For this model, we compared relative weightings of success loss to language matching loss of $0., 0.5, 1.0$.

\subsubsection{Losses}

In addition to the standard supervised loss, we compared two auxiliary loss options whose weighting was controlled by hyperparameters. These auxiliary losses helped the model to learn better representations and generalise to unseen episodes. We computed these losses in the same place as the standard supervised loss by passing augmented batches through the model (and potentially adding a separate head), then we summed the weighted losses. 
\begin{enumerate}
\item \textit{ELM loss:} The full-episode variant of the language matching loss, as defined in Equation~\ref{eq:elangmatch}, was computed on only successful episodes in the batch, yielding a batch size equal to half the total batch size. We augmented the batch by shuffling the instruction field for half of the successful episodes, holding the video and the solver emission field constant. We then used a boolean array denoting whether the language instruction field was shuffled or not as the targets. For the concatenation version of the model, another hyperparameter determined whether or not to share the same weights for the success MLP head and the language matching MLP head.
\item \textit{Success frame prediction loss:} The success frame prediction loss helps the model overcome the difference in distribution of human episodes and agent episodes. In human episodes, the moment of success is skewed towards the end of the episode, whereas in agent episodes the moment of success is skewed towards the beginning of the episode (for more details, see Section~\ref{app:eval_model_additional_if_results} below). We computed the success frame prediction loss by using a separate MLP head to predict a sequence of length $\text{num-frames-selected}$ where a $1$ at index $i$ signifies that success occurred at sampled frame $i$. We then use a cross entropy loss to classify the moment of success we derived from the reward sketches, computing the loss on successful episodes only. This loss was only used on the ResNet-based evaluation model.
\end{enumerate}

\subsection{Model Selection}
\label{app:eval_models_selection}

In Table~\ref{tab:evaluation_model_comparison}, all of the evaluation models are listed, with architectural details, active losses, and number of input observations.

\begin{table}[H]
    \centering
    \small
    \setlength{\tabcolsep}{1.2pt}
    \begin{tabular}{ccccccc}
    \toprule
    Name           & TSM & \multirow{2}{2cm}{\centering Concat/ Product} & \multirow{2}{2cm}{\centering ELM\\ Loss} & \multirow{2}{2cm}{\centering Success Frame \\Loss} & Transformer & \multirow{2}{2cm}{\centering Number of \\Frames} \\ \\
    \midrule
    \textrm{RC$\cdot$S$\cdot$Tr}        & \no & C  & \yes  & \no & \yes  & 32 \\
    \textrm{RCT$\cdot$S$\cdot$SF}       & \yes & C  & \yes  & \yes & \no  & 32 \\
    \textrm{RC$\cdot$S}          & \no & C  & \yes  & \no & \no  & 32 \\
    \textrm{RP$\cdot$L}          & \no & P  & \yes  & \no & \no  & 48 \\
    \textrm{RPT$\cdot$L}          & \yes  & P & \yes  & \no  & \no  & 48 \\
    \textrm{RPT$\cdot$S}          & \yes & P & \yes  & \no  & \no  & 32 \\
    \textrm{RCT$\cdot$S}         & \yes & C & \yes  & \no & \no  & 32 \\
    \textrm{RC$\cdot$S$\cdot$Tr (no ELM)} & \no & C & \no  & \no & \yes  & 32 \\
    \textrm{RPT$\cdot$S (ELM only)} & \yes & P & \yes  & \no & \no  & 32 \\
    \textrm{RC$\cdot$S$\cdot$Tr (ELM only)} & \no & C & \yes  & \no & \yes  & 32 \\
    \textrm{RC$\cdot$S (no ELM)} & \no & C & \no  & \no & \no  & 32 \\
    \bottomrule
    \end{tabular}
    \caption{\textbf{Evaluation Model Property List.} We name the models based on the features they contain, where R denotes using a ResNet to embed the video frames, C or P denotes the method used to combine modalities (concatenation or product), T denotes using TSM, S or L denotes the length of the video (short=$32$ frames or long=$48$ frames), SF denotes using the success frame prediction loss, and Tr denotes using the transformer-based architecture.}
    \label{tab:evaluation_model_comparison}
\end{table}

We used a ``validation score'' to both select the best model among those presented in Table~\ref{tab:evaluation_model_comparison} and to select the best hyperparameter combination per model. 
The formula for the validation score was as follows:
\begin{align*}
 \text{validation-score} & =  0.5 \times \text{balanced-acc-human} + 0.25 \times \text{balanced-acc-weak-agent} \\ & \, \, \, \, \, + 0.25 \times \text{balanced-acc-strong-agent},
\end{align*}
where $\text{weak-agent}$ and $\text{strong-agent}$ were previously trained agents. 
We selected the model, best hyperparameter combination (including the model's threshold for success), and model training step from smoothed online evaluation of the validation score. 

\begin{figure}[H]
    \centering
    \includegraphics[scale=1]{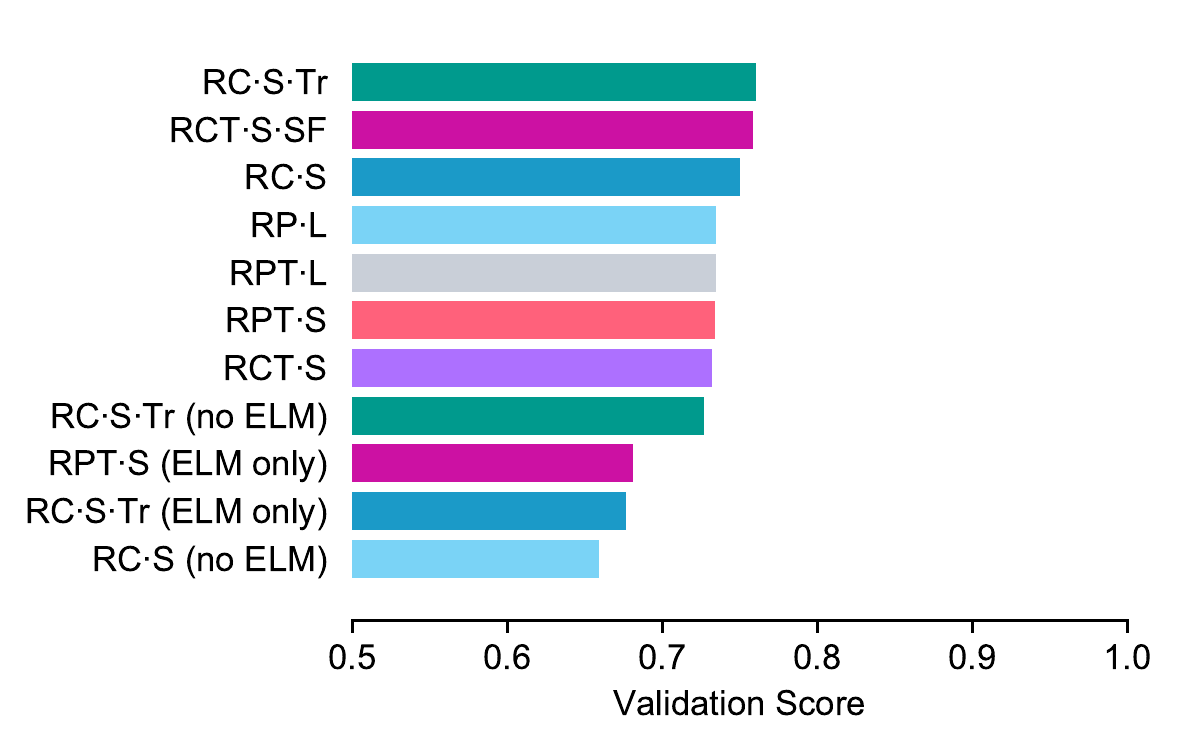}
    \caption{Validation scores for model ablations.}
\end{figure}

\subsection{Additional Instruction-Following Results}
\label{app:eval_model_additional_if_results}

The success frame prediction loss, in conjunction with the TSM, yielded a model almost as good as the transformer model. The validation score was only slightly lower for the success frame model, and it achieved higher balanced accuracy measures for some of the test agents. We hypothesise that the combination of transformer based models with success prediction may produce stronger models, though we have not tested this here.

\begin{figure}[H]
    \centering
    \includegraphics[width=\textwidth]{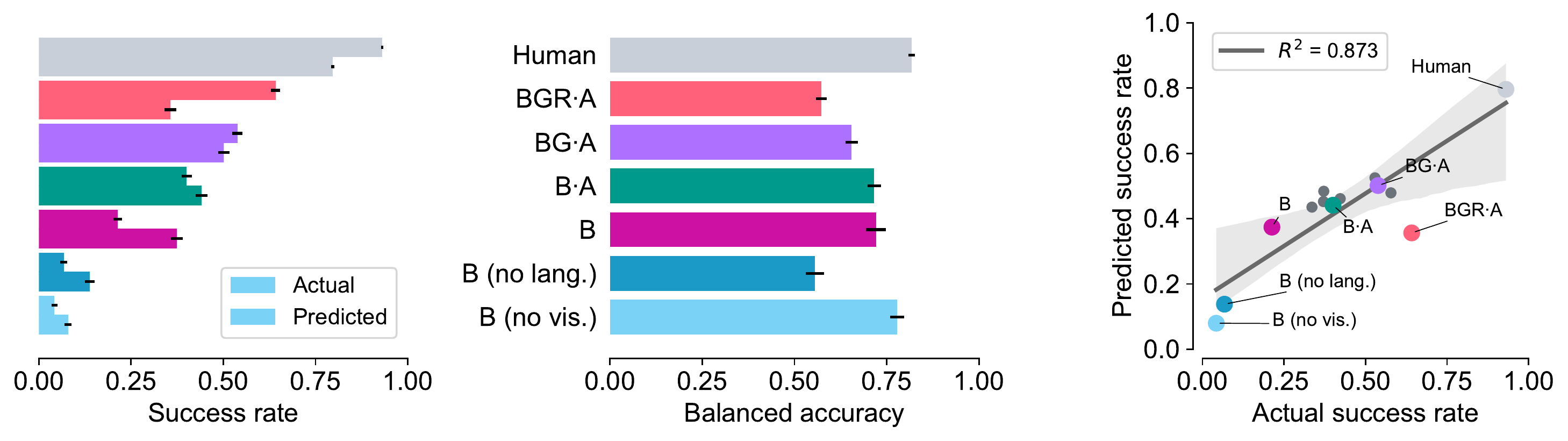}
    \caption{Success frame prediction results.}
\end{figure}

\begin{figure}[H]
    \centering
    \includegraphics[width=\textwidth]{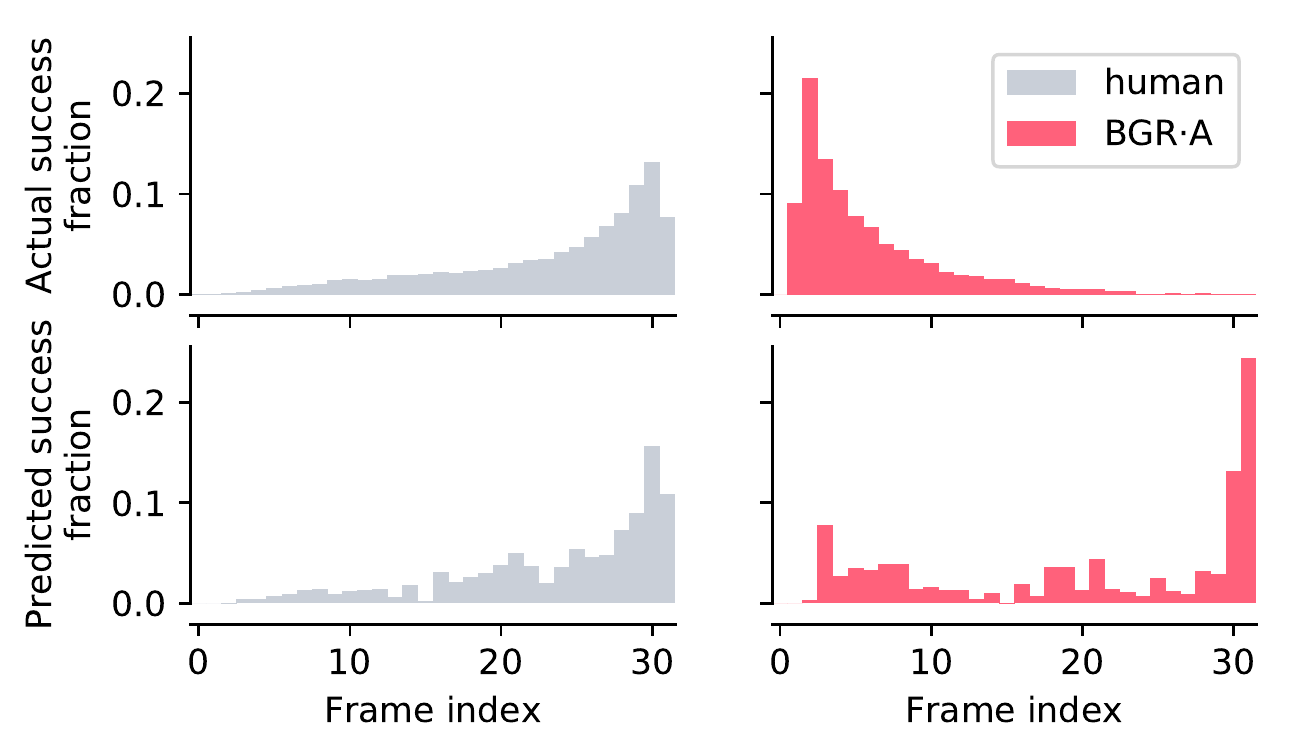}
    \caption{\textbf{Moment of success prediction.} The evaluation model predicts the moment of success well for human data (on which it's trained), and is able to partially overcome the distributional shift on previously unseen agent data. Frames are indexed after decimating the video.}
    \label{fig:success_prediction}
\end{figure}

In the human demonstrations, the episodes were stopped shortly after the moment of success, leading to a distribution of moment of success that was heavily skewed towards the end of the episode (see Figure~\ref{fig:success_prediction}). 
In the agent episodes, however, episodes were run for a fixed length of time, skewing the distribution of moment of success to earlier in the episode. 
Thus, since our evaluation models were trained only on human episodes, this distribution mismatch presents a challenge for the evaluation models. 
The success frame prediction loss encourages the model to understand the moment of success regardless of when success happens within the episode.

\subsection{Question-Answering Results}

\begin{figure}[H]
    \centering
    \includegraphics[width=\textwidth]{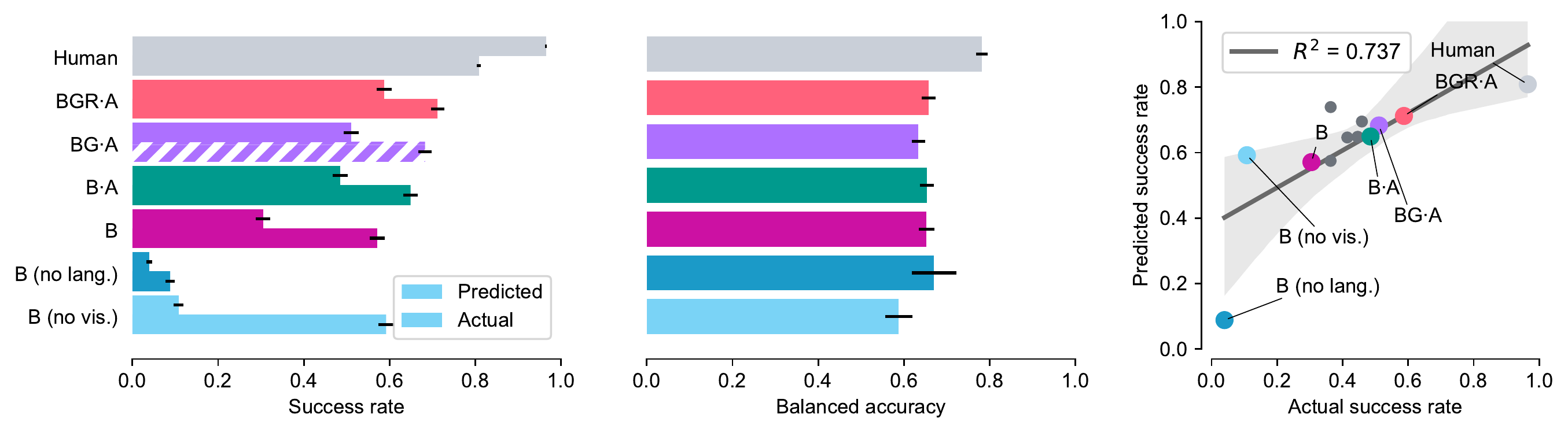}
    \caption{Question-Answering results.}
    \label{fig:qa_eval}
\end{figure}

The results previously highlighted focused on instruction-following tasks. 
Ultimately we want to develop evaluation models that work across both instruction-following and question-answering tasks. 
The question-answering domain is more difficult than the instruction-following domain for several reasons, including the challenge of combining three modalities (instruction, video, and response). 
Figure~\ref{fig:qa_eval} shows the results with an early model developed to evaluation question-answering episodes.

For this model, we use a ResNet augmented with FiLM \citep{perez2018film}, which has elsewhere proved effective for visual reasoning tasks. 
Otherwise, the model was constructed according to the ResNet Product model with 32 decimated frames and the auxiliary ELM loss. 
Though the ordering of agents it produces is close to correct, the balanced accuracy per agent and correlation are significantly worse than the models for instruction-following tasks. 
Additionally, the success rate per agent was overpredicted, likely because the human episodes used as training data had few negative examples. 

\section{Automated Evaluation Metrics}
\label{app:auto_metrics}
    
\subsection{Setter Language Metrics}
\label{app:setter_metrics}
To evaluate our setters' language output we used log probabilities as well as several heuristics that captured the agents' ability to refer to objects present in the scene. The \emph{object error rate} measures the fraction of setter sentences that mention an object type that exists in the room (e.g., a toy duck or a helicopter). The error rate captures the ability of the agent to recognise object categories but not necessarily their colour. 

To understand whether our setters are also able to detect the correct object colour, we introduced the \emph{colour object error rate}, which measures the proportion of setter sentences that mention valid objects with their true colour out of all the sentences that mention coloured objects. When calculating these error rates we also applied a mapping to canonicalise the colour vocabulary: e.g., mapping ``turquoise'' to ``blue.''

We used these metrics to measure training progress and to evaluate in an automated manner how contextually relevant setter instructions were. We acknowledge that these metrics do not capture many desiderata in language output. For example, they express nothing about the relative positioning of objects; errors related to descriptions of spatial relationships were undetected in our automated metrics.   

\subsection{Procedural Tasks}
\subsubsection{Automated Metrics}
We designed a set of automated metrics for evaluating solver agent performance during training as well. These metrics compared an agent trajectory to a human demonstration in a setter replay episode, with an instruction and inital room pulled from the human dataset. In many of our language games, holding and lifting specific objects was part of the human demonstration. In the ``first-object-lifted by-both'' metric, we checked whether the first object lifted by the agent was the same as that lifted by the human in the corresponding episode. This was a useful correlational indicator of agent performance. However, multiple objects can exist in the environment that satisfy the instruction, and some episodes involve no such lifting instructions, so it was only useful for relative comparisons across agents and during training. We additionally computed metrics that measured if the colour and type of the two objects lifted were the same, even if not the exact same instance, in the ``number-of-colour-objects-lifted-by-both’' metric. We adjusted this metric to account for the average number of objects that the human and agent lifted within an episode for each language game.

\subsubsection{Scripted Probe Tasks}
\label{app:probe_tasks}
Our agents were not trained in reinforcement learning environments providing unambiguous rewards for reaching particular goal states. Nevertheless, we found it valuable to program a small set of such tasks for the purpose of evaluating our agents since the reward function provides an objective and repeatable measure of success. In each of these tasks, as in the agent's training environment, the distribution of objects and the initial position of the agent were randomised on a per-episode basis. We typically ran trained agents for $1,000$ episodes in each task to obtain an estimate of its expected success rate.  

\paragraph{Go Somewhere} After a random delay of up to 10 seconds, an instruction is presented to the agent of the form \texttt{Go near the X}, where \texttt{X} can be an object (ball, teddy bear, box), furniture (table, bed) or landmark (door, window, shelf). The agent must move to within 1m of the target before the 30 second episode time limit is reached. 
\paragraph{Lift Something} After a random delay of up to 10 seconds, an instruction is presented to the agent of the form \texttt{Lift an X}, where \texttt{X} is a movable object (ball, teddy bear, box etc.). The agent must pick the object up higher than 1m. If the agent at any point picks up an object that is not an X, or if the 2 minute time limit is reached, the episode ends with a score of zero. If the agent picks up an X, the episode also ends, with a score of one. 

\paragraph{Position Relative} The environment checks the episode's initial conditions to ensure that no objects of type X are within 1m of any objects of type Y. After a random delay of up to 10 seconds, an instruction is presented to the agent of the form \texttt{Put an X near a Y}, where \texttt{X} is a movable object (ball, teddy bear, box etc.) and \texttt{Y} is a movable object or furniture (bed, table etc.). If at any timestep an object of type X is within 1m of an object of type Y, the episode ends with a score of one. Otherwise, the episode ends after 2 minutes with a score of zero. 

\paragraph{Ask about Colour} After a random delay of up to 10 seconds, an question is presented to the agent of the form \texttt{What colour is the X}, where \texttt{X} is a movable object (ball, teddy bear, box etc.) or furniture (bed, table  etc.). The episode ends when the agent generates some language or when the 2 minute time limit is reached. If the agent's response is among the set of allowed correct answers for the episode (corresponding to the colour of object X), the score is one, otherwise it is zero. 

\paragraph{Ask about Existence} After a random delay of up to 10 seconds, an question is presented to the agent of the form \texttt{Is there an  X in the room}, where \texttt{X} is a movable object (ball, teddy bear, box etc.) or furniture (bed, table  etc.). The episode ends when the agent generates some language or when the 2 minute time limit is reached. If the agent's response is correct (either \emph{yes} or \emph{no}), a score of one is awarded. If the response is incorrect or the agent does not respond within the 2 minute time limit, the score is zero. Note that the task generator is designed to construct approximately equal numbers of episodes for which the correct answer is yes and no. 

\paragraph{Count Something} After a random delay of up to 10 seconds, an question is presented to the agent of the form \texttt{How many X are there  in the room}, where \texttt{X} is a movable object (ball, teddy bear, box etc.) or furniture (bed, table  etc.). The episode ends when the agent generates some language or when the 2 minute time limit is reached. If the agent's response is correct (correct answers range between zero and five), a score of one is awarded. If the response is incorrect or the agent does not respond within the 2 minute time limit, the score is zero. Note that the distribution of answers across the set $\{0 \dots 5\}$ is not uniform, with \emph{zero}, \emph{one} and \emph{two} being the most frequent responses, so comparing agent scores to baseline controls (e.g. blind, language-only agents) is important for valid interpretation of the scores.

Numerical performance of each agent on these tasks is show in Table~\ref{tab:scripted_probes}.

\begin{table}[ht]
    \small
    \centering
    \begin{tabular}{rcccccc}
\toprule
{} & Colour & Existence & Count & Go & Lift & Position \\
\midrule
Human & $0.73$ & $0.8$ & $0.85$ & $0.84$ & $0.93$ & $0.65$  \\
\mainagent{} & $\mathbf{0.57}$ & $\mathbf{0.72}$ & $0.28$ & $\mathbf{0.69}$ & $\mathbf{0.79}$ & $\mathbf{0.37}$  \\ 
\BGA{} & $0.47$ & $0.61$ & $0.32$ & $0.64$ & $0.76$ & $0.32$ \\ 
\BA{} & $0.43$ & $0.69$ & $\mathbf{0.34}$ & $0.47$ & $0.55$ & $0.19$  \\ 
\B{} & $0.18$ & $0.59$ & $0.27$ & $0.35$ & $0.13$ & $0.14$ \\ 
\BnoL{} & $0.04$ & $0.03$ & $0.02$ & $0.19$ & $0.04$ & $0.12$ \\ 
\BnoV{} & $0.14$ & $0.56$ & $0.28$ & $0.17$ & $0.0$ & $0.02$ \\ 
\bottomrule
\end{tabular}

    \caption[Agent performance on scripted probe tasks]{
    \textbf{Agent performance on the scripted probe tasks}. Standard errors were all less than $0.03$. \\}
    \label{tab:scripted_probes}
\end{table}

\section{Scaling Experiments}
In this set of experiments we studied the scaling properties of our main models. Focusing on the scripted probe tasks, we quantified the performance of our agents as we increased the amount of training data. We ran these experiments for the two models \BA{} and \BGA{}. We controlled the size of the datasets by randomly subsampling our original data, resulting in the following number of episodes for each fraction of the complete dataset:
\begin{center}
\begin{tabular}{ r c c c c c }
 Relative size & 1 & $1/2$ & $1/4$ & $1/8$ & $1/16$ \\
 Number of episodes & 548K & 274K & 137K & 68K & 34K   \\
\end{tabular}
\end{center}
Figure~\ref{fig:scaling}A shows the accumulated reward averaged over instruction-following (\emph{lift something, go somewhere and position relative to}) and question-answering (\emph{ask about color, ask about existence and count something}) scripted probe tasks. For both, \BA{} and \BGA{}, the performance increased smoothly with the amount of data. However, particularly for the instruction-following levels, the rate of increase in performance for the agent using GAIL (\BGA{}) was higher than for the agent trained with BC (\BA{}).

\section{Transfer Experiments} 
In this set of experiments we tested an agents' ability to generalise behaviour to situations that are either unseen or rare in the training data. We used a slightly smaller network (4 layers instead of 8 in the multi-modal transformer and an embedding size of 256 instead of 512) for this set of experiments compared to our default setup, however, we reran relevant baselines in this setup in order to maintain fair comparisons. We used the \BGA{} agent in all of these experiments.

\subsection{Multi-task Transfer}
The next group of experiments aimed to evaluate transfer of agent skills across different tasks. We studied two aspects of transfer learning: 1) Do agents trained on multiple task perform better than agents trained in each task separately? and 2) Does training on multiple tasks lead to agents that require less data to learn a new task? To answer these questions, we performed two sets of experiments: one where we \emph{varied the type of task we trained on}, and one where we \emph{varied the amount of data we saw from within a task}.

\subsubsection{Single-task versus Multi-task}
In the first setup we trained our agent on a subsets of the data containing only a particular task (such as \emph{lift something}, or \emph{ask about color}), and compared the performance on these tasks against an agent trained on all data from all tasks. Figure~\ref{fig:scaling}B shows that using data from all tasks yielded higher rewards on the scripted probe tasks than the single-task case. This occurred presumably because motor behaviours like navigation of the room and grasping objects, along with linguistic knowledge like representations of object names, transferred across tasks.

\subsubsection{Multi-task Data Efficiency}
In this experiment, we wanted to analyse whether training on multiple tasks led to agents that required less data to learn a new task. We studied this by entirely removing data from a particular task and adding it back in controlled amounts. 
Specifically, we tested how much data was required to learn the behaviour of positioning an object relative to another by transferring skills learned from other types of interactions (such as lifting and counting). We constructed a ``position'' dataset of 73K episodes with human instructions containing one of the verbs ``put,'' ``place,'' or ``position'' and excluded these episodes from the pool of 548 background tasks. Then, we incrementally added some proportion of the ``position'' dataset ($1/8$, $1/4$, and $1/2$) to these background tasks, effectively creating several new datasets with a different amount of ``position'' data. 

Figure~\ref{fig:scaling}C shows that with as little as $1/8$th of the ``position'' data, the agent could easily learn to position objects relative to one another provided that it was also exposed to the background of multi-task data from other tasks during training. As we increased the amount of positioning data, performance improved for both the single and multi-task cases. However, to achieve good scores on the scripted probe tasks, we needed much less data when we trained on all tasks in the multi-task condition.

\subsection{Colour-Object Generalisation}
To check how well our agents generalised over different features of the environment, we performed the following experiment: we removed all occurrences of a certain colour-object combination from the environment during training (both from the dataset and the environment), we created probe tasks that require the agent to interact with the held-out coloured object, and  we compared the performance on these tasks with agents trained without any object restrictions. Because the agent was exposed to that particular colour in other objects and to that particular object in other colours, we were testing to what extent it could generalise to unseen combinations of known features.

Without loss of generality, we picked orange ducks as the held-out coloured object. We removed all instances of it from the training data and from the environment, and we created scripted probe tasks that refer to this object in particular (e.g. \emph{lift the orange duck} or \emph{what is the color of the duck?}) 

There were approximately $23$K episodes in the whole dataset across all language games that contained orange ducks. Note that to exclude relevant episodes, we inspected ground truth information about what objects were in the Playroom instance.

We designed special-case scripted probe tasks to evaluate the agents that were similar to the previously described scripted probe tasks (lift something, ask about color, position relative to, and go somewhere). In this case, they specifically examined performance with respect to this particular colour-object combination. Each task was balanced by providing appropriate distractors. For example, when asking to lift an orange duck, there was always a non-orange duck in the room as well. This guaranteed that the agent did not just interact with this type of object regardless of its colour.

Figure~\ref{fig:scaling}D summarises the results. Overall, our agents were able to successfully interact with the colour-object combinations that they had never been exposed to in the training data, with only a small performance drop across all games we evaluated on compared to the control experiment that used an equal total number of episodes but no restrictions on the types of objects seen in the data. 

\section{Vocabulary and Spelling Correction Table}
\label{sec:vocab}
\subsection{Vocabulary}
\begin{footnotesize}
\begin{flushleft}
\begin{multicols}{8}
a\\
able\\
about\\
above\\
activity\\
adjacent\\
after\\
again\\
air\\
airplane\\
all\\
along\\
also\\
am\\
among\\
an\\
and\\
animal\\
another\\
answer\\
any\\
anything\\
anywhere\\
aqua\\
aquamarine\\
are\\
arent\\
arm\\
armchair\\
around\\
arrange\\
arrangement\\
article\\
as\\
aside\\
ask\\
at\\
available\\
away\\
baby\\
back\\
bad\\
ball\\
basket\\
basketball\\
bat\\
be\\
bear\\
bed\\
before\\
behind\\
below\\
beneath\\
beside\\
between\\
big\\
bigger\\
biggest\\
bike\\
bin\\
black\\
block\\
blue\\
board\\
boat\\
book\\
bookshelf\\
both\\
box\\
brighter\\
bring\\
brown\\
bus\\
bush\\
but\\
by\\
cabinet\\
cactus\\
can\\
car\\
carriage\\
case\\
cat\\
catch\\
ceiling\\
center\\
chair\\
change\\
check\\
chest\\
choice\\
circle\\
circular\\
clean\\
clear\\
clock\\
close\\
closer\\
closet\\
coffee\\
collect\\
color\\
come\\
common\\
compare\\
comparing\\
compartment\\
consist\\
contain\\
container\\
corner\\
correct\\
cot\\
couch\\
could\\
count\\
cream\\
cube\\
cup\\
cupboard\\
currently\\
cushion\\
cyan\\
cylinder\\
cylindrical\\
dance\\
dark\\
darker\\
describe\\
desk\\
diagonal\\
did\\
difference\\
different\\
dining\\
direction\\
disturb\\
disturbing\\
do\\
does\\
dog\\
doll\\
door\\
down\\
drag\\
drawer\\
drier\\
drop\\
duck\\
each\\
ear\\
ears\\
edge\\
eight\\
eighteen\\
eighty\\
either\\
elevate\\
eleven\\
else\\
empty\\
end\\
engine\\
enter\\
equal\\
ever\\
every\\
everything\\
exact\\
exactly\\
except\\
exist\\
existed\\
existing\\
explain\\
face\\
facing\\
falling\\
fan\\
far\\
favorite\\
feet\\
few\\
fifteen\\
fifty\\
find\\
finished\\
first\\
five\\
flight\\
flip\\
floor\\
flower\\
flying\\
foot\\
football\\
for\\
forty\\
four\\
fourteen\\
frame\\
from\\
front\\
gather\\
get\\
give\\
go\\
good\\
grab\\
gray\\
green\\
grey\\
ground\\
group\\
had\\
hair\\
hairdryer\\
hand\\
handle\\
hang\\
has\\
have\\
he\\
head\\
headphone\\
headphones\\
height\\
helicopter\\
help\\
here\\
hide\\
hit\\
hold\\
how\\
human\\
hundred\\
i\\
identical\\
if\\
im\\
in\\
including\\
inside\\
instrument\\
into\\
is\\
isnt\\
it\\
item\\
its\\
jug\\
jump\\
just\\
keep\\
key\\
keyboard\\
kick\\
know\\
lamp\\
large\\
larger\\
largest\\
lavender\\
laying\\
ledge\\
left\\
leg\\
legged\\
less\\
lies\\
lift\\
light\\
lighter\\
like\\
liked\\
line\\
located\\
location\\
locomotive\\
long\\
look\\
looking\\
lying\\
magenta\\
main\\
make\\
man\\
many\\
maroon\\
mattress\\
maximum\\
may\\
me\\
mention\\
mess\\
middle\\
mine\\
mirror\\
moment\\
more\\
most\\
mostly\\
move\\
moving\\
much\\
mug\\
musical\\
my\\
name\\
navy\\
near\\
nearer\\
nearest\\
neat\\
next\\
nine\\
nineteen\\
no\\
non\\
none\\
not\\
notice\\
noticed\\
noticing\\
now\\
number\\
object\\
objective\\
observing\\
of\\
off\\
olive\\
on\\
one\\
only\\
onto\\
opposite\\
or\\
orange\\
order\\
other\\
out\\
oval\\
over\\
own\\
paint\\
painted\\
painting\\
pale\\
parallel\\
parrot\\
pass\\
peach\\
pen\\
perform\\
phone\\
photo\\
piano\\
pick\\
picture\\
pillow\\
pink\\
place\\
placed\\
plane\\
plant\\
play\\
player\\
please\\
pointed\\
pointy\\
position\\
possess\\
pot\\
potted\\
present\\
purple\\
push\\
put\\
question\\
rack\\
racket\\
rail\\
rectangle\\
rectangular\\
red\\
refer\\
relative\\
remove\\
replace\\
right\\
roam\\
robot\\
rocket\\
roof\\
room\\
round\\
rounded\\
row\\
rubber\\
run\\
same\\
say\\
sea\\
see\\
seeing\\
seen\\
self\\
sequence\\
set\\
setter\\
seven\\
seventeen\\
seventy\\
shape\\
shaped\\
sheet\\
shelf\\
shift\\
ship\\
show\\
side\\
silver\\
similar\\
simple\\
single\\
sit\\
situated\\
six\\
sixteen\\
sixty\\
size\\
sizes\\
sky\\
small\\
smaller\\
smallest\\
so\\
soccer\\
soda\\
sofa\\
some\\
something\\
somewhere\\
spacious\\
specific\\
square\\
squared\\
stand\\
standing\\
staring\\
stool\\
storage\\
straight\\
study\\
table\\
take\\
takeoff\\
taller\\
task\\
tea\\
teal\\
teddy\\
tell\\
ten\\
tennis\\
than\\
that\\
thats\\
the\\
their\\
them\\
then\\
there\\
these\\
they\\
thing\\
think\\
thirteen\\
thirty\\
this\\
those\\
thousand\\
three\\
through\\
throw\\
time\\
to\\
together\\
top\\
total\\
touch\\
touching\\
towards\\
toy\\
train\\
tree\\
triangle\\
try\\
tube\\
turn\\
turquoise\\
tv\\
twelve\\
twenty\\
two\\
under\\
underneath\\
until\\
up\\
upon\\
upside\\
use\\
used\\
using\\
van\\
vehicle\\
very\\
view\\
violet\\
visible\\
wait\\
walk\\
wall\\
want\\
wardrobe\\
was\\
watching\\
way\\
we\\
were\\
what\\
whatever\\
wheels\\
when\\
where\\
whether\\
which\\
white\\
window\\
wise\\
wish\\
with\\
without\\
wooden\\
word\\
yellow\\
yes\\
you\\
your\\
yourself\\
\end{multicols}
\end{flushleft}
\end{footnotesize}
\subsection{Typo Table}
The custom typo dictionary is enumerated below. Please note that this also ``corrects'' for pluralisation. Future work will explore the use of subword tokenisation to better construct and learn large vocabularies:
\begin{footnotesize}
\begin{flushleft}
\begin{multicols}{4}
'-yesno': 'yes or no'\\
'03': 'three'\\
'0bjects': 'object'\\
'1': 'one'\\
'10': 'ten'\\
'100': 'one hundred'\\
'1000': 'one thousand'\\
'10000': 'ten thousand'\\
'100objects': 'one hundred object'\\
'10objects': 'ten object'\\
'11': 'eleven'\\
'111': 'eleven'\\
'12': 'twelve'\\
'120': 'twelve'\\
'13': 'thirteen'\\
'14': 'fourteen'\\
'15': 'fifteen'\\
'150': 'fifteen'\\
'16': 'sixteen'\\
'17': 'seventeen'\\
'18': 'eighteen'\\
'19': 'nineteen'\\
'2': 'two'\\
'20': 'twenty'\\
'200': 'two hundred'\\
'2000': 'two thousand'\\
'20teddys': 'twenty teddy bear'\\
'21': 'twenty one'\\
'22': 'twenty two'\\
'23': 'twenty three'\\
'24': 'twenty four'\\
'25': 'twenty five'\\
'26': 'twenty six'\\
'27': 'twenty seven'\\
'28': 'twenty eight'\\
'29': 'twenty nine'\\
'2objects': 'two object'\\
'3': 'three'\\
'30': 'thirty'\\
'300': 'three hundred'\\
'32': 'thirty two'\\
'33': 'thirty three'\\
'34': 'thirty four'\\
'35': 'thirty five'\\
'36': 'thirty six'\\
'3objects': 'three object'\\
'4': 'four'\\
'40': 'forty'\\
'42': 'forty two'\\
'43': 'forty three'\\
'45': 'forty five'\\
'46': 'forty six'\\
'4objects': 'four object'\\
'5': 'five'\\
'50': 'fifty'\\
'500': 'five hundred'\\
'5000': 'five thousand'\\
'50objects': 'fifty object'\\
'52': 'fifty two'\\
'55': 'fifty five'\\
'5balls': 'five ball'\\
'5objects': 'five object'\\
'6': 'six'\\
'60': 'sixty'\\
'69': 'sixty nine'\\
'7': 'seven'\\
'70': 'seventy'\\
'75': 'seventy five'\\
'8': 'eight'\\
'88': 'eighty eight'\\
'9': 'nine'\\
'90': 'ninety'\\
'`place': 'place'\\
'a': 'a'\\
'aa': 'a'\\
'aany': 'any'\\
'abed': 'a bed'\\
'abject': 'object'\\
'abjects': 'object'\\
'able': 'able'\\
'abll': 'a ball'\\
'ablue': 'a blue'\\
'about': 'about'\\
'above': 'above'\\
'ac': 'a'\\
'according': 'according'\\
'acircle': 'a circle'\\
'acorner': 'a corner'\\
'acr': 'car'\\
'action': 'action'\\
'active': 'active'\\
'activity': 'activity'\\
'adjacent': 'adjacent'\\
'adjust': 'adjust'\\
'adn': 'and'\\
'aero': 'airplane'\\
'aeroplain': 'airplane'\\
'aeroplane': 'airplane'\\
'aeroplanes': 'airplane'\\
'after': 'after'\\
'again': 'again'\\
'against': 'against'\\
'aid': 'aid'\\
'air': 'air'\\
'aircraft': 'airplane'\\
'aircrafts': 'airplane'\\
'airplane': 'airplane'\\
'airplanes': 'airplane'\\
'airport"': 'airplane'\\
'al': 'all'\\
'aline': 'a line'\\
'all': 'all'\\
'alla': 'all of'\\
'alll': 'all'\\
'almirah': 'wardrobe'\\
'along': 'along'\\
'also': 'also'\\
'am': 'am'\\
'amd': 'and'\\
'ame': 'and'\\
'amny': 'many'\\
'among': 'among'\\
'an': 'an'\\
'and': 'and'\\
'andplace': 'and place'\\
'andstand': 'and stand'\\
'angle': 'angle'\\
'animal': 'animal'\\
'another': 'another'\\
'ans': 'and'\\
'answer': 'answer'\\
'ant': 'any'\\
'anu': 'any'\\
'any': 'any'\\
'anyof': 'any of'\\
'anyone': 'anyone'\\
'anyother': 'another'\\
'anything': 'anything'\\
'anywhere': 'anywhere'\\
'anyy': 'any'\\
'apart': 'apart'\\
'apink': 'a pink'\\
'approximately': 'approximately'\\
'aqua': 'aqua'\\
'ar': 'are'\\
'araange': 'arrange'\\
'arange': 'arrange'\\
'arch': 'arch'\\
'are': 'are'\\
'area': 'area'\\
'ared': 'a red'\\
'aree': 'are'\\
'areoplane': 'airplane'\\
'arew': 'are'\\
'arm': 'arm'\\
'armchair': 'armchair'\\
'armchairs': 'armchair'\\
'arocket': 'a rocket'\\
'aroe': 'a row'\\
'aroom': 'a room'\\
'around': 'around'\\
'arounf': 'around'\\
'arow': 'a row'\\
'arragement': 'arrangement'\\
'arrange': 'arrange'\\
'arrangement': 'arrangement'\\
'arrangements': 'arrangement'\\
'arrangment': 'arrangement'\\
'arre': 'are'\\
'arrrange': 'arrange'\\
'arte': 'are'\\
'articles': 'article'\\
'as': 'as'\\
'ash': 'grey'\\
'aside': 'aside'\\
'asimple': 'a simple'\\
'asingle': 'a single'\\
'ask': 'ask'\\
'assemble': 'assemble'\\
'at': 'at'\\
'atand': 'and'\\
'ate': 'at'\\
'athe': 'the'\\
'atleast': 'at least'\\
'att': 'at'\\
'auto': 'car'\\
'available': 'available'\\
'avalable': 'available'\\
'away': 'away'\\
'b': ''\\
'ba': 'bed'\\
'baal': 'ball'\\
'baasket': 'basket'\\
'baby': 'baby'\\
'back': 'back'\\
'backside': 'back'\\
'bad': 'bed'\\
'bag': 'bag'\\
'bal': 'ball'\\
'balcony': 'balcony'\\
'ball': 'ball'\\
'balll': 'ball'\\
'ballls': 'ball'\\
'balloon': 'balloon'\\
'balls': 'ball'\\
'bals': 'ball'\\
'baot': 'boat'\\
'bar': 'bear'\\
'bas': 'ball'\\
'bascket': 'basket'\\
'basket': 'basket'\\
'basketball': 'basketball'\\
'basketballs': 'basketball'\\
'basketbox': 'basket'\\
'baskets': 'basket'\\
'basketsboxes': 'basket'\\
'baskety': 'basket'\\
'bat': 'bat'\\
'bathroom': 'bathroom'\\
'bats': 'bat'\\
'bax': 'bat'\\
'bbed': 'bed'\\
'bbok': 'book'\\
'bd': 'bed'\\
'bde': 'bed'\\
'be': 'be'\\
'bear': 'bear'\\
'bears': 'bear'\\
'beat': 'beat'\\
'beautiful': 'beautiful'\\
'bec': 'bed'\\
'bed': 'bed'\\
'bed"': 'bed'\\
'bed-': 'bed'\\
'bedand': 'bed and'\\
'bedd': 'bed'\\
'bedds': 'bed'\\
'bede': 'bed'\\
'bedf': 'bed'\\
'bedlamp': 'bed lamp'\\
'bedright': 'bed right'\\
'beds': 'bed'\\
'bedsheet': 'bed sheet'\\
'bedyes': 'bed yes'\\
'bedyesno': 'bed yes no'\\
'bedyn': 'bed yes no'\\
'beed': 'bed'\\
'beer': 'bear'\\
'bef': 'bed'\\
'before': 'before'\\
'behind': 'behind'\\
'being': 'being'\\
'below': 'below'\\
'bench': 'bench'\\
'beneath': 'beneath'\\
'bera': 'bear'\\
'berd': 'bed'\\
'bes': 'bed'\\
'besdie': 'beside'\\
'besid': 'beside'\\
'beside': 'beside'\\
'besides': 'beside'\\
'besie': 'beside'\\
'best': 'best'\\
'bet': 'bed'\\
'betweeen': 'between'\\
'between': 'between'\\
'big': 'big'\\
'bigger': 'bigger'\\
'biggest': 'biggest'\\
'bike': 'bike'\\
'bin': 'bin'\\
'bins': 'bin'\\
'biplane': 'airplane'\\
'bird': 'duck'\\
'bix': 'box'\\
'bjects': 'object'\\
'bkue': 'blue'\\
'black': 'black'\\
'blackyesno': 'black yes no'\\
'ble': 'blue'\\
'bll': 'ball'\\
'block': 'block'\\
'blocks': 'block'\\
'blow': 'below'\\
'blu': 'blue'\\
'blue': 'blue'\\
'bluebox': 'blue box'\\
'bluish': 'blue'\\
'blur': 'blue'\\
'bluw': 'blue'\\
'bo': 'boat'\\
'boa': 'boat'\\
'boar': 'boat'\\
'board': 'board'\\
'boards': 'board'\\
'boart': 'boat'\\
'boat': 'boat'\\
'boats': 'boat'\\
'boc': 'box'\\
'bojects': 'object'\\
'bok': 'book'\\
'boll': 'ball'\\
'bolls': 'ball'\\
'boo': 'book'\\
'boojs': 'book'\\
'book': 'book'\\
'bookl': 'book'\\
'bookon': 'book on'\\
'books': 'book'\\
'bookshelf': 'bookshelf'\\
'bool': 'book'\\
'boook': 'book'\\
'boox': 'box'\\
'bot': 'robot'\\
'both': 'both'\\
'bothe': 'both'\\
'bottle': 'bottle'\\
'box': 'box'\\
'boxbasket': 'box'\\
'boxe': 'box'\\
'boxes': 'box'\\
'boxex': 'box'\\
'boxs': 'box'\\
'boxtill': 'box until'\\
'boxx': 'box'\\
'bpox': 'box'\\
'brd': 'bed'\\
'brig': 'bring'\\
'brighter': 'brighter'\\
'brightest': 'brightest'\\
'brimg': 'bring'\\
'brin': 'bring'\\
'bring': 'bring'\\
'broen': 'brown'\\
'brow': 'brown'\\
'brown': 'brown'\\
'bsket': 'basket'\\
'bsll': 'ball'\\
'bthe': 'the'\\
'bucket': 'bucket'\\
'bue': 'blue'\\
'bule': 'blue'\\
'bus': 'bus'\\
'buses': 'bus'\\
'bush': 'bush'\\
'bushes': 'bush'\\
'busket': 'basket'\\
'but': 'but'\\
'bved': 'bed'\\
'bw': 'between'\\
'bwd': 'bed'\\
'bx': 'box'\\
'by': 'by'\\
'bye': 'by'\\
'c': ''\\
'cabinet': 'cabinet'\\
'cabinets': 'cabinet'\\
'cactus': 'cactus'\\
'cacuts': 'cactus'\\
'cae': 'car'\\
'cahir': 'chair'\\
'cair': 'chair'\\
'came': 'come'\\
'camera': 'camera'\\
'camper': 'camper'\\
'can': 'can'\\
'cant': 'cant'\\
'canu': 'can you'\\
'car': 'car'\\
'cardboard': 'cupboard'\\
'care': 'car'\\
'cars': 'car'\\
'casio': 'keyboard'\\
'cat': 'cat'\\
'catch': 'catch'\\
'catcus': 'cactus'\\
'cautus': 'cactus'\\
'ccolor': 'color'\\
'ceiling': 'ceiling'\\
'ceilling': 'ceiling'\\
'celing': 'ceiling'\\
'celling': 'ceiling'\\
'center': 'center'\\
'centre': 'center'\\
'chai': 'chair'\\
'chair': 'chair'\\
'chairs': 'chair'\\
'chait': 'chair'\\
'change': 'change'\\
'chaor': 'chair'\\
'char': 'chair'\\
'chari': 'chair'\\
'chauffer': 'chauffeur'\\
'check': 'check'\\
'chiar': 'chair'\\
'chir': 'chair'\\
'choice': 'choice'\\
'chopper': 'helicopter'\\
'choppers': 'helicopter'\\
'chzir': 'chair'\\
'circle': 'circle'\\
'circular': 'circular'\\
'clean': 'clean'\\
'cleaner': 'cleaner'\\
'clear': 'clear'\\
'cleat': 'clear'\\
'clock': 'clock'\\
'clor': 'color'\\
'clored': 'color'\\
'close': 'close'\\
'closer': 'closer'\\
'closet': 'closet'\\
'closets': 'closet'\\
'clour': 'color'\\
'cn': 'can'\\
'cna': 'can'\\
'co': 'go'\\
'coffe': 'coffee'\\
'coffee': 'coffee'\\
'coffeemug': 'coffee mug'\\
'collect': 'collect'\\
'collors': 'color'\\
'colo': 'color'\\
'colo0r': 'color'\\
'coloe': 'color'\\
'colof': 'color'\\
'cololr': 'color'\\
'color': 'color'\\
'colore': 'color'\\
'colored': 'color'\\
'colored"': 'color'\\
'coloredd': 'color'\\
'colorof': 'color'\\
'coloror': 'color'\\
'colorright': 'color right'\\
'colors': 'color'\\
'colorsyn': 'color yes no'\\
'colort': 'color'\\
'coloryesno': 'color yes no'\\
'coloryn': 'color yes no'\\
'colot': 'color'\\
'coloum': 'color'\\
'colour': 'color'\\
'coloured': 'color'\\
'colours': 'color'\\
'colouryesno': 'color yes no'\\
'colr': 'color'\\
'colred': 'color'\\
'colro': 'color'\\
'column': 'column'\\
'colur': 'color'\\
'com': 'come'\\
'combined': 'combine'\\
'come': 'come'\\
'compare': 'compare'\\
'compared': 'compare'\\
'comparing': 'comparing'\\
'compartment': 'compartment'\\
'compartments': 'compartment'\\
'conditioner': 'conditioner'\\
'conner': 'corner'\\
'consist': 'consist'\\
'consists': 'consist'\\
'cont': 'count'\\
'contain': 'contain'\\
'container': 'container'\\
'containers': 'container'\\
'contains': 'contain'\\
'contents': 'contents'\\
'coor': 'color'\\
'coored': 'color'\\
'coour': 'color'\\
'copunt': 'count'\\
'corner': 'corner'\\
'corners': 'corner'\\
'cornor': 'corner'\\
'cornr': 'corner'\\
'correct': 'correct'\\
'cot': 'cot'\\
'cou': ''\\
'couch': 'couch'\\
'couches': 'couch'\\
'could': 'could'\\
'coumt': 'count'\\
'coun': 'count'\\
'counr': 'count'\\
'count': 'count'\\
'counts': 'count'\\
'countthe': 'count the'\\
'cout': 'count'\\
'cover': 'cover'\\
'cream': 'cream'\\
'cricket': 'cricket'\\
'crimson': 'crimson'\\
'csn': 'can'\\
'cub': 'cup'\\
'cubboard': 'cupboard'\\
'cube': 'cube'\\
'cubebox': 'box'\\
'cubes': 'cube'\\
'cuboard': 'cupboard'\\
'cude': 'could'\\
'cuo': 'cup'\\
'cuoboard': 'cupboard'\\
'cup': 'cup'\\
'cupboard': 'cupboard'\\
'cupboards': 'cupboard'\\
'cupboarf': 'cupboard'\\
'cups': 'cup'\\
'currently': 'currently'\\
'cushion': 'cushion'\\
'cushions': 'cushion'\\
'cusion': 'cushion'\\
'cute': 'cute'\\
'cyan': 'cyan'\\
'cycle': 'cycle'\\
'd': ''\\
'dame': 'same'\\
'dance': 'dance'\\
'dancew': 'dance'\\
'dark': 'dark'\\
'darker': 'darker'\\
'dck': 'duck'\\
'ddoor': 'door'\\
'dear': 'door'\\
'deck': 'deck'\\
'decribe': 'describe'\\
'der': ''\\
'describe': 'describe'\\
'desk': 'desk'\\
'dhelf': 'shelf'\\
'diagonal': 'diagonal'\\
'diagonally': 'diagonal'\\
'did': 'did'\\
'diferent': 'different'\\
'diff': 'different'\\
'diffent': 'different'\\
'differ': 'different'\\
'difference': 'difference'\\
'different': 'different'\\
'differentiate': 'differentiate'\\
'differently': 'differently'\\
'differnet': 'different'\\
'differnt': 'different'\\
'diffferent': 'different'\\
'diffrent': 'different'\\
'diiferent': 'different'\\
'dining': 'dining'\\
'direction': 'direction'\\
'distance': 'distance'\\
'distrub': 'disturb'\\
'disturb': 'disturb'\\
'disturbing': 'disturbing'\\
'dloor': 'floor'\\
'do': 'do'\\
'doddle': ''\\
'doe': 'does'\\
'does': 'does'\\
'dofferent': 'different'\\
'dog': 'dog'\\
'doing': 'doing'\\
'doll': 'doll'\\
'dolls': 'doll'\\
'dont': 'dont'\\
'doo': 'door'\\
'dooe': 'door'\\
'dooor': 'door'\\
'door': 'door'\\
'doors': 'door'\\
'doot': 'door'\\
'dor': 'door'\\
'dorr': 'door'\\
'dose': 'does'\\
'down': 'down'\\
'dplace': 'place'\\
'dr': 'drop'\\
'drag': 'drag'\\
'dragon': 'drag on'\\
'draier': 'drawer'\\
'drawer': 'drawer'\\
'dresser': 'wardrobe'\\
'drier': 'drier'\\
'driers': 'drier'\\
'dries': 'drier'\\
'driyer': 'drier'\\
'drop': 'drop'\\
'dropping': 'dropping'\\
'dryeer': 'drier'\\
'dryer': 'drier'\\
'dryers': 'drier'\\
'dsame': 'same'\\
'dtool': 'tool'\\
'duch': 'duck'\\
'duck': 'duck'\\
'duckl': 'duck'\\
'ducks': 'duck'\\
'ducky': 'duck'\\
'duckys': 'duck'\\
'ducts': 'duck'\\
'duk': 'duck'\\
'dull': 'duck'\\
'dusk': 'duck'\\
'duxk': 'duck'\\
'dyer': 'drier'\\
'e': ''\\
'each': 'each'\\
'eachother': 'each other'\\
'ear': 'ear'\\
'earphone': 'headphone'\\
'earphones': 'headphone'\\
'ebd': 'bed'\\
'ebed': 'bed'\\
'eblue': 'blue'\\
'ecist': 'exist'\\
'ecolour': 'color'\\
'ed': 'bed'\\
'edoor': 'floor'\\
'efloor': 'floor'\\
'egreen': 'green'\\
'ehat': 'what'\\
'ehere': 'where'\\
'ehich': 'which'\\
'eight': 'eight'\\
'eighteen': 'eighteen'\\
'eith': 'with'\\
'either': 'either'\\
'elevate': 'elevate'\\
'else': 'else'\\
'em': 'me'\\
'eme': 'me'\\
'empty': 'empty'\\
'end': 'end'\\
'engine': 'engine'\\
'engines': 'engine'\\
'entire': 'entire'\\
'eobjects': 'object'\\
'eoof': 'roof'\\
'eorange': 'orange'\\
'eow': 'row'\\
'epink': 'pink'\\
'eplace': 'place'\\
'equal': 'equal'\\
'ered': 'red'\\
'eroof': 'roof'\\
'eroom': 'room'\\
'et': 'what'\\
'etable': 'table'\\
'ethe': 'the'\\
'ever': 'ever'\\
'every': 'every'\\
'everything': 'everything'\\
'exact': 'exact'\\
'exactly': 'exactly'\\
'except': 'except'\\
'exist': 'exist'\\
'exista': 'exist'\\
'existed': 'existed'\\
'existing': 'existing'\\
'exists': 'exist'\\
'existsin': 'exist in'\\
'exit': 'exist'\\
'exits': 'exist'\\
'exixts': 'exist'\\
'exsists': 'exist'\\
'exsits': 'exist'\\
'f': ''\\
'face': 'face'\\
'facing': 'facing'\\
'fall': 'fall'\\
'falling': 'falling'\\
'fan': 'fan'\\
'far': 'far'\\
'farme': 'far'\\
'favorite': 'favorite'\\
'favourate': 'favorite'\\
'favourite': 'favorite'\\
'feel': 'feel'\\
'feet': 'feet'\\
'few': 'few'\\
'fifteen': 'fifteen'\\
'fifty': 'fifty'\\
'fill': 'fill'\\
'find': 'find'\\
'finished': 'finished'\\
'finishing': 'finishing'\\
'first': 'first'\\
'five': 'five'\\
'flight': 'flight'\\
'flip': 'flip'\\
'flloor': 'floor'\\
'fllor': 'floor'\\
'floo': 'floor'\\
'flooe': 'floor'\\
'flooer': 'floor'\\
'flooor': 'floor'\\
'floor': 'floor'\\
'flooring': 'floor'\\
'flooryn': 'floor'\\
'floot': 'floor'\\
'flor': 'floor'\\
'florr': 'floor'\\
'flower': 'flower'\\
'flowerpot': 'flower pot'\\
'flying': 'flying'\\
'fo': 'of'\\
'font': 'front'\\
'food': 'food'\\
'foor': 'floor'\\
'foorball': 'flootball'\\
'foot': 'foot'\\
'footbal': 'football'\\
'football': 'football'\\
'footballs': 'football'\\
'for': 'for'\\
'force': 'force'\\
'form': 'from'\\
'formate': 'formation'\\
'fornt': 'front'\\
'foue': 'four'\\
'four': 'four'\\
'fourteen': 'fourteen'\\
'frame': 'frame'\\
'framed': 'frame'\\
'frames': 'frame'\\
'freen': 'green'\\
'fridge': 'refridgerator'\\
'frier': 'drier'\\
'frm': 'from'\\
'frnt': 'front'\\
'from': 'from'\\
'fron': 'front'\\
'front': 'front'\\
'fryer': 'drier'\\
'fuck': 'duck'\\
'furniture': 'furniture'\\
'g': ''\\
'gadget': 'gadget'\\
'gat': 'at'\\
'gather': 'gather'\\
'gave': 'gave'\\
'geen': 'green'\\
'gereen': 'green'\\
'get': 'get'\\
'gho': 'go'\\
'gimme': 'give me'\\
'give': 'give'\\
'glass': 'glass'\\
'glider': 'airplane'\\
'go': 'go'\\
'gold': 'gold'\\
'golden': 'gold'\\
'gonear': 'go near'\\
'good': 'good'\\
'goodbad': 'good bad'\\
'gostand': 'go stand'\\
'got': 'got'\\
'goto': 'go to'\\
'gp': 'go'\\
'grab': 'grab'\\
'gray': 'gray'\\
'gree': 'green'\\
'greeb': 'green'\\
'greeen': 'green'\\
'greem': 'green'\\
'green': 'green'\\
'greenbook': 'green book'\\
'greentable': 'green table'\\
'gren': 'green'\\
'grenn': 'green'\\
'grey': 'grey'\\
'grey+white': 'grey and white'\\
'ground': 'ground'\\
'group': 'group'\\
'grreen': 'green'\\
'grren': 'green'\\
'gthe': 'the'\\
'gun': ''\\
'gyrocopter': 'helicopter'\\
'ha': 'have'\\
'had': 'had'\\
'haedset': 'headphones'\\
'hai': 'hair'\\
'haie': 'hair'\\
'hair': 'hair'\\
'hairdrier': 'hair drier'\\
'hairdriers': 'hair drier'\\
'hairdryer': 'hair drier'\\
'hairdryers': 'hair drier'\\
'hairdyer': ' hair drier'\\
'hand': 'hand'\\
'handle': 'handle'\\
'handover': 'hand over'\\
'hands': 'hand'\\
'hang': 'hang'\\
'hanndover': 'hand over'\\
'has': 'has'\\
'have': 'have'\\
'havethe': 'have the'\\
'havew': 'have'\\
'having': 'have'\\
'he': 'he'\\
'head': 'head'\\
'headphoes': 'headphone'\\
'headphon': 'headphone'\\
'headphone': 'headphone'\\
'headphones': 'headphone'\\
'headphons': 'headphone'\\
'headpones': 'headphone'\\
'headset': 'headphone'\\
'headsets': 'headphone'\\
'headst': 'headphone'\\
'heap': 'heap'\\
'heart': 'heart'\\
'heassets': 'headphone'\\
'hed': 'head'\\
'height': 'height'\\
'heir': 'hair'\\
'helcopter': 'helicopter'\\
'held': 'held'\\
'helicap': 'helicopter'\\
'helicaptor': 'helicopter'\\
'helicofter': 'helicopter'\\
'helicopeter': 'helicopter'\\
'helicopter': 'helicopter'\\
'helicopters': 'helicopter'\\
'helicoptor': 'helicopter'\\
'helicoter': 'helicopter'\\
'helipad': 'helicopter'\\
'help': 'help'\\
'here': 'here'\\
'hesdset': 'headphone'\\
'hide': 'hide'\\
'his': 'his'\\
'hit': 'hit'\\
'hnd': 'hand'\\
'ho': 'how'\\
'hoe': 'how'\\
'hoist': 'hoist'\\
'hold': 'hold'\\
'holder': 'hold'\\
'holding': 'hold'\\
'house': 'house'\\
'how': 'how'\\
'howmany': 'how many'\\
'hows': 'how'\\
'hte': 'the'\\
'human': 'human'\\
'hundred': 'hundred'\\
'i': 'i'\\
'ibjects': 'object'\\
'id': 'is'\\
'identical': 'identical'\\
'ids': 'is'\\
'if': 'if'\\
'iift': 'lift'\\
'iin': 'in'\\
'iis': 'is'\\
'im': 'im'\\
'in': 'in'\\
'in-front': 'in front'\\
'ina': 'in a'\\
'inbetween': 'in between'\\
'including': 'including'\\
'infont': 'in front'\\
'infornt': 'in front'\\
'infron': 'in front'\\
'infront': 'in front'\\
'infrontof': 'in front of'\\
'infrot': 'in front'\\
'inj': 'in'\\
'ink': 'in'\\
'inn': 'in'\\
'insame': 'in same'\\
'inside': 'inside'\\
'instead': 'instead'\\
'int': 'in'\\
'international': 'international'\\
'inthe': 'in the'\\
'into': 'into'\\
'ion': 'in'\\
'ios': 'is'\\
'is': 'is'\\
'isblue': 'is blue'\\
'ison': 'is on'\\
'ispink': 'is pink'\\
'isthe': 'is the'\\
'isthere': 'is there'\\
'it': 'it'\\
'iteams': 'item'\\
'item': 'item'\\
'items': 'item'\\
'its': 'its'\\
'itself': 'itself'\\
'jeep': 'jeep'\\
'jug': 'jug'\\
'jugs': 'jug'\\
'jump': 'jump'\\
'just': 'just'\\
'keeep': 'keep'\\
'keep': 'keep'\\
'keeping': 'keeping'\\
'kep': 'keep'\\
'kepp': 'keep'\\
'kept': 'keep'\\
'key': 'key'\\
'keybaord': 'keyboard'\\
'keyboard': 'keyboard'\\
'keyboards': 'keyboard'\\
'keyboars': 'keyboard'\\
'khan': ''\\
'kick': 'kick'\\
'kicking': 'kicking'\\
'kift': 'lift'\\
'kit': 'it'\\
'knock': 'knock'\\
'know': 'know'\\
'l': ''\\
'lace': 'place'\\
'laeger': 'larger'\\
'lager': 'larger'\\
'lam': 'lamp'\\
'lamb': 'lamp'\\
'lamo': 'lamp'\\
'lamp': 'lamp'\\
'lampp': 'lamp'\\
'lamps': 'lamp'\\
'lane': 'lane'\\
'lap': 'lamp'\\
'lapms': 'lamp'\\
'laptop': 'laptop'\\
'large': 'large'\\
'larger': 'larger'\\
'largerbed': 'larger bed'\\
'largermirror': 'larger mirror'\\
'largersmaller': 'larger smaller'\\
'largersmalleror': 'larger smaller'\\
'largersmallersame': 'larger smaller same'\\
'largerthe': 'larger the'\\
'largest': 'largest'\\
'last': 'last'\\
'later': 'later'\\
'lavender': 'lavender'\\
'laying': 'laying'\\
'ledge': 'ledge'\\
'left': 'left'\\
'leg': 'leg'\\
'leged': 'legged'\\
'legs': 'leg'\\
'length': 'length'\\
'less': 'less'\\
'let': 'let'\\
'lft': 'lift'\\
'li': 'lift'\\
'lidt': 'lift'\\
'lie': 'like'\\
'liek': 'like'\\
'lies': 'lies'\\
'lif': 'lift'\\
'life': 'lift'\\
'lifft': 'lift'\\
'lift': 'lift'\\
'liftb': 'lift'\\
'lifted': 'lifted'\\
'lifting': 'lifting'\\
'liftt': 'lift'\\
'liftthe': 'lift the'\\
'lify': 'lift'\\
'ligher': 'lighter'\\
'light': 'light'\\
'lighter': 'lighter'\\
'ligt': 'lift'\\
'like': 'like'\\
'liked': 'liked'\\
'line': 'line'\\
'lines': 'line'\\
'link': 'link'\\
'linw': 'line'\\
'lion': 'lion'\\
'list': 'list'\\
'lit': 'lift'\\
'lite': 'light'\\
'litf': 'lift'\\
'lith': 'lift'\\
'llift': 'lift'\\
'lloking': 'looking'\\
'lmap': 'lamp'\\
'loacted': 'located'\\
'locate': 'locate'\\
'located': 'located'\\
'location': 'location'\\
'locking': 'looking'\\
'loco': 'locomotive'\\
'locomotive': 'locomotive'\\
'locomotives': 'locomotive'\\
'loft': 'lift'\\
'loking': 'looking'\\
'lokking': 'looking'\\
'longer': 'longer'\\
'looing': 'looking'\\
'look': 'look'\\
'lookig': 'looking'\\
'lookin': 'looking'\\
'looking': 'looking'\\
'lookingat': 'looking at'\\
'lookng': 'looking'\\
'looks': 'look'\\
'loooking': 'looking'\\
'loor': 'door'\\
'ls': 'is'\\
'luft': 'lift'\\
'lying': 'lying'\\
'm': ''\\
'ma': 'a'\\
'magenta': 'magenta'\\
'main': 'main'\\
'majority': 'majority'\\
'make': 'make'\\
'man': 'man'\\
'manner': 'manner'\\
'mant': 'many'\\
'manty': 'many'\\
'many': 'many'\\
'maroon': 'maroon'\\
'mat': 'mat'\\
'matching': 'matching'\\
'matrees': 'mattress'\\
'matress': 'mattress'\\
'matresses': 'mattress'\\
'matters': 'mattress'\\
'mattres': 'mattress'\\
'mattress': 'mattress'\\
'mattresses': 'mattress'\\
'maximum': 'maximum'\\
'may': 'may'\\
'me': 'me'\\
'mear': 'near'\\
'mee': 'me'\\
'mention': 'mention'\\
'mer': 'me'\\
'mess': 'mess'\\
'methe': 'me the'\\
'mew': 'me'\\
'mg': 'mug'\\
'middle': 'middle'\\
'mind': 'mind'\\
'mine': 'mine'\\
'miror': 'mirror'\\
'mirors': 'mirror'\\
'mirrior': 'mirror'\\
'mirriors': 'mirror'\\
'mirrir': 'mirror'\\
'mirroe': 'mirror'\\
'mirron': 'mirror'\\
'mirror': 'mirror'\\
'mirrors': 'mirror'\\
'mirros': 'mirror'\\
'mirrow': 'mirror'\\
'mirrror': 'mirror'\\
'mnay': 'many'\\
'mne': 'me'\\
'mobile': 'mobile'\\
'moment': 'moment'\\
'more': 'more'\\
'morror': 'mirror'\\
'most': 'most'\\
'movable': 'movable'\\
'move': 'move'\\
'movee': 'move'\\
'moving': 'moving'\\
'mow': 'now'\\
'mp': ''\\
'mr': 'me'\\
'mu': 'mug'\\
'much': 'much'\\
'muf': 'mug'\\
'mug': 'mug'\\
'mugs': 'mug'\\
'muh': 'mug'\\
'mw': 'me'\\
'my': 'my'\\
'myg': 'my'\\
'n': ''\\
'nad': 'and'\\
'naer': 'near'\\
'nall': 'ball'\\
'name': 'name'\\
'navy': 'navy'\\
'nay': 'navy'\\
'nbed': 'bed'\\
'nbox': 'box'\\
'nd': 'and'\\
'nder': 'under'\\
'ne': 'me'\\
'nea': 'near'\\
'near': 'near'\\
'nearby': 'nearby'\\
'neare': 'near'\\
'nearest': 'nearest'\\
'nearme': 'near me'\\
'neart': 'near to'\\
'nearto': 'near to'\\
'neat': 'neat'\\
'neath': 'beneath'\\
'ned': 'bed'\\
'neqar': 'near'\\
'ner': 'near'\\
'nera': 'near'\\
'nerar': 'near'\\
'nesr': 'near'\\
'nect': 'next'\\
'nice': 'nice'\\
'nine': 'nine'\\
'nineteen': 'nineteen'\\
'niw': 'now'\\
'nk': 'no'\\
'no': 'no'\\
'nonyellow': 'non yellow'\\
'nonred': 'non red'\\
'nonblue': 'non blue'\\
'nongreen': 'non green'\\
'nonviolet': 'non violet'\\
'nonwhite': 'non white'\\
'nonblack': 'non black'\\
'nonbrown': 'non brown'\\
'nonpink': 'non pink'\\
'noe': 'now'\\
'noew': 'now'\\
'nof': 'number of'\\
'noof': 'number of'\\
'not': 'not'\\
'notice': 'notice'\\
'noticed': 'noticed'\\
'noticeing': 'noticing'\\
'noticing': 'noticing'\\
'now': 'now'\\
'no\}': 'now'\\
'nrar': 'near'\\
'nthe': 'the'\\
'nug': 'mug'\\
'num': 'number'\\
'numbe': 'number'\\
'number': 'number'\\
'numberof': 'number of'\\
'numbers': 'number'\\
'ny': 'my'\\
'nxt': 'next'\\
'o': ''\\
'ob': 'object'\\
'obects': 'object'\\
'obejcts': 'object'\\
'obejects': 'object'\\
'obhects': 'object'\\
'obhjects': 'object'\\
'obj': 'object'\\
'objct': 'object'\\
'objcts': 'object'\\
'objec': 'object'\\
'objeccts': 'object'\\
'objecs': 'object'\\
'objecst': 'object'\\
'object': 'object'\\
'objecta': 'object'\\
'objectives': 'object'\\
'objects': 'object'\\
'objectsa': 'object'\\
'objectsin': 'object in'\\
'objectys': 'object'\\
'objest': 'object'\\
'objetcs': 'object'\\
'objets': 'object'\\
'objevts': 'object'\\
'objs': 'object'\\
'obkjects': 'object'\\
'observe': 'observe'\\
'observing': 'observing'\\
'occupies': 'occupies'\\
'ocean': 'ocean'\\
'od': 'of'\\
'odf': 'off'\\
'of': 'of'\\
'ofbed': 'of bed'\\
'off': 'off'\\
'ofthings': 'of things'\\
'ofwall': 'of wall'\\
'og': 'of'\\
'oh': 'on'\\
'oin': 'on'\\
'oink': 'pink'\\
'ois': 'is'\\
'ojects': 'object'\\
'olace': 'place'\\
'olive': 'olive'\\
'olor': 'color'\\
'om': 'on'\\
'omn': 'on'\\
'on': 'on'\\
'onbed': 'on bed'\\
'onbjects': 'object'\\
'once': 'once'\\
'one': 'one'\\
'onfloor': 'on floor'\\
'ongreen': 'on green'\\
'onj': 'on'\\
'onjects': 'object'\\
'onlt': 'only'\\
'only': 'only'\\
'ont': 'on'\\
'ontable': 'on table'\\
'onthe': 'on the'\\
'onto': 'onto'\\
'onwhite': 'on white'\\
'oobjects': 'object'\\
'ook': 'book'\\
'ooks': 'book'\\
'oom': 'room'\\
'oon': 'on'\\
'open': 'open'\\
'oposite': 'opposite'\\
'opp': 'opposite'\\
'opposite': 'opposite'\\
'opposte': 'opposite'\\
'or': 'or'\\
'oraange': 'orange'\\
'orage': 'orange'\\
'oragne': 'orange'\\
'oramge': 'orange'\\
'orane': 'orange'\\
'orang': 'orange'\\
'orange': 'orange'\\
'oranges': 'orange'\\
'order': 'order'\\
'orgreen': 'or green'\\
'ornage': 'orange'\\
'ornge': 'orange'\\
'orsnge': 'orange'\\
'os': 'is'\\
'osfa': 'sofa'\\
'ot': 'not'\\
'otange': 'orange'\\
'other': 'other'\\
'others': 'other'\\
'our': 'our'\\
'out': 'out'\\
'outside': 'outside'\\
'over': 'over'\\
'ow': 'own'\\
'own': 'own'\\
'ox': 'box'\\
'p': ''\\
'pa': ''\\
'pace': 'place'\\
'paino': 'piano'\\
'paint': 'paint'\\
'painted': 'painted'\\
'painting': 'painting'\\
'pair': 'pair'\\
'palace': 'place'\\
'palce': 'place'\\
'pale': 'pale'\\
'pants': 'plant'\\
'parallel': 'parallel'\\
'parrot': 'parrot'\\
'particular': 'particular'\\
'parts': 'parts'\\
'pass': 'pass'\\
'peach': 'peach'\\
'pen': 'pen'\\
'per': 'per'\\
'perform': 'perform'\\
'pet': 'pet'\\
'pf': 'of'\\
'phone': 'phone'\\
'phones': 'phone'\\
'photo': 'photo'\\
'photoframe': 'photo frame'\\
'pi': 'pink'\\
'piano': 'keyboard'\\
'pianos': 'keyboard'\\
'pic': 'pick'\\
'pich': 'pick'\\
'pick': 'pick'\\
'pickup': 'pick up'\\
'picture': 'picture'\\
'pik': 'pick'\\
'pilllow': 'pillow'\\
'pillo': 'pillow'\\
'pilloe': 'pillow'\\
'pillow': 'pillow'\\
'pillowa': 'pillow'\\
'pillows': 'pillow'\\
'piloow': 'pillow'\\
'pilow': 'pillow'\\
'pilows': 'pillow'\\
'pin': 'pink'\\
'ping': 'pink'\\
'pink': 'pink'\\
'pink-': 'pink'\\
'pinkball': 'pink ball'\\
'pinkbook': 'pink book'\\
'pinkl': 'pink'\\
'pinkmirror': 'pink mirror'\\
'pinl': 'pink'\\
'piono': 'piano'\\
'pit': 'put'\\
'pivk': 'pick'\\
'plaace': 'place'\\
'plac': 'place'\\
'place': 'place'\\
'placed': 'placed'\\
'placee': 'place'\\
'places': 'place'\\
'placewhite': 'place white'\\
'placing': 'placing'\\
'plain': 'airplane'\\
'plan': 'airplane'\\
'plane': 'airplane'\\
'planes': 'airplane'\\
'plant': 'plant'\\
'plantpot': 'plant pot'\\
'plants': 'plant'\\
'plave': 'place'\\
'play': 'play'\\
'player': 'player'\\
'players': 'player'\\
'playing': 'playing'\\
'plcae': 'place'\\
'plce': 'place'\\
'plced': 'placed'\\
'please': 'please'\\
'plotted': 'potted'\\
'plus': 'plus'\\
'pn': 'on'\\
'pnik': 'pink'\\
'pnk': 'pink'\\
'pod': 'pot'\\
'poillow': 'pillow'\\
'poition': 'position'\\
'pon': 'on'\\
'ponk': 'pink'\\
'position': 'position'\\
'positionplace': 'position place'\\
'possess': 'possess'\\
'possible': 'possible'\\
'postion': 'position'\\
'pot': 'pot'\\
'pots': 'pot'\\
'potted': 'potted'\\
'pottedplant': 'potted plant'\\
'poy': 'put'\\
'ppink': 'pink'\\
'present': 'present'\\
'presently': 'presently'\\
'products': 'product'\\
'proper': 'proper'\\
'pu': 'put'\\
'puch': 'push'\\
'pudh': 'push'\\
'puink': 'pink'\\
'puit': 'put'\\
'pull': 'pull'\\
'puple': 'purple'\\
'puprle': 'purple'\\
'pur': 'put'\\
'purple': 'purple'\\
'purplr': 'purple'\\
'purpple': 'purple'\\
'pus': 'push'\\
'push': 'push'\\
'pusht': 'push'\\
'pust': 'push'\\
'put': 'put'\\
'putb': 'put'\\
'putt': 'put'\\
'putthe': 'put the'\\
'r': ''\\
'rable': 'table'\\
'rack': 'rack'\\
'racket': 'racket'\\
'rackets': 'racket'\\
'rackl': 'rack'\\
'racks': 'rack'\\
'racqet': 'racket'\\
'raild': 'rail'\\
'raise': 'raise'\\
'randomly': 'randomly'\\
'raw': 'row'\\
'rd': 'red'\\
're': 'red'\\
'real': 'real'\\
'rectangle': 'rectangle'\\
'rectangular': 'rectangular'\\
'red': 'red'\\
'red-': 'red'\\
'redball': 'red ball'\\
'redbook': 'red book'\\
'redbox': 'red box'\\
'reddy': 'teddy'\\
'redmug': 'red mug'\\
'redtable': 'red table'\\
'redyesno': 'red yes no'\\
'reed': 'red'\\
'ref': 'red'\\
'related': 'related'\\
'relative': 'relative'\\
'remove': 'remove'\\
'replace': 'replace'\\
'replacing': 'replacing'\\
'res': 'red'\\
'respect': 'respect'\\
'rhere': 'are here'\\
'right': 'right'\\
'rightnow': 'right now'\\
'ro': 'row'\\
'roa': 'roam'\\
'roam': 'roam'\\
'robat': 'robot'\\
'robbo': 'robot'\\
'robo': 'robot'\\
'roboat': 'robot'\\
'roboats': 'robot'\\
'robor': 'robot'\\
'robort': 'robot'\\
'robos': 'robot'\\
'robot': 'robot'\\
'robots': 'robot'\\
'roboy': 'robot'\\
'robt': 'robot'\\
'rock': 'rocket'\\
'rockect': 'rocket'\\
'rocker': 'rocket'\\
'rockert': 'rocket'\\
'rocket': 'rocket'\\
'rockets': 'rocket'\\
'rockt': 'rocket'\\
'rockts': 'rocket'\\
'roe': 'row'\\
'roew': 'row'\\
'rof': 'row'\\
'roght': 'right'\\
'roiw': 'row'\\
'roket': 'rocket'\\
'rom': 'room'\\
'romm': 'room'\\
'ronot': 'robot'\\
'roo': 'room'\\
'roobo': 'robot'\\
'rood': 'roof'\\
'roof': 'roof'\\
'roofs': 'roof'\\
'rooftop': 'roof'\\
'room': 'room'\\
'room"': 'room'\\
'room-': 'room'\\
'room-yesno': 'room yes no'\\
'roomm': 'room'\\
'roomn': 'room'\\
'roomright': 'room right'\\
'rooms': 'room'\\
'roomyes': 'room yes'\\
'roomyesno': 'room yes no'\\
'roomyn': 'room yes no'\\
'roon': 'room'\\
'roonm': 'room'\\
'rooom': 'room'\\
'root': 'robot'\\
'ropom': 'room'\\
'rorbot': 'robot'\\
'rotate': 'rotate'\\
'round': 'round'\\
'rovket': 'rocket'\\
'rovkets': 'rockets'\\
'row': 'row'\\
'rows': 'row'\\
'roww': 'row'\\
'royal': 'royal'\\
'rpbot': 'robot'\\
'rpoom': 'room'\\
'rpw': 'row'\\
'rred': 'red'\\
'rrom': 'room'\\
'rroom': 'room'\\
'rtable': 'table'\\
'rthe': 'the'\\
'rubber': 'rubber'\\
'run': 'run'\\
'rw': 'row'\\
'rwo': 'row'\\
's': ''\\
'safe': 'safe'\\
'salman': ''\\
'sam': 'same'\\
'same': 'same'\\
'samecolor': 'same color'\\
'samedifferent': 'same different'\\
'samelarger': 'same larger'\\
'samesize': 'same size'\\
'samesmall': 'same small'\\
'sameyesno': 'same yes no'\\
'sameyn': 'same yes no'\\
'samller': 'smaller'\\
'samw': 'same'\\
'sand': 'same'\\
'sane': 'same'\\
'saofa': 'sofa'\\
'satand': 'stand'\\
'say': 'say'\\
'se': 'see'\\
'sea': 'sea'\\
'sea-green': 'sea green'\\
'seagreen': 'sea green'\\
'sealing': 'ceiling'\\
'searching': 'searching'\\
'seconds': 'seconds'\\
'see': 'see'\\
'seee': 'see'\\
'seeing': 'seeing'\\
'self': 'self'\\
'sequence': 'sequence'\\
'set': 'set'\\
'sets': 'set'\\
'setter': 'setter'\\
'setting': 'setting'\\
'seven': 'seven'\\
'seventeen': 'seventeen'\\
'shaded': 'shaded'\\
'shalf': 'shelf'\\
'shape': 'shape'\\
'sheet': 'sheet'\\
'shef': 'shelf'\\
'shekf': 'shelf'\\
'shelf': 'shelf'\\
'shelf"s': 'shelf'\\
'shelfs': 'shelf'\\
'shelve': 'shelf'\\
'shelves': 'shelf'\\
'shettle': 'rocket'\\
'shift': 'shift'\\
'ship': 'ship'\\
'ships': 'ship'\\
'shlef': 'shelf'\\
'shoe': 'show'\\
'should': 'should'\\
'show': 'show'\\
'shuttle': 'rocket'\\
'si': 'is'\\
'side': 'side'\\
'sides': 'side'\\
'sifa': 'sofa'\\
'silver': 'silver'\\
'similar': 'similar'\\
'simple': 'simple'\\
'sin': 'in'\\
'single': 'single'\\
'sit': 'sit'\\
'sitting': 'sitting'\\
'situated': 'situated'\\
'situation': 'situation'\\
'six': 'six'\\
'sixe': 'six'\\
'size': 'size'\\
'sized': 'sized'\\
'sizes': 'sizes'\\
'sizw': 'size'\\
'sky': 'sky'\\
'skyblue': 'sky blue'\\
'sleep': 'sleep'\\
'smae': 'same'\\
'smal': 'small'\\
'small': 'small'\\
'smaller': 'smaller'\\
'smallerlarger': 'smaller larger'\\
'smalleror': 'smaller or'\\
'smallest': 'smallest'\\
'smalllarge': 'small large'\\
'smalllargesame': 'small large same'\\
'smallsamelarge': 'small same large'\\
'sme': 'some'\\
'so': 'so'\\
'soccer': 'soccer'\\
'soda': 'soda'\\
'sodas': 'soda'\\
'sof': 'sofa'\\
'sofa': 'sofa'\\
'sofas': 'sofa'\\
'sofe': 'sofa'\\
'sofs': 'sofa'\\
'solver': 'solver'\\
'some': 'some'\\
'something': 'something'\\
'somewhere': 'somewhere'\\
'son': 'on'\\
'song': 'song'\\
'soor': 'door'\\
'space': 'space'\\
'spacious': 'spacious'\\
'specific': 'specific'\\
'spfa': 'sofa'\\
'sqaure': 'square'\\
'square': 'square'\\
'sre': 'are'\\
'ss': ''\\
'ssame': 'same'\\
'stairs': 'stairs'\\
'stamd': 'stand'\\
'stan': 'stand'\\
'stand': 'stand'\\
'standing': 'standing'\\
'stands': 'stand'\\
'stanf': 'stand'\\
'stans': 'stand'\\
'star': 'stare'\\
'staring': 'staring'\\
'starring': 'staring'\\
'stay': 'stay'\\
'sthe': 'the'\\
'still': 'still'\\
'stnd': 'stand'\\
'stol': 'stool'\\
'stole': 'stool'\\
'stoll': 'stool'\\
'stoo': 'stool'\\
'stookl': 'stool'\\
'stool': 'stool'\\
'stoole': 'stool'\\
'stools': 'stool'\\
'stop': 'stop'\\
'storage': 'storage'\\
'straight': 'straight'\\
'study': 'study'\\
'suck': 'duck'\\
'swap': 'swap'\\
't': ''\\
'ta': 'at'\\
'taable': 'table'\\
'tabble': 'table'\\
'tabe': 'table'\\
'tabel': 'table'\\
'tabke': 'table'\\
'tabkle': 'table'\\
'tabl': 'table'\\
'table': 'table'\\
'tabled': 'table'\\
'tablee': 'table'\\
'tablel': 'table'\\
'tables': 'table'\\
'tablw': 'table'\\
'tabvle': 'table'\\
'take': 'take'\\
'takeoff': 'takeoff'\\
'taking': 'taking'\\
'tale': 'table'\\
'tall': 'tall'\\
'taller': 'taller'\\
'tand': 'and'\\
'tanle': 'table'\\
'tarin': 'duck'\\
'task': 'task'\\
'tavle': 'table'\\
'tbale': 'table'\\
'tble': 'table'\\
'te': 'the'\\
'tea': 'tea'\\
'teaddy': 'teddy'\\
'teaddybear': 'teddy bear'\\
'teaddybears': 'teddy bear'\\
'teady': 'teddy'\\
'teadybear': 'teddy bear'\\
'teal': 'teal'\\
'teble': 'table'\\
'ted': 'red'\\
'tedddy': 'teddy'\\
'teddies': 'teddy'\\
'teddt': 'teddy'\\
'teddu': 'teddy'\\
'teddy': 'teddy'\\
'teddy"s': 'teddy'\\
'teddybear': 'teddy bear'\\
'teddybears': 'teddy bear'\\
'teddys': 'teddy'\\
'tedy': 'teddy'\\
'teedy': 'teddy'\\
'teedys': 'teddy'\\
'teh': 'the'\\
'tel': 'tell'\\
'television': 'television'\\
'tell': 'tell'\\
'tellow': 'yellow'\\
'ten': 'ten'\\
'tennis': 'tennis'\\
'tesno': 'yes no'\\
'tge': 'the'\\
'tghe': 'the'\\
'th': 'the'\\
'tha': 'the'\\
'tham': 'than'\\
'than': 'than'\\
'that': 'that'\\
'thats': 'thats'\\
'the': 'the'\\
'theb': 'the'\\
'theball': 'the ball'\\
'theballs': 'the ball'\\
'thebed': 'the bed'\\
'theblue': 'the blue'\\
'thebox': 'the box'\\
'thechair': 'the chair'\\
'thecolor': 'the color'\\
'thed': 'the'\\
'thedoor': 'the door'\\
'theduck': 'the duck'\\
'thee': 'the'\\
'theere': 'there'\\
'thefloor': 'the floor'\\
'thegreen': 'the green'\\
'their': 'their'\\
'them': 'them'\\
'then': 'then'\\
'theorange': 'the orange'\\
'thepink': 'the pink'\\
'thepurple': 'the purple'\\
'ther': 'there'\\
'there': 'there'\\
'thered': 'the red'\\
'theres': 'the red'\\
'theroof': 'the roof'\\
'theroom': 'the room'\\
'these': 'these'\\
'thetable': 'the table'\\
'theviolet': 'the violet'\\
'thew': 'the'\\
'thewhite': 'the white'\\
'they': 'they'\\
'theyellow': 'the yellow'\\
'thge': 'the'\\
'thhe': 'the'\\
'thier': 'their'\\
'thing': 'thing'\\
'things': 'thing'\\
'think': 'think'\\
'thinks': 'think'\\
'third': 'third'\\
'thirteen': 'thirteen'\\
'this': 'this'\\
'thje': 'the'\\
'thjere': 'there'\\
'those': 'those'\\
'thr': 'the'\\
'thre': 'the'\\
'thred': 'the red'\\
'three': 'three'\\
'threre': 'there'\\
'through': 'through'\\
'throw': 'throw'\\
'throwing': 'throwing'\\
'ths': 'the'\\
'tht': 'that'\\
'thtree': 'three'\\
'thw': 'the'\\
'thwe': 'the'\\
'thye': 'the'\\
'ti': 'to'\\
'till': 'until'\\
'time': 'time'\\
'times': 'time'\\
'tis': 'is'\\
'tje': 'the'\\
'tjhe': 'the'\\
'to': 'to'\\
'toch': 'touch'\\
'tocket': 'rocket'\\
'tof': 'of'\\
'together': 'together'\\
'tome': 'to me'\\
'too': 'to'\\
'top': 'top'\\
'tot': 'toy'\\
'total': 'total'\\
'tou': 'two'\\
'touch': 'touch'\\
'touching': 'touching'\\
'touvh': 'touching'\\
'tow': 'two'\\
'towards': 'towards'\\
'tower': 'tower'\\
'toy': 'toy'\\
'toyrobot': 'toy robot'\\
'toys': 'toy'\\
'tpouch': 'touch'\\
'tpuch': 'touch'\\
'trable': 'table'\\
'train': 'train'\\
'traingle': 'triangle'\\
'trains': 'train'\\
'tray': 'tray'\\
'tree': 'tree'\\
'trhe': 'the'\\
'triangle': 'triangle'\\
'triangular': 'triangular'\\
'trow': 'throw'\\
'truck': 'truck'\\
'trucks': 'trucks'\\
'true': 'true'\\
'try': 'try'\\
'tsble': 'table'\\
'tthe': 'the'\\
'tub': 'tub'\\
'tubelight': 'tube light'\\
'turn': 'turn'\\
'turquoise': 'turqoise'\\
'tv': 'tv'\\
'twenty': 'twenty'\\
'two': 'two'\\
'tyhe': 'the'\\
'types': 'types'\\
'u': 'you'\\
'uder': 'under'\\
'ug': 'mug'\\
'uing': 'using'\\
'uisng': 'using'\\
'unde': 'under'\\
'undeer': 'under'\\
'under': 'under'\\
'underneath': 'underneath'\\
'underthe': 'under the'\\
'undr': 'under'\\
'undrneath': 'underneath'\\
'until': 'until'\\
'up': 'up'\\
'upon': 'upon'\\
'upside': 'upside'\\
'ur': 'your'\\
'us': 'is'\\
'use': 'use'\\
'usimg': 'using'\\
'usin': 'using'\\
'usinf': 'using'\\
'using': 'using'\\
'usinggreen': 'using green'\\
'usingthe': 'using the'\\
'usingyellow': 'using yellow'\\
'vaccum': 'vacuum'\\
'van': 'van'\\
'vch': 'which'\\
'vchs': 'which is'\\
've': ''\\
'vechile': 'vehicle'\\
'vechiles': 'vehicle'\\
'vehicle': 'vehicle'\\
'vehicles': 'vehicle'\\
'very': 'very'\\
'vhair': 'chair'\\
'view': 'view'\\
'viloet': 'violet'\\
'violet': 'violet'\\
'visible': 'visible'\\
'vme': 'me'\\
'voilet': 'violet'\\
'volley': 'volley'\\
'volor': 'color'\\
'volour': 'color'\\
'volvo': 'volvo'\\
'vount': 'count'\\
'vrs': 'where is'\\
'vt': 'what'\\
'vth': 'with'\\
'vts': 'what is'\\
'w': ''\\
'wa': ''\\
'wadrobe': 'wardrobe'\\
'waht': 'what'\\
'wait': 'wait'\\
'wakll': 'walk'\\
'wal': 'wall'\\
'walk': 'walk'\\
'wall': 'wall'\\
'walla': 'wall'\\
'walll': 'wall'\\
'walls': 'wall'\\
'wals': 'wall'\\
'want': 'want'\\
'wardrobe': 'wardrobe'\\
'wardrobes': 'wardrobe'\\
'wardrof': 'wardrobe'\\
'was': 'was'\\
'wat': 'what'\\
'watch': 'watch'\\
'watching': 'watching'\\
'way': 'way'\\
'we': 'we'\\
'weather': 'whether'\\
'weed': 'weed'\\
'well': 'well'\\
'were': 'were'\\
'wether': 'whether'\\
'wh': ''\\
'wha': 'what'\\
'whaere': 'where'\\
'whar': 'where'\\
'whare': 'where'\\
'what': 'what'\\
'whatare': 'what are'\\
'whatever': 'whatever'\\
'whats': 'what is'\\
'whatyou': 'what you'\\
'whch': 'which'\\
'whcih': 'which'\\
'wheather': 'whether'\\
'wheels': 'wheels'\\
'wheer': 'wheel'\\
'whellers': 'wheeler'\\
'when': 'when'\\
'wher': 'where'\\
'where': 'where'\\
'whereis': 'where is'\\
'wheres': 'where is'\\
'wherre': 'where'\\
'whers': 'where'\\
'wherte': 'where'\\
'whether': 'whether'\\
'whhite': 'white'\\
'whic': 'which'\\
'which': 'which'\\
'whichis': 'which is'\\
'whick': 'which'\\
'whiite': 'white'\\
'while': 'while'\\
'whit': 'white'\\
'white': 'white'\\
'whiteball': 'white ball'\\
'whiteduck': 'white duck'\\
'whitee': 'white'\\
'whites': 'white'\\
'whitw': 'white'\\
'who': 'who'\\
'whote': 'white'\\
'whr': 'where'\\
'whre': 'where'\\
'whst': 'what'\\
'wht': 'what'\\
'whta': 'what'\\
'whte': 'white'\\
'will': 'will'\\
'wimdow': 'window'\\
'windo': 'window'\\
'windoe': 'window'\\
'windos': 'window'\\
'window': 'window'\\
'windowa': 'window'\\
'windows': 'window'\\
'wise': 'wise'\\
'wish': 'wish'\\
'wit': 'with'\\
'wite': 'white'\\
'with': 'with'\\
'witha': 'with a'\\
'withe': 'with the'\\
'wither': 'whether'\\
'withorange': 'with orange'\\
'without': 'without'\\
'wll': 'wall'\\
'wodden': 'wooden'\\
'wondow': 'window'\\
'wooden': 'wooden'\\
'wordrobe': 'wardrobe'\\
'would': 'would'\\
'wt': 'what'\\
'wthat': 'what'\\
'wts': 'what is'\\
'ww': ''\\
'wwww': ''\\
'wwwww': ''\\
'wwwwww': ''\\
'wwwwwwwwww': ''\\
'xylophone': 'xylophone'\\
'y': 'yes'\\
'ye': 'yes'\\
'yeallow': 'yellow'\\
'yelllow': 'yellow'\\
'yello': 'yellow'\\
'yelloe': 'yellow'\\
'yelloew': 'yellow'\\
'yellow': 'yellow'\\
'yellow-': 'yellow'\\
'yellowball': 'yellow ball'\\
'yelolow': 'yellow'\\
'yeloow': 'yellow'\\
'yelow': 'yellow'\\
'yeno': 'yes no'\\
'yes': 'yes'\\
'yesno': 'yes no'\\
'yesno0': 'yes no'\\
'yesor': 'yes or'\\
'yhe': 'the'\\
'yn': 'yes no'\\
'yno': 'yes no'\\
'yo': 'you'\\
'yor': 'your'\\
'you': 'you'\\
'youlooking': 'you looking'\\
'your': 'your'\\
'youre': 'youre'\\
'yourself': 'yourself'\\
'youself': 'yourself'\\
'ypu': 'you'\\
'ytellow': 'yellow'\\
'ywllow': 'yellow'\\
'\{yes': 'yes'\\
\end{multicols}
\end{flushleft}
\end{footnotesize}

\end{document}